\documentclass[10pt,twocolumn,letterpaper]{article}

\usepackage{3dv}
\usepackage{times}
\usepackage{epsfig}
\usepackage{graphicx}
\usepackage{amsmath}
\usepackage{amssymb}

\usepackage[pagebackref=true,breaklinks=true,letterpaper=true,colorlinks,bookmarks=false]{hyperref}

\usepackage{multirow}
\usepackage{booktabs}
\usepackage{inconsolata}
\usepackage{authblk}

\usepackage[dvipsnames]{xcolor}
\definecolor{pink}{rgb}{1,0,1}
\definecolor{asparagus}{rgb}{0.53, 0.66, 0.42}

\newcommand{\revision}[1]{{\leavevmode\color{black}#1}}

\newcommand{\denselist}{\itemsep 0pt\parsep=0pt\partopsep 0pt\vspace{-\topsep}}

\threedvfinalcopy 

\def\threedvPaperID{66} 

\ifthreedvfinal\fi
\begin{document}

\title{Shape from Tracing: Towards Reconstructing 3D Object Geometry and\\SVBRDF Material from Images via Differentiable Path Tracing}



\renewcommand*{\Authsep}{\quad}
\renewcommand*{\Authand}{\quad}
\renewcommand*{\Authands}{\quad}
\makeatletter \renewcommand\AB@affilsepx{\quad \protect\Affilfont} \makeatother

\author[1, 2]{Purvi Goel}
\author[1]{Loudon Cohen}
\author[1]{James Guesman}
\author[1]{Vikas Thamizharasan}
\author[1]{\\James Tompkin}
\author[1]{Daniel Ritchie}
\affil[1]{\small{Brown University}}
\affil[2]{\small{Stanford University}}

\maketitle

\begin{abstract}

Reconstructing object geometry and material from multiple views typically requires optimization. Differentiable path tracing is an appealing framework as it can reproduce complex appearance effects. However, it is difficult to use due to high computational cost. In this paper, we explore how to use differentiable ray tracing to refine an initial coarse mesh and per-mesh-facet material representation. In simulation, we find that it is possible to reconstruct fine geometric and material detail from low resolution input views, allowing high-quality reconstructions in a few hours despite the expense of path tracing. The reconstructions successfully disambiguate shading, shadow, and global illumination effects such as diffuse interreflection from material properties. We demonstrate the impact of different geometry initializations, including space carving, multi-view stereo, and 3D neural networks. Finally, with input captured using smartphone video and a consumer $360^\circ$ camera for lighting estimation, we also show how to refine initial reconstructions of real-world objects in unconstrained environments.
\end{abstract}

\section{Introduction}
\label{sec:intro}

%
Reconstructing digital representations of the appearance of objects is important to many industries, including visualization, cultural heritage, and entertainment. 
At a minimum, this task requires estimating the shape of the object via its surface geometry, and estimating the material appearance properties of the object.
Recreating these properties accurately by hand requires skill and labor, so automatic reconstruction techniques are useful to complete this task.

Many techniques have been proposed with a common high-level approach: capture multiple views of the object with an imaging sensor, often under varying illumination, to describe the underlying geometry and material properties under appearance assumptions. These techniques can be forward or `bottom up,' by directly estimating object properties from observed sensor data, or can be inverse or `top down,' by optimizing an underlying model until its rendering is consistent with the captured sensor data.

For bottom-up methods, multi-view stereo approaches directly estimate the depth of points on the object surface from calibrated RGB cameras, under a Lambertian surface reflectance assumption.
Time-of-flight and structured light sensors can also directly estimate depth
under simplified reflectance assumptions; depth point clouds can then be fused into volumes for surface reconstruction.
Photometric stereo approaches use RGB cameras to directly estimate surface normal directions from objects exposed to light from different directions,
typically with non-spatially-varying surface albedo and Lambertian or restricted BRDF reflectance models. These material reflectance assumptions cause limitations or inaccuracy in complex shape and material reconstruction. Further, methods may also be limited by their light transport assumptions, e.g., that no diffuse interreflection exists for Lambertian materials.


Top-down approaches suffer these problems in reverse, as the renderer must be able to accurately reproduce the appearance of objects under as few assumptions as possible for shape, material, and light transport. While realistic rendering is possible, any renderer must also be efficient to use in optimization to fit a model to the captured camera view. That is, it must provide gradients which describe the direction of error with respect to the object's shape and material. As such, many differentiable renderers support only simplified camera and geometry (e.g., simplified visibility~\cite{OpenDR,Rhodin_2015_ICCV}), simplified material (e.g., diffuse only~\cite{kato2018renderer}), or simplified light transport (e.g., rasterization~\cite{TensorflowGraphics}). 

However, differentiable \emph{path tracing} methods~\cite{redner,Mitsuba2} capable of simulating global illumination can reproduce and optimize complex appearance with fewer assumptions about geometry, material, and light transport. Path tracing is a theoretically elegant approach, but its application to multi-view object reconstruction is difficult in practice due to the computational complexity of computing derivatives with respect to the object's shape and material properties.

In this paper, we investigate how to reconstruct an object from multi-view images via differentiable path tracing. Given multiple calibrated views of an object under known lighting, represented either by point lights or an HDR environment map, we explore how to reconstruct both the 3D geometry of the object as a surface mesh and the surface material as a spatially-varying Torrance-Sparrow BRDF model. We refine an initial coarse mesh, produced by any one of a variety of reconstruction methods, at the triangle level with a mesh colors SVBRDF representation.  This combination provides coarse-to-fine optimization of both shape and material through geometry subdivision, simplification, and remeshing stages.



We discover with simulated objects that this approach can reconstruct fine geometric and material detail from low-resolution (128$\times$128) target camera views. These reconstructions include the disambiguation of shading and shadows from material variation, the disambiguation of global illumination effects on surface albedo like color bleeding from diffuse interreflection, and the reconstruction of spatially-varying materials with different roughness and specularity. Our efficient representations provide reconstructions within a few hours of optimization, versus naive approaches which can settle at incorrect local minima during gradient-based optimization for inverse rendering. 

In addition, we use physically-based differentiable pathtracing to reconstruct from nearly-unconstrained unstructured real-world data.
Given only a hand-held smartphone camera video of a target object and an environment map captured by a consumer HDR 360 camera, we explore the challenging problem of reconstruction `in the wild.'


In short, we show that efficient representation and optimization of surface geometry and material makes differentiable path tracing a promising technique for high-quality object reconstruction. 
\revision{
We contribute:
\begin{itemize}
\denselist
    \item An investigation into benefits, limitations, and design choices (e.g., parameter space, optimization ordering, and initialization choices) for applying differentiable path tracing to joint geometry/SVBRDF reconstruction in both simulated and real-world settings.
    \item A differentiable mesh colors texture representation~\cite{Yuksel:2010:MC:1731047.1731053} suitable for optimization problems involving meshes with continually-evolving topology.
\denselist
\end{itemize}
}
\revision{Our code and real-world data is available at  \href{http://www.github.com/brownvc/shapefromtracing}{http://www.github.com/brownvc/shapefromtracing}.}
This includes our implementation of mesh colors~\cite{Yuksel:2010:MC:1731047.1731053, Mallett2019}, which to our knowledge has no prior public implementation.


\begin{figure*}[t!]
    
    \begin{minipage}[c]{0.64\textwidth}
        \centering
        \includegraphics[clip,trim={7.35cm 1.35cm 5cm 1.25cm},width=\linewidth]{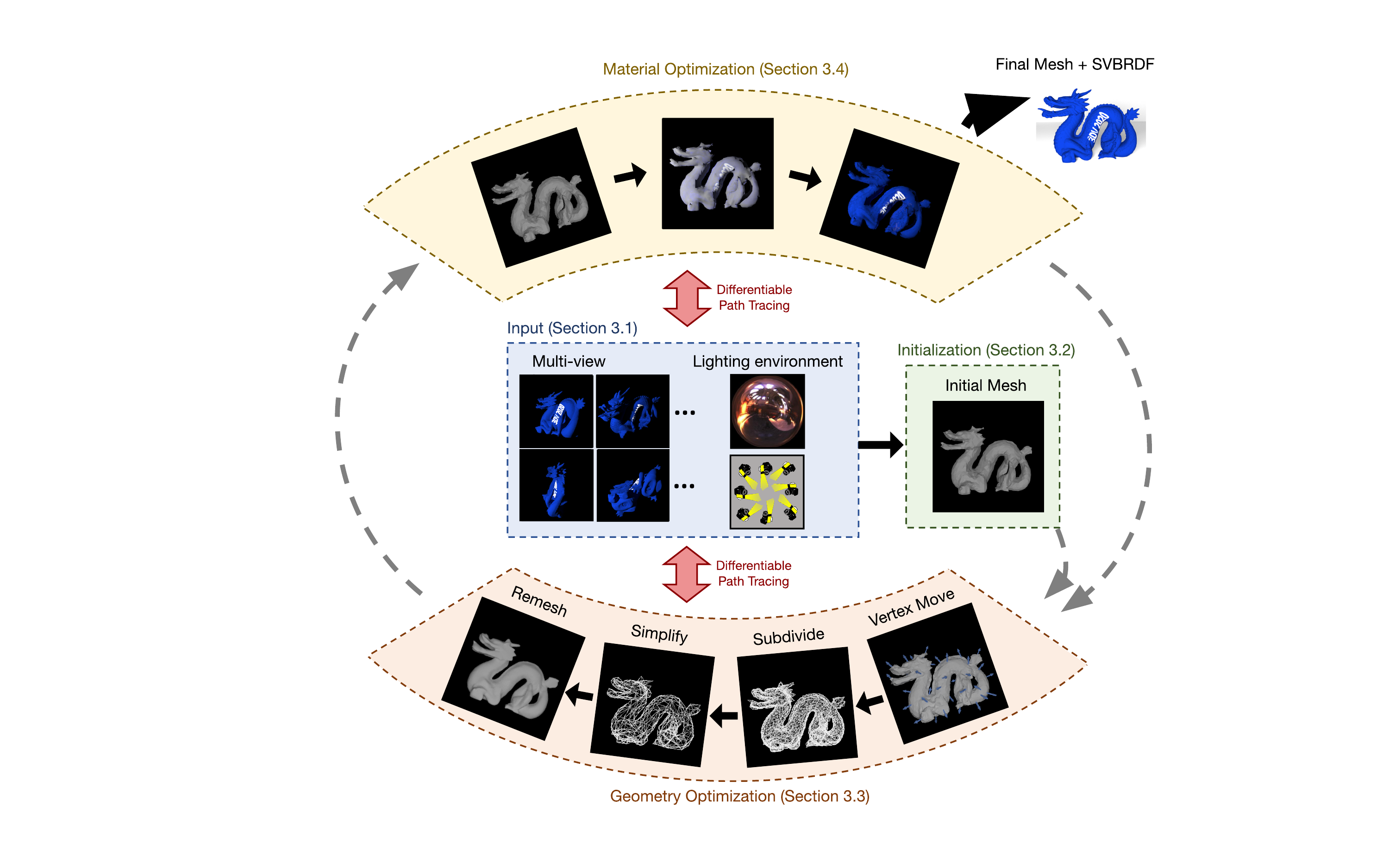}
    \end{minipage}
    \hfill
    \begin{minipage}[c]{0.29\textwidth}
    \caption{
        \revision{Proposed steps for simultaneous geometry} and spatially varying material reconstruction via inverse path tracing.
        Given input views of an object under known lighting, our pipeline begins with a coarse shape initialization, produced through any one of a number of approaches such as space carving or multiview stereo.
        It then alternates between material and geometry optimization, adjusting surface mesh vertices and SVBRDF material texels via stochastic gradient descent with respect to path traced renderings of the current reconstruction.
        This process is made coarse-to-fine by the subdivision of the surface geometry, which then implicitly and automatically subdivides the surface texture via a multi-scale per-facet SVBRDF texture representation.
    }
    \vfill
    \label{fig:pipeline}
  \end{minipage}
  \vspace{-0.3cm}
\end{figure*}


\section{Related Work}
\label{sec:relatedWork}

We focus our discussion on differentiable rendering as applied to inverse problems and on methods which recover shape and spatially-varying non-diffuse material.


\subsection{Differentiable rendering}

With aims to optimize through or ``invert'' the rendering process, the past decade has seen many efforts to develop renderers which are differentiable in output pixels with respect to different input scene properties~\cite{OpenDR,Rhodin_2015_ICCV,SoftRas,DIBR,DIRT}. Modern deep learning toolkits such as Tensorflow and Pytorch3D also now provide differentiable rendering, currently through rasterization~\cite{TensorflowGraphics, Pytorch3D}. These renderers operate on mesh representations of 3D geometry; parallel efforts have also explored differentiable variants of ray marching for rendering implicit surfaces~\cite{SDFDiff,FieldProbing,NeuralVolumes,sitzmann2019scene, mildenhall2020nerf}.
All of the above consider either only geometry, or geometry plus local illumination.
Recently, differentiable formulations of global illumination rendering have been proposed, resulting in physically-based inverse renderers~\cite{redner,Mitsuba2,DiffRadiativeTransfer}. 

Differentiable renderers have been used to fit morphable human face models to images~\cite{FitMorphableModel,Gaurav} and to optimize for more general classes of objects~\cite{PerspectiveTransformerNets,SoftRas,DIBR,Pix2Vex}, to acquire materials~\cite{MaterialsForMasses} and optimize for effects like caustic reflections~\cite{Mitsuba2}, paired with an encoder to predict subsurface scattering parameters~\cite{9105209} and to simultaneously estimate materials and lighting in 3D scenes~\cite{InversePathTracing}.
We show that geometry and material refinement via differentiable physically-based rendering can account for complex light transport effects. This strategy also makes it feasible to reconstruct real-world objects exhibiting reflections, specular highlights, and soft shadows, within unconstrained environments, given calibrated views and an HDR environment map.

\subsection{Geometry and material reconstruction}

Many works reconstruct geometry and material; we refer to Weinmann et al.~\cite{egt.20161032} for a recent review. Geometry methods include multi-view stereo~\cite{MultiViewStereoComparison,DeepMVS} techniques to reconstruct point clouds with diffuse color, space carving~\cite{SpaceCarving} techniques to reconstruct voxel volumes with diffuse color, or photometric stereo techniques to reconstruct surface normals~\cite{wu2011high,langguth2016shading} and spatially-varying specular materials~\cite{Goldman2010}.

Some approaches reconstruct complex material with simplified geometry. Lin et al.~\cite{Lin2019MaterialAcquisition} present a shape‐agnostic method for on-site BRDF capture, and Gao et al.~\cite{gao2019deep} use data-driven methods to reconstruct SVBRDFs under planar assumptions. Other methods implicitly perform reconstruction via view synthesis. Xu et al.~\cite{xu2019deep} use data-driven photometric stereo to generate new views from sparse views, which then drive reconstruction via multi-view stereo~\cite{schoenberger2016sfm}. Li et al.~\cite{li2018learning} present a learning-based method to reconstruct SVBRDF and geometry from a single image.

Other methods reconstruct spatially-varying BRDFs with specular components and whole-object geometry. Tunwattanapong et al.~\cite{Tunwattanapong2013} use a dense lighting capture setup and turntable to simulate varying spherical harmonic environment maps. Xia et al.~\cite{RSR-UI} reconstruct geometry and SVBRDF under unknown illumination from coarse initializations by using temporal traces of the reflected illumination as the object rotates over time, though it cannot handle interreflections or occlusions. Kang et al.~\cite{Kang2019IllumMultiplexing} use a controlled light box with matching synthetic training data to learn detailed geometry and SVBRDF reconstruction, though it cannot handle interreflection and self-shadowing. Most flexibly, Nam et al.~\cite{Nam2018PracticalAcquisition} present a practical smartphone-based geometry and SVBRDF capture system which uses interactive inverse-rendering, although the system is constrained to blacked-out room with point illumination. None of these approaches explicitly model global illumination effects like interreflection and self-shadowing.

Some methods explicitly model interreflection. Lombardi and Nishino~\cite{Lombardi_2016} model multiple bounces of light through path tracing and compute derivatives with respect to reflectance and illumination. Geometry adjustment is modeled from an initial depth fusion through a linear combination of surface normals, which can inflate or deflate the surface. Park et al.~\cite{park2020seeing} 
model interreflection and Fresnel reflectance in their learning-based recovery of scene properties from RGBD imagery, via surface light field and specular reflectance map reconstructions. Both approaches assume accurate geometry initialization, while we include results on reconstructions from coarser initializations. Overall, the problem of simultaneous geometry and SVBRDF capture under global illumination effects is still difficult.



\section{Method}
\label{sec:method}

Figure~\ref{fig:pipeline} shows our exploratory reconstruction pipeline based on differentiable path tracing.
Starting with a set of images captured from known viewpoints and under known illumination, we propose a procedure which first constructs an initial coarse estimate of object geometry using existing methods, and then alternates between optimizing this geometry and a mesh colors~\cite{Yuksel:2010:MC:1731047.1731053} spatially-varying material using gradient descent with a differentiable path tracer.
Our proposed procedure has multiple sub-components; the remainder of this section motivates and describes each in detail.

\subsection{Input}
\label{sec:input}

\begin{figure}[t!]
  \centering
  \setlength{\tabcolsep}{1pt}
    \begin{tabular}{ccc}
        32 views & 16 views & 8 views
        \\
        \includegraphics[trim=0 0 0 160,clip,width=.32\linewidth]{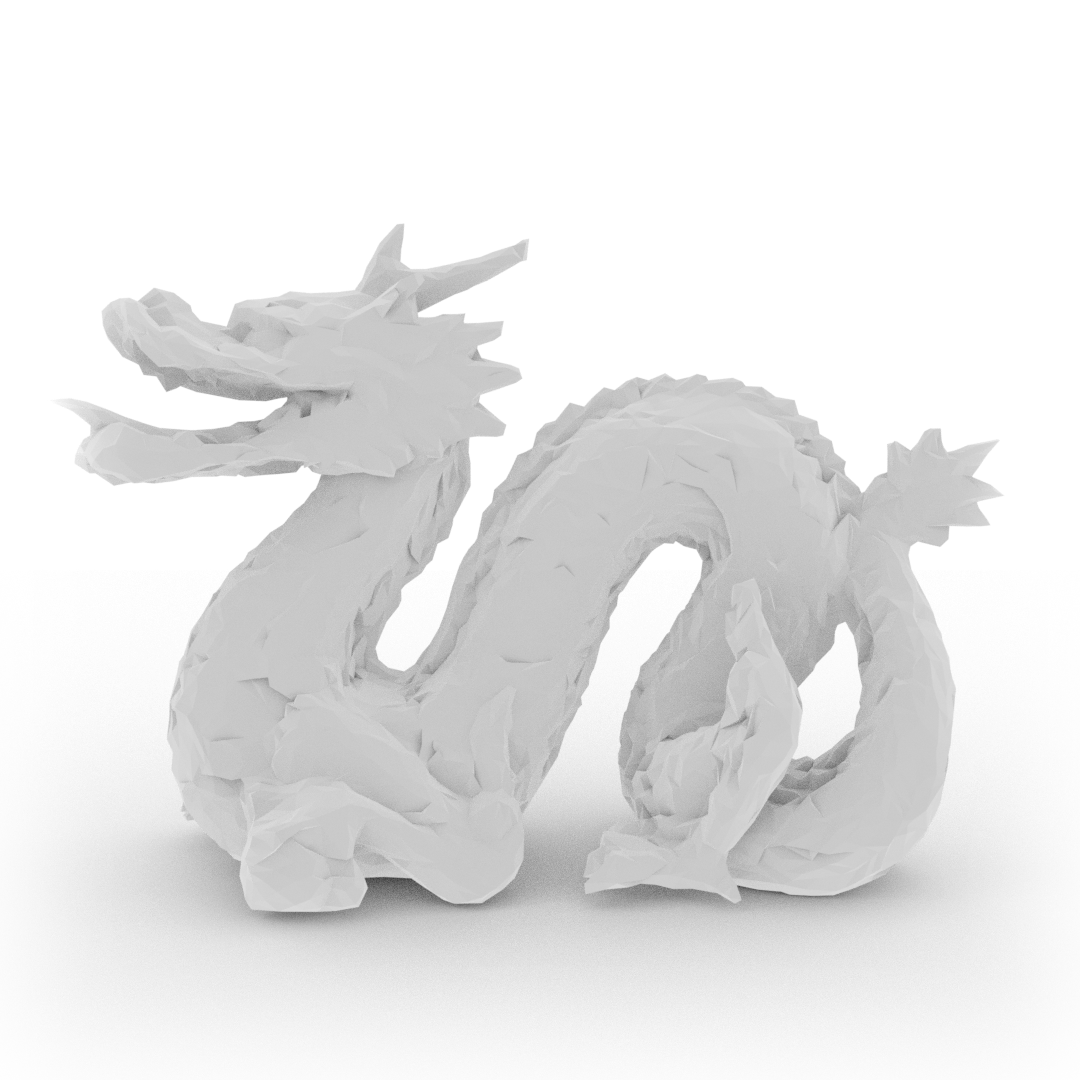} &
        \includegraphics[trim=0 0 0 160,clip,width=.32\linewidth]{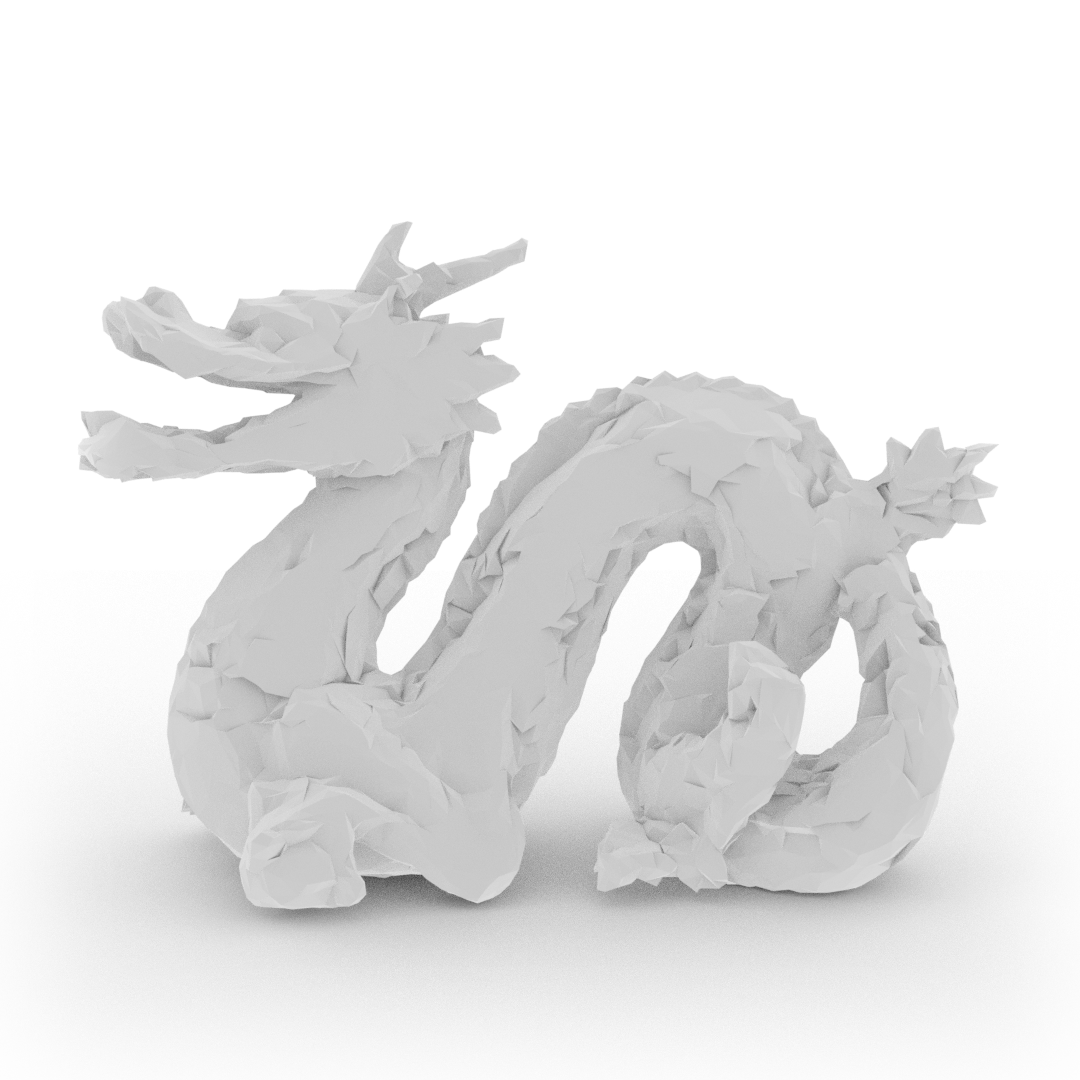} &
        \includegraphics[trim=0 0 0 160,clip,width=.32\linewidth]{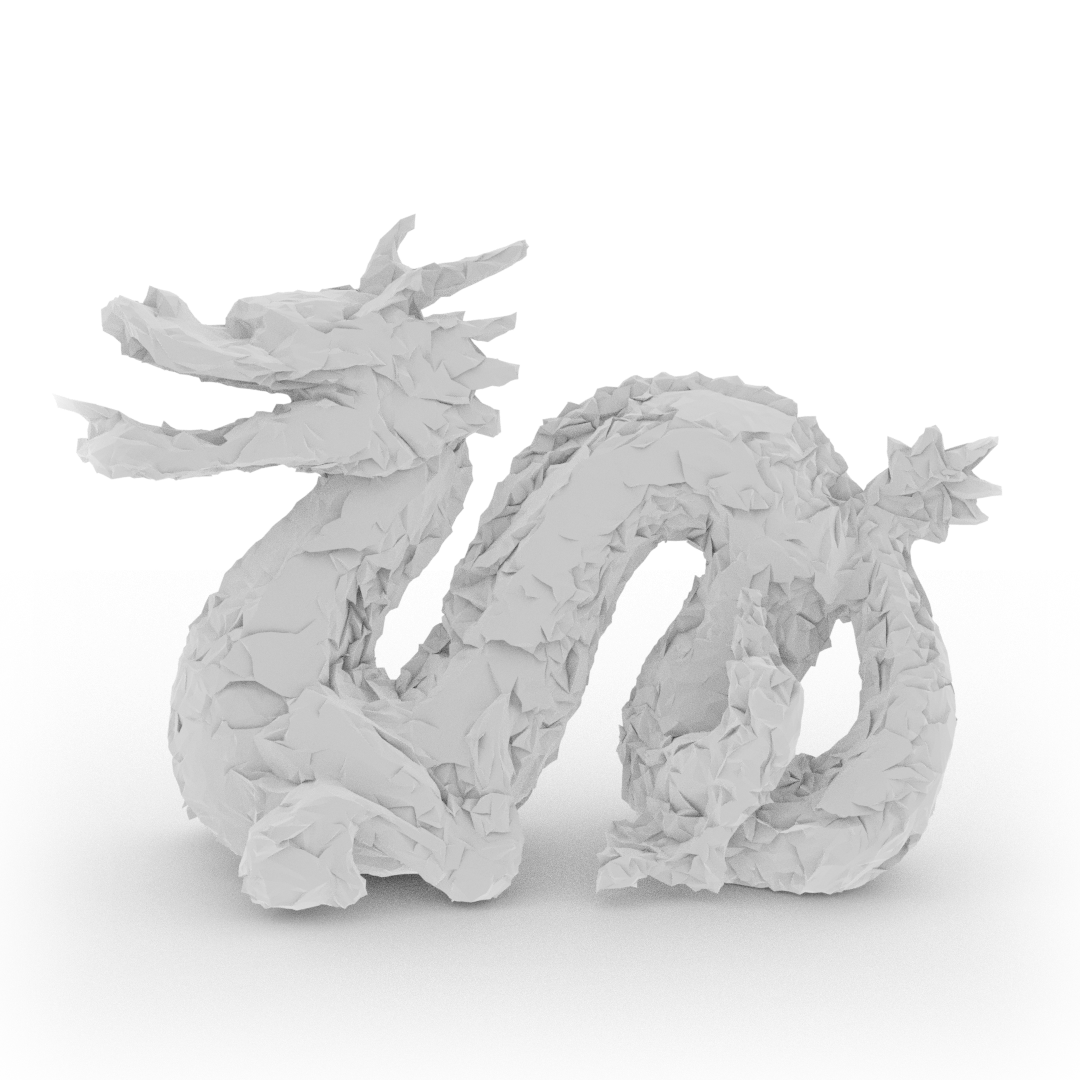}
        \\
        F1: \textbf{87.51} & F1: 78.40 & F1: 62.31
    \end{tabular}
    \vspace{0.2cm}
\caption{
Degradation of reconstruction quality with decreasing number of input views. As the number of views to constrain the optimization decreases, more noise appears in the geometry reconstruction.
F1 score is computed with a tolerance of 0.01.
}
\label{fig:num_views}
\end{figure}

The input to our pipeline is a set of images of the target object plus the scene lighting, represented either by an HDR environment map or a set of point lights. 
Images are captured from known poses and under a known lighting configuration as might be captured by a light stage~\cite{Tunwattanapong2013} or box~\cite{Kang2019IllumMultiplexing}, from a multi-view stereo setup with known camera/light offset~\cite{Nam2018PracticalAcquisition}, or as frames from a low-dynamic-range video sequence from a hand-held cell phone with known environment lighting. Typically, the greater the number of views or frames, the higher the quality of reconstruction. Figure~\ref{fig:num_views} shows the effects of number of input views on reconstruction quality.


For scene lighting input, we use point lights to reconstruct objects in simulation, and environment maps to reconstruct objects in real-world scenes.
In the supplemental material, we investigate the effect of illumination model (environment vs. point lights) on reconstruction quality.
We find that reconstruction is more accurate under point illumination than environment map illumination. However, environment maps better model real-world scenes. 

\revision{We  only consider input image pixels covering the object and mask away the image background.}
Masks can be created manually or via classic or machine-learned image segmentation.
In simulation, we compute binary foreground masks via a white albedo render of the relevant scene geometry.
For real-world imagery, we use the \emph{X101-FPN} model from \emph{detectron2}, pretrained on the COCO dataset, to perform instance segmentation~\cite{wu2019detectron2}.

Similarly, image-space masks for separate materials that appear in the same object  (e.g., distinguishing a plastic bottle from a metal bottle cap) are helpful for controlling the material optimization landscape. These can be computed in a variety of ways: manually, for finest precision, or clustering image pixels by their RGB color or normalized intensity.





\subsection{Geometry Optimization}
\label{sec:geoopt}

Given the above inputs, our approach is to alternate between optimizing the geometry and material of the reconstruction.
\revision{We detail the geometry phase of this alternating minimization scheme: the representation, initialization, and a multi-step approach to perform gradient-based refinement on geometry while controlling its resolution and quality.}

\paragraph{Representation}
We use a triangle mesh to represent object geometry.
First, it provides local control over geometry, allowing for optimization to locally capture fine detail.
Second, it supports coarse-to-fine refinement, which our optimization schedule heavily exploits.
Third, it naturally accommodates SVBRDF specification via a per-mesh-facet representation.
Finally, it facilitates highly-optimized ray-surface intersection, which forms the bulk of path tracing's computational cost.
The major limitation of a mesh, as opposed to an implicit representation, is that it is more difficult to change topology during optimization.
\revision{As we will see later in this section, 
periodic remeshing during optimization can overcome this difficulty.}

\paragraph{Initialization}
Since our optimization is based on gradient-based local optimization, it is important to start with an initial mesh that captures large-scale topological features to place the optimizer in the right basin of attraction.
Possible initialization strategies are to start with a simple proxy geometry, e.g., spheres or boxes, which can be used in any setting but may require significant hand-tuning to work well; or to leveraging existing bottom-up reconstruction methods---these give more accurate initial results but may make assumptions about the underlying scene.
We have experimented with the multi-view stereo pipeline COLMAP \cite{schoenberger2016mvs}, voxel carving, and `sphere clouds,' which we detail in the supplemental material.

Figure~\ref{fig:geo_init} illustrates the behavior of different initalization strategies when refining them with our suggested procedure.
MVS produces the highest quality initializations and therefore the best reconstruction, but requires a large number of input views ($>$100).
Voxel carving operates more reliably under a range of camera views (as few as six) at the cost of some geometric detail.
The sphere cloud approach, while general-purpose, is least accurate.
In the supplemental material, we also explore refining initial geometry produced by a deep-learning based reconstruction method.

\begin{figure}[t!]
  \centering
  \setlength{\tabcolsep}{1pt}
    \begin{tabular}{rccc}
        & MVS & Space Carving & Sphere Cloud
        \\
        \raisebox{1em}{\rotatebox{90}{Initial}} &
        \includegraphics[width=.32\linewidth]{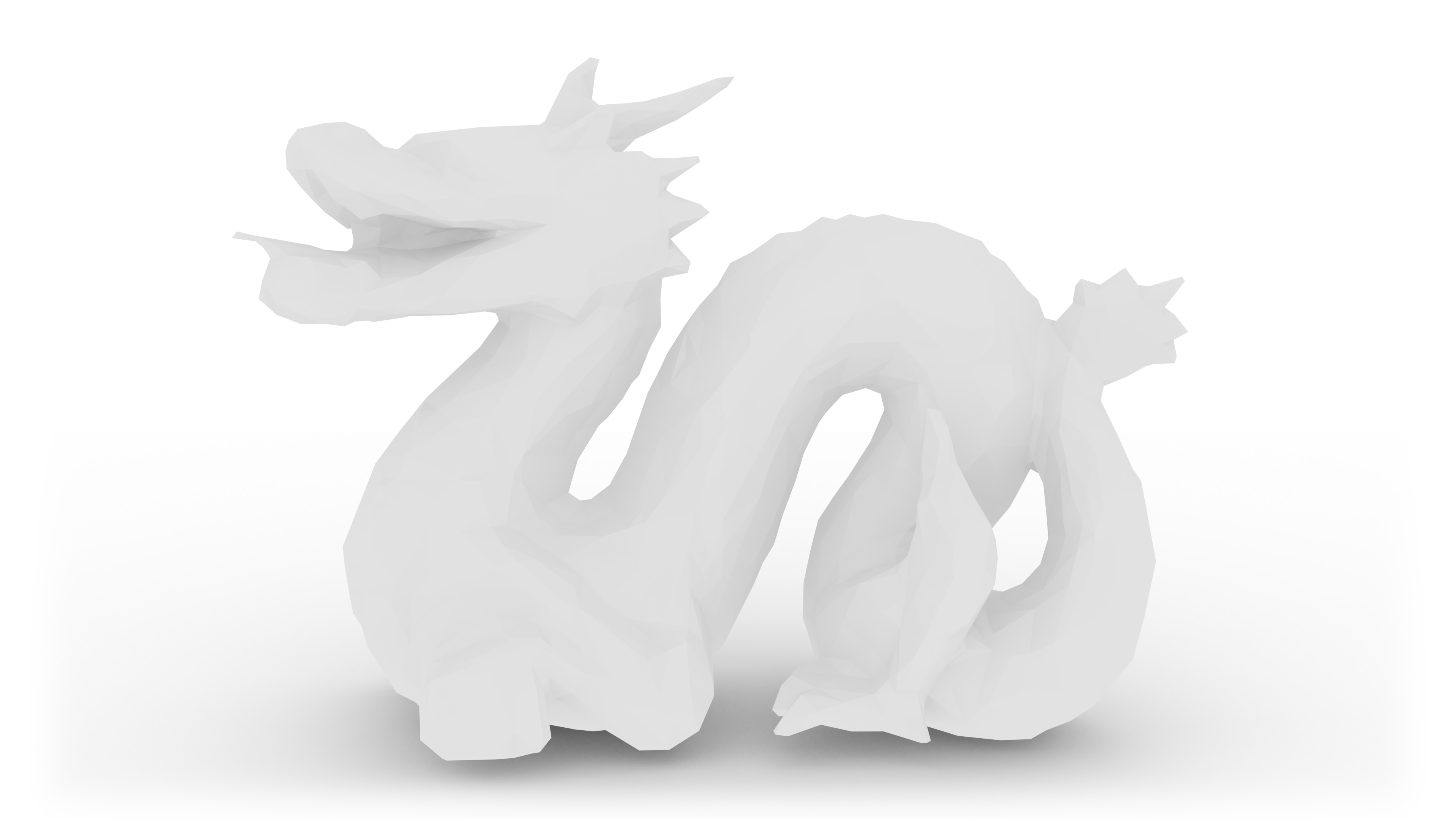} &
        \includegraphics[width=.32\linewidth]{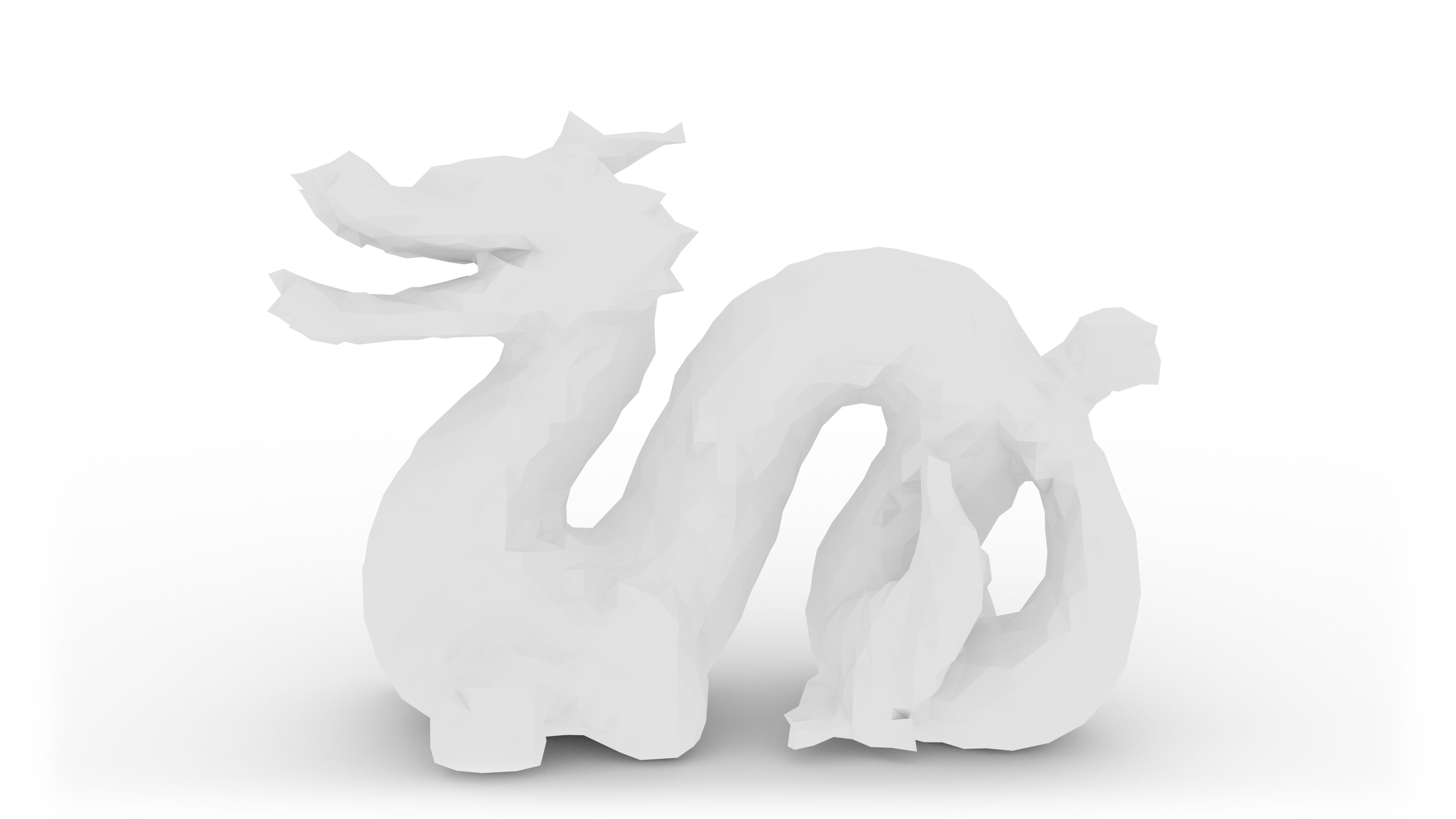} &
        \includegraphics[width=.32\linewidth]{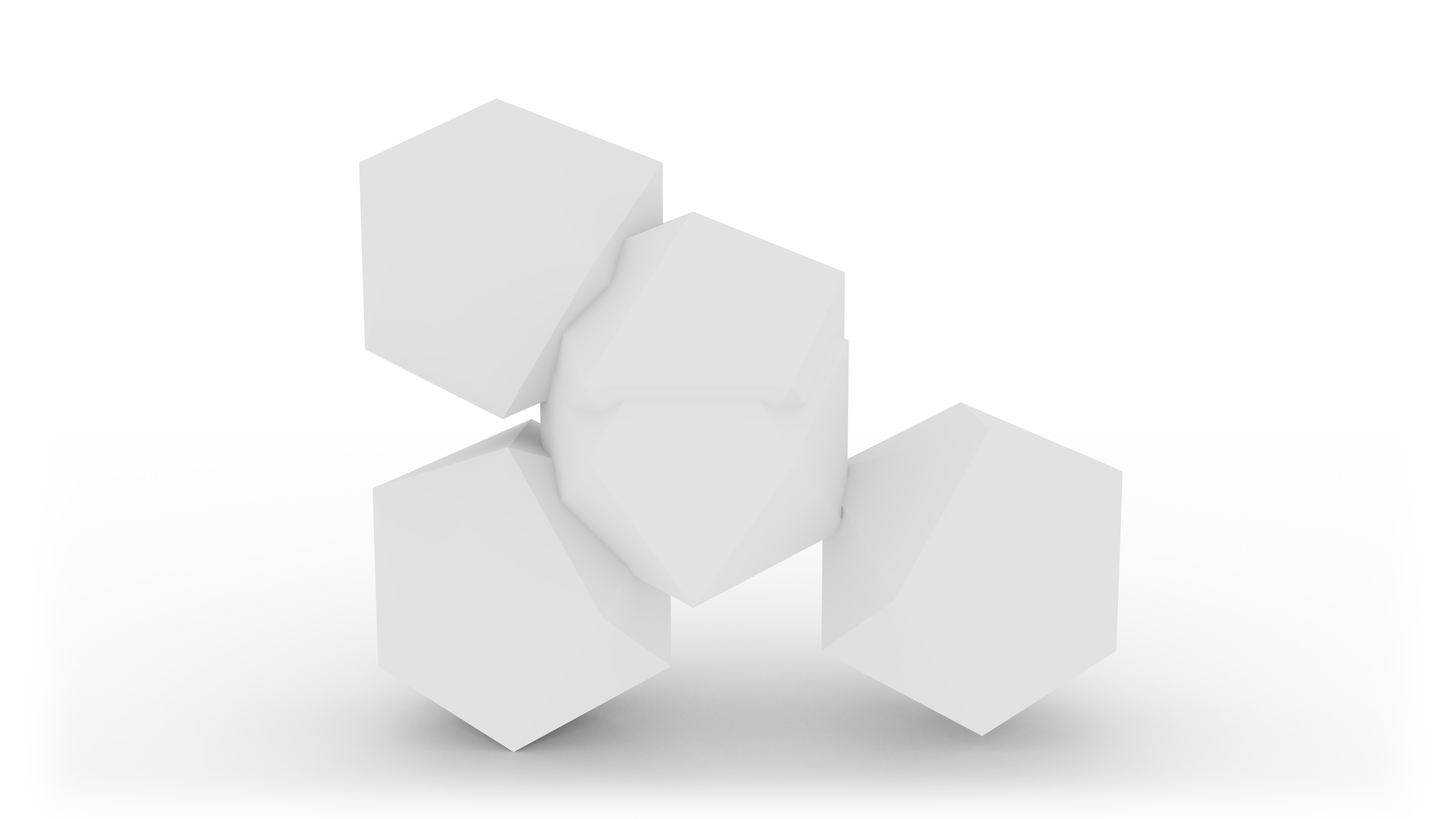}
        \\
        \raisebox{1em}{\rotatebox{90}{Final}} &
        \includegraphics[width=.32\linewidth]{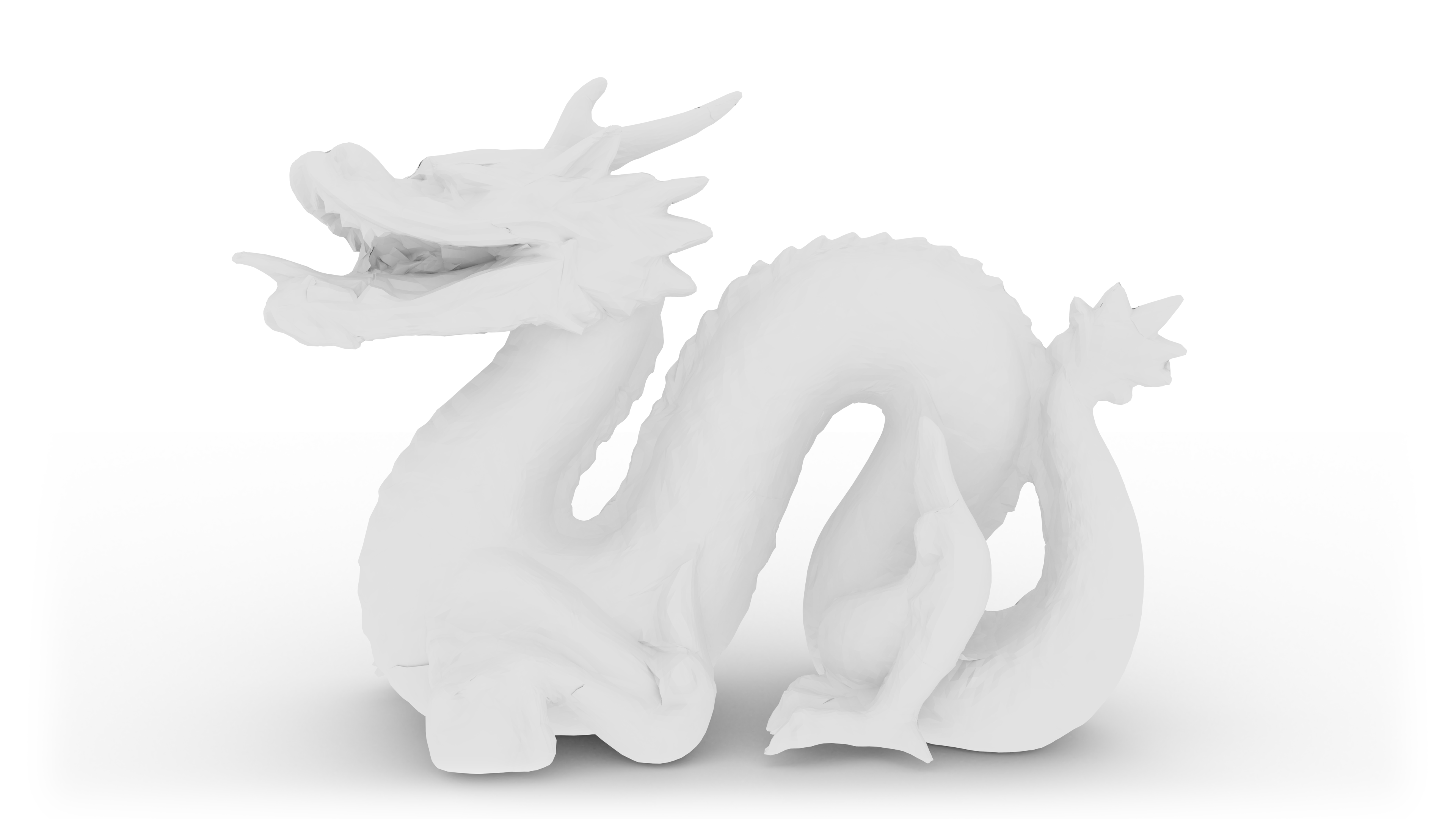} &
        \includegraphics[width=.32\linewidth]{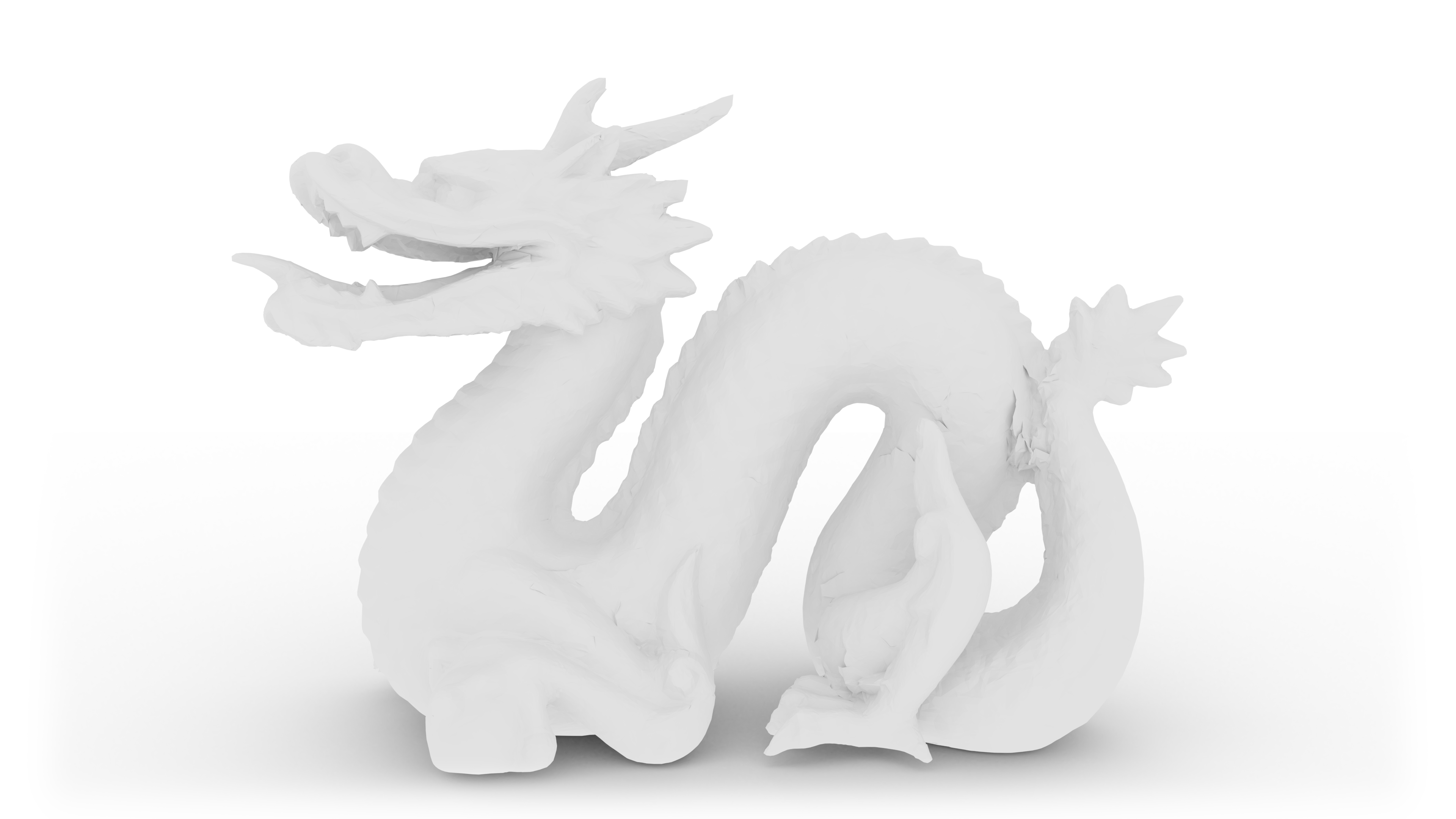} &
        \includegraphics[width=.32\linewidth]{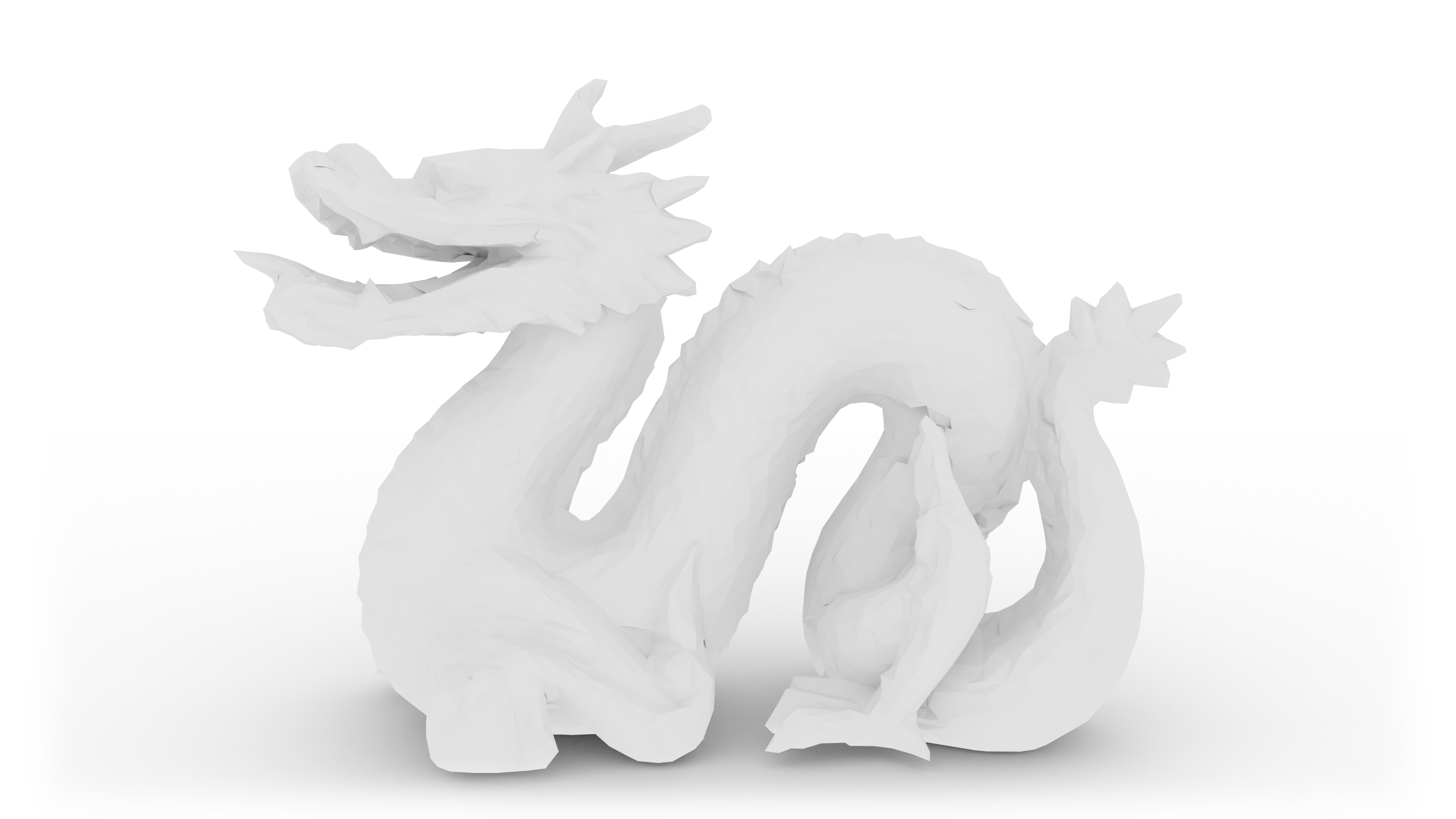}
        \\
        \raisebox{1em}{\rotatebox{90}{Target}} &
        \includegraphics[width=.32\linewidth]{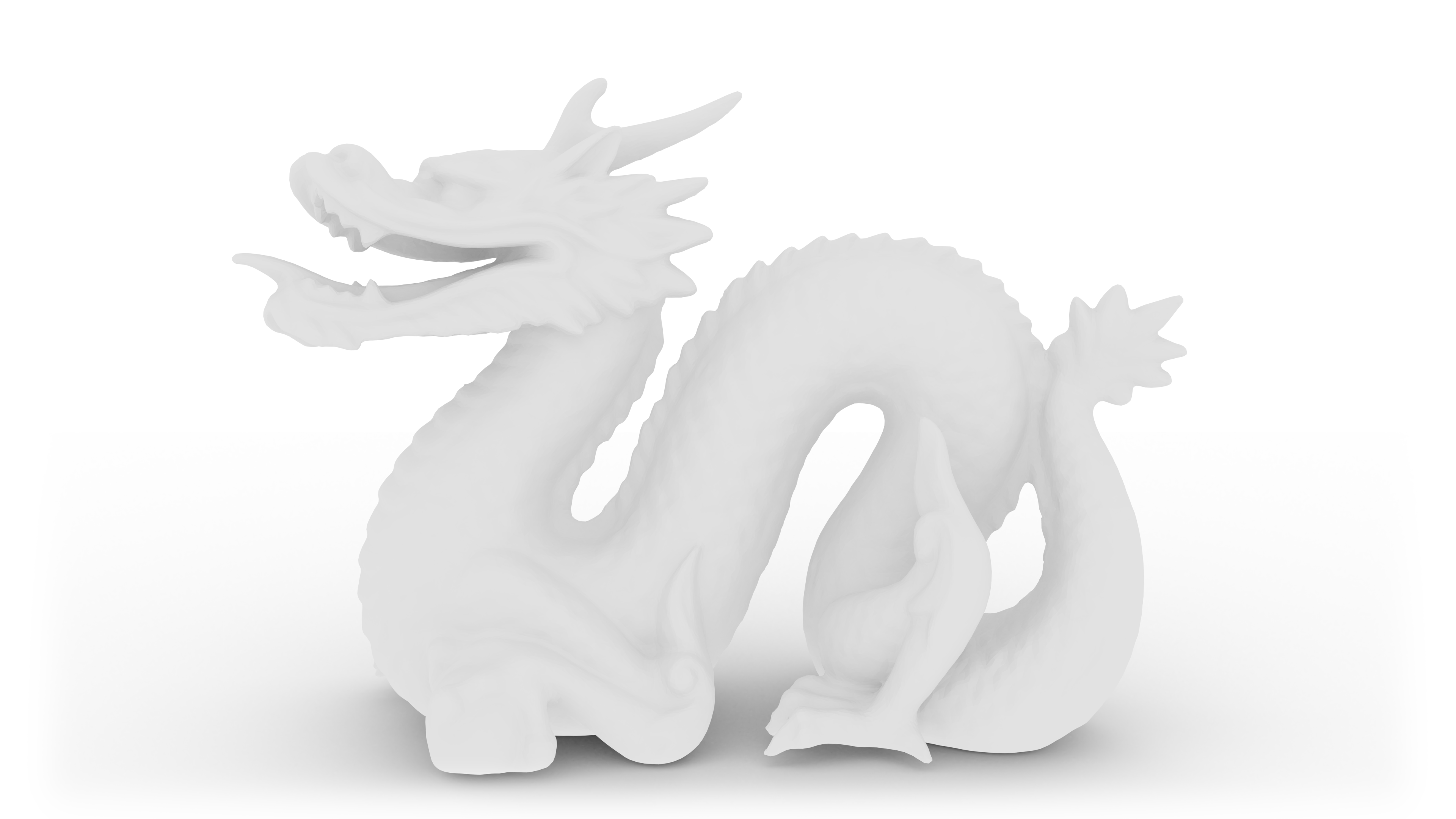} &
        \includegraphics[width=.32\linewidth]{figs/dragon-target.png} &
        \includegraphics[width=.32\linewidth]{figs/dragon-target.png}
        \\
        \raisebox{1em}{\rotatebox{90}{Diff}} &
        \includegraphics[width=.32\linewidth]{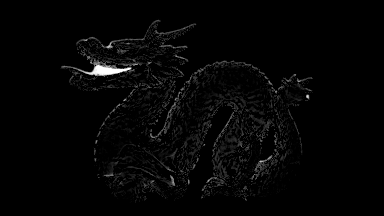} &
        \includegraphics[width=.32\linewidth]{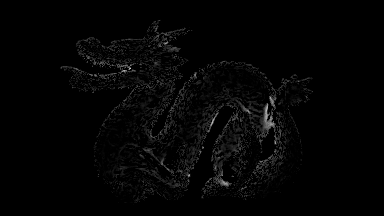} &
        \includegraphics[width=.32\linewidth]{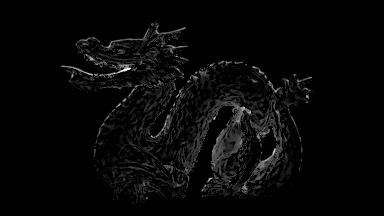}
    \end{tabular}
\caption{
Reconstruction results on different initializations. Top row: initializations produced by multi-view stereo \emph{(Top Left)}, voxel carving \emph{(Top Middle)}, and `sphere clouds' \emph{(Top Right)}. Second row: results of geometry-only optimization starting from each initialization. Bottom row: difference (2$\times$ magnified) between the final reconstruction and the target input image.
}
\label{fig:geo_init}
\end{figure}

\paragraph{Vertex optimization}
Given the initial mesh geometry, the first (and core) step of the geometry phase is to optimize mesh vertex positions $\mathbf{x}$ via gradient descent with respect to a mean-squared-error loss:
\begin{equation*}
    \mathcal{L}(\mathbf{x}) = \frac{1}{n}\sum_{i=1}^{n}(\mathcal{M}_i^t \cdot \mathcal{T}_i - \mathcal{M}_i^r \cdot R(\mathbf{x}, c_i) )^2
\end{equation*}
where $\mathcal{T}_i$ is the $i$th target image, $c_i$ is the $i$th camera pose, $\mathcal{M}_i^t$ is the $i$th target mask, $\mathcal{M}_i^r$ is the $i$th mask of the current reconstructed object, and $R$ is a differentiable physically-based rendering function.
We use the differentiable path tracer of Li et al.~\cite{redner}, as it provides gradients of output pixels with respect to input geometry.



\paragraph{Subdivision}
To avoid poor local optima, we found that coarse-to-fine optimization works well.
We begin with a low-resolution initial mesh, optimize its vertices, and then increase mesh resolution once this optimization converges.
To increase resolution, we subdivide every triangular face into four by splitting each edge at the midpoint.
This approach helps balance the granularity of geometric and material detail at each iteration, preventing geometry from compensating for missing texture detail and preventing texture from `baking in' geometric detail. 

\begin{figure}[t]
  \centering
  \setlength{\tabcolsep}{1pt}
    \begin{tabular}{ccc}
        Input Mesh & Surface Samples & Output Mesh
        \\
        \includegraphics[width=.33\linewidth]{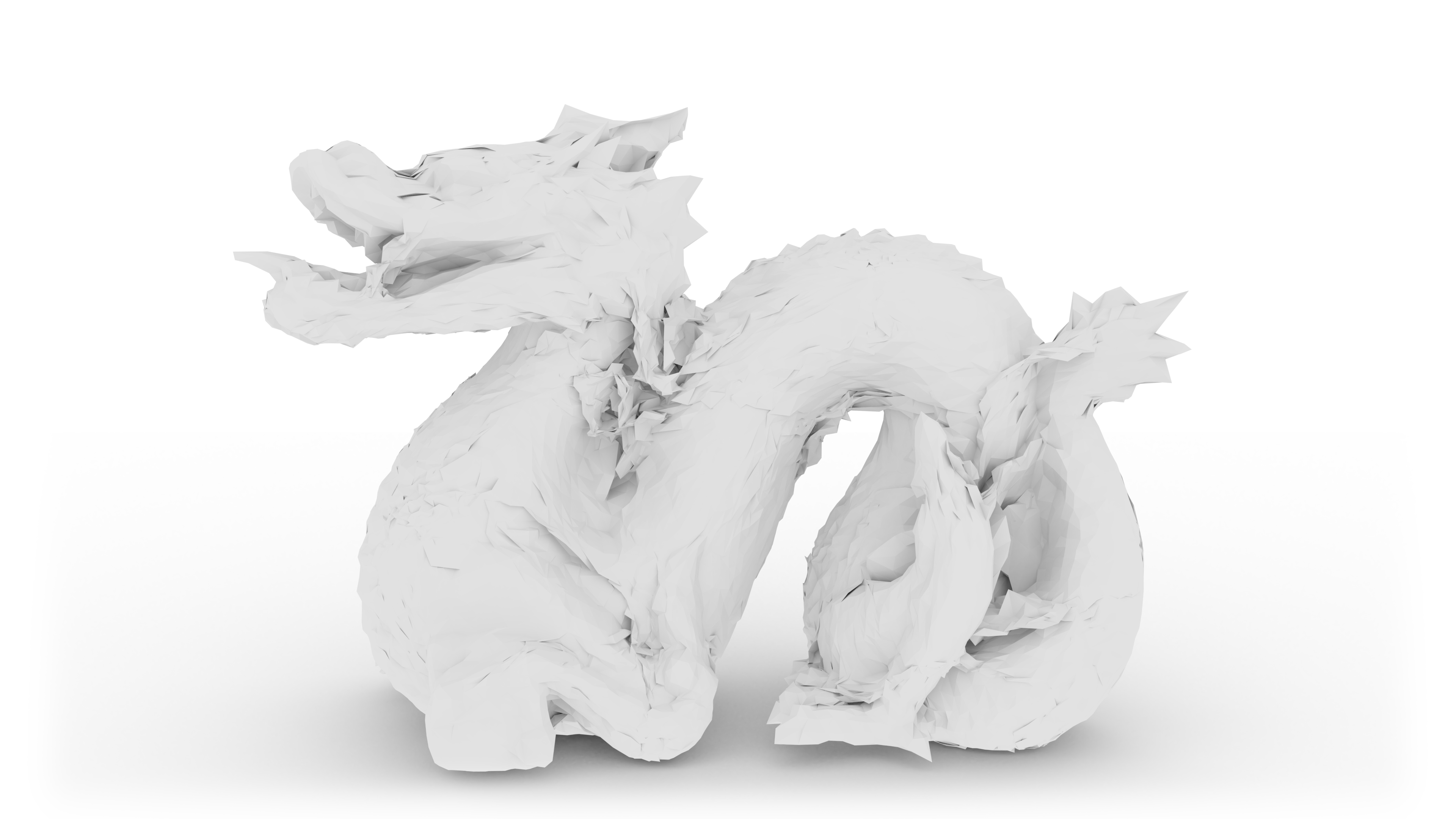} &
        \includegraphics[width=.33\linewidth]{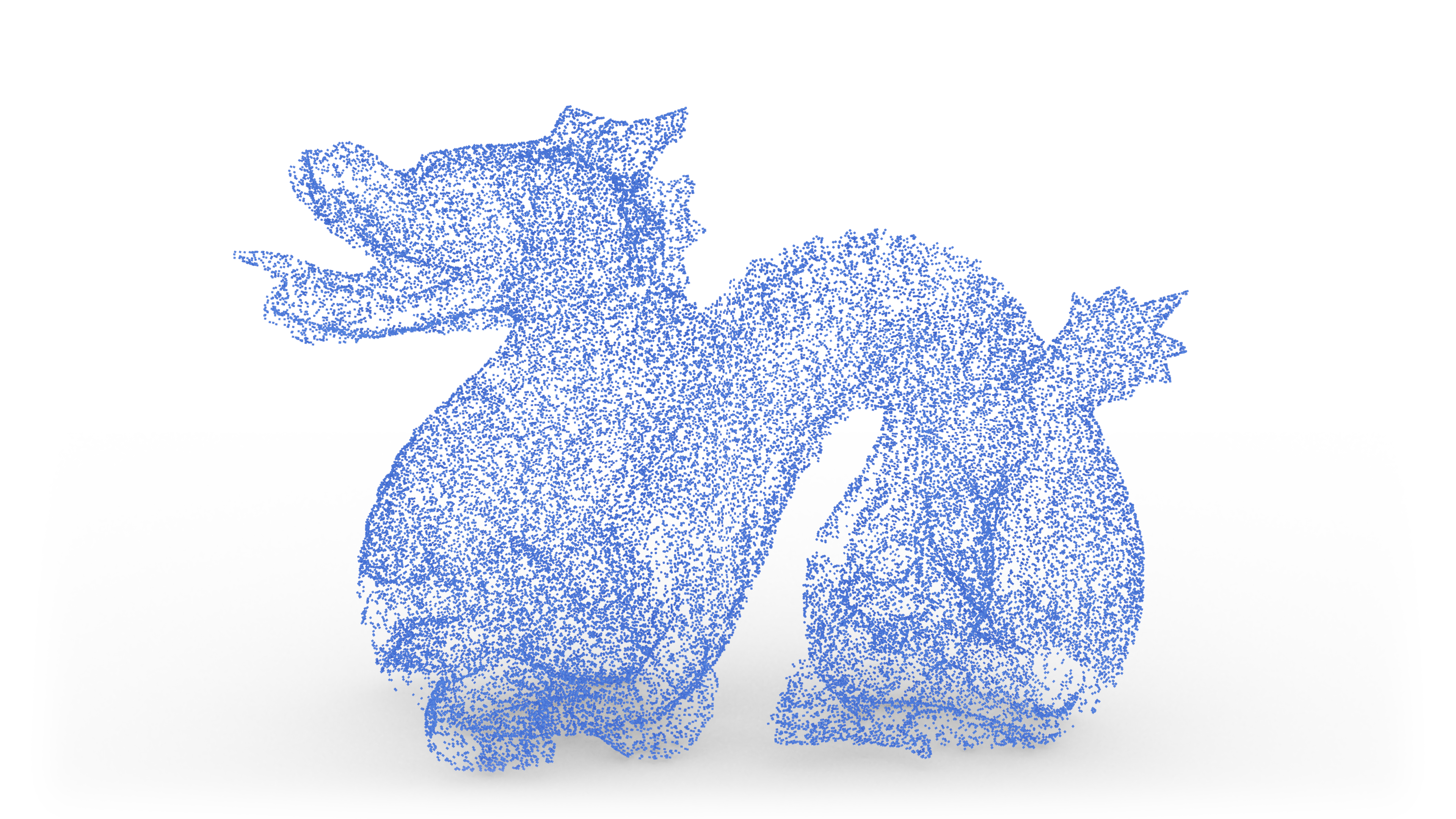} &
        \includegraphics[width=.33\linewidth]{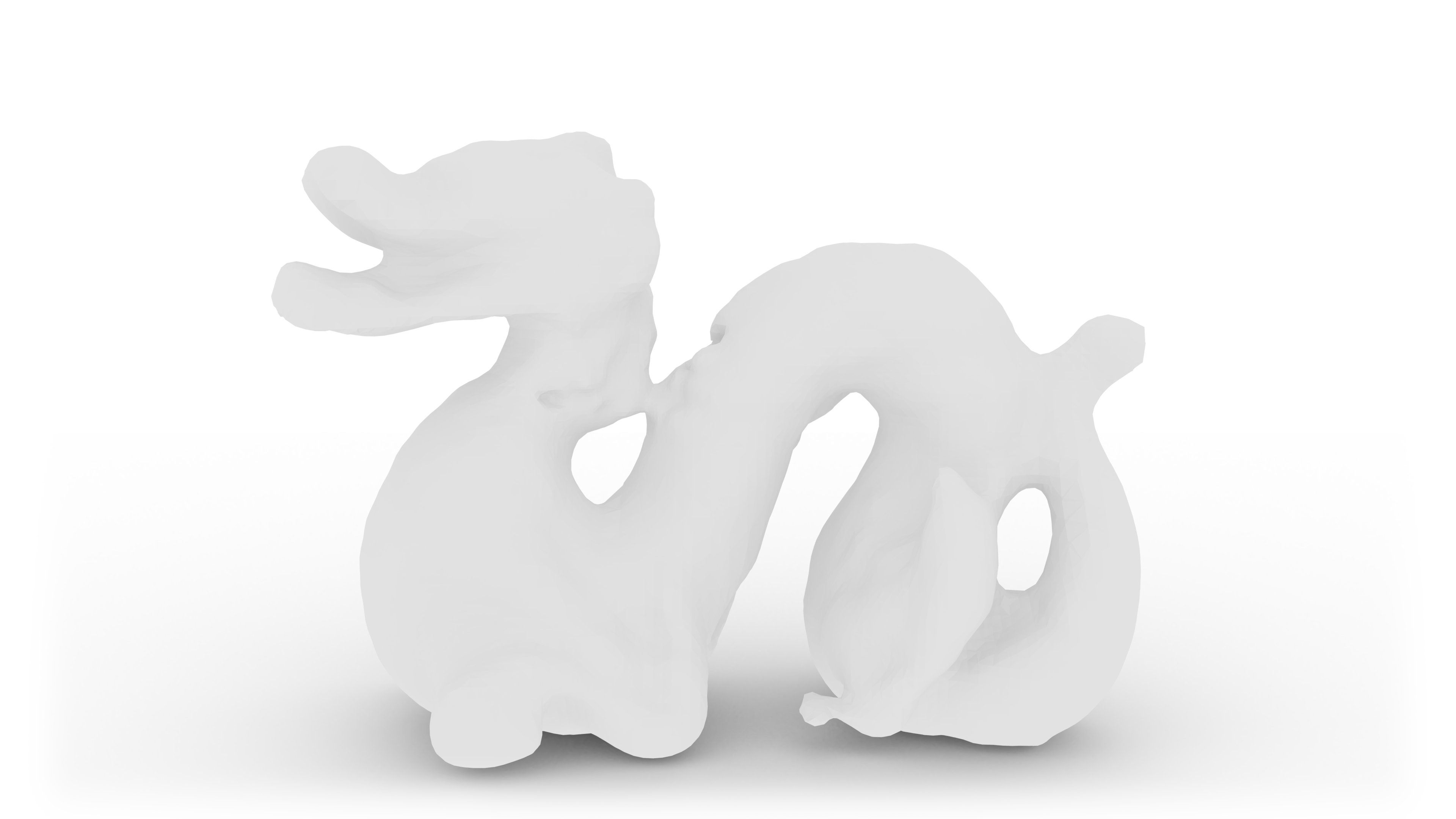}
    \end{tabular}
    \begin{tabular}{cc}
        No Remeshing & With Remeshing
        \\
        \includegraphics[width=.48\linewidth]{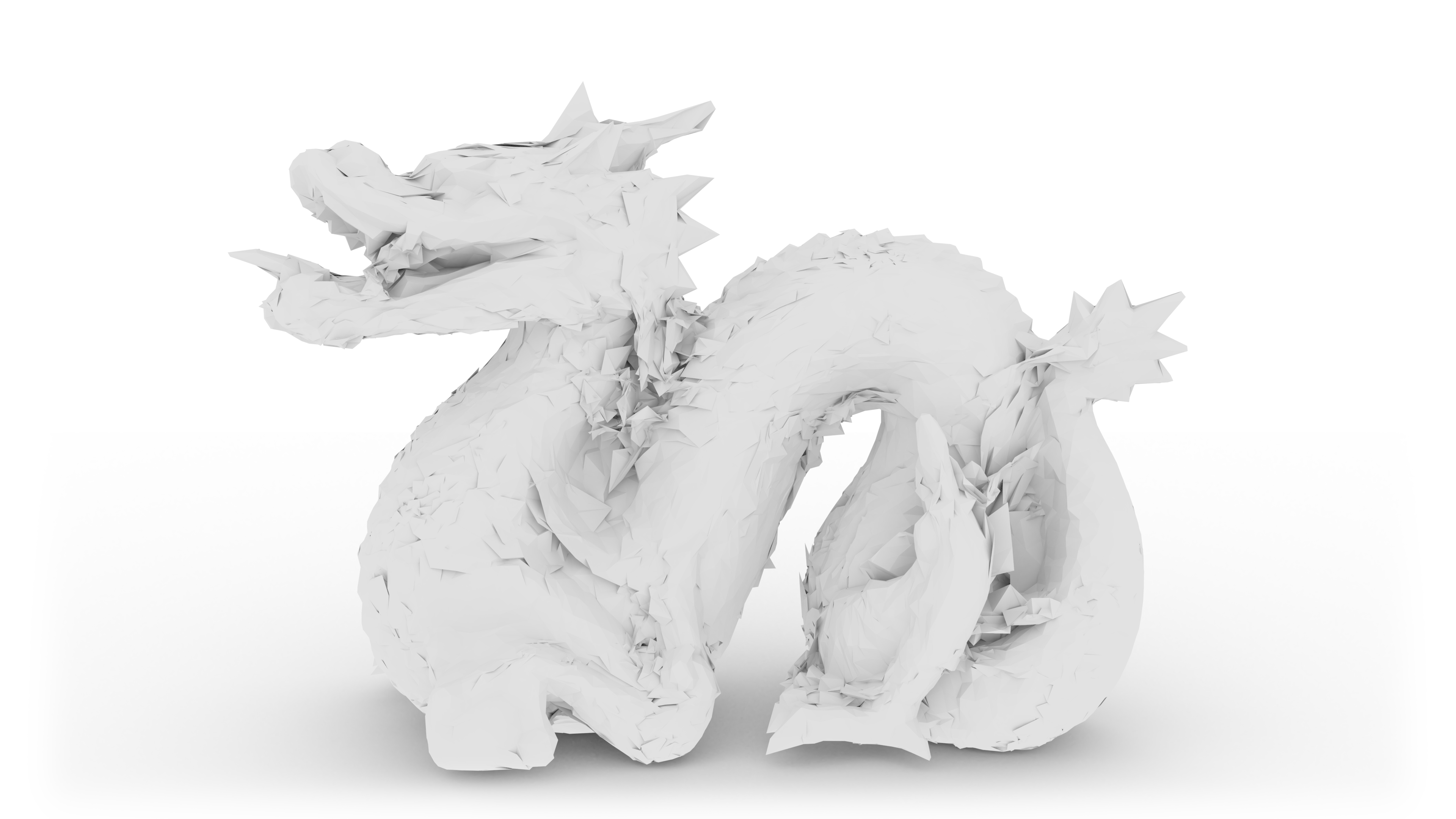} &
        \includegraphics[width=.48\linewidth]{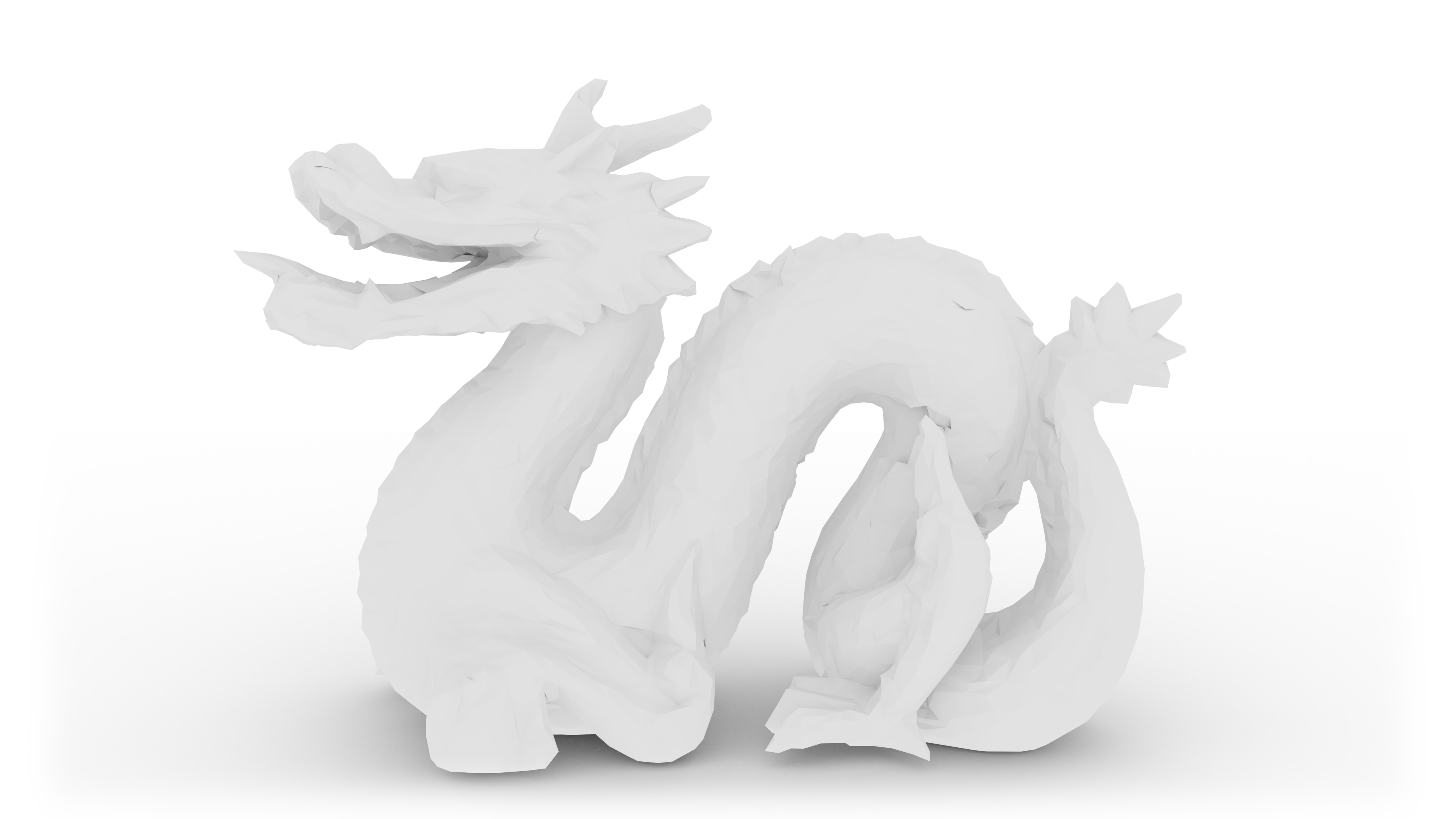}
    \end{tabular}
\caption{
Effect of the remeshing step.
\emph{(Left)} A degraded mesh mid-optimization.
\emph{(Middle)} Point samples on the exterior surface.
\emph{(Right)} Poisson surface reconstruction on points.
Artifacts disappear and fine detail can be recovered by subsequent optimization.
The genus of the object changes from 0 to 1.
\emph{(Bottom)} Final outputs. Self-intersecting geometry degrades the reconstruction and hinders further optimization. Both optimizations are initialized as sphere clouds to emphasize behavior of this step.
}
\label{fig:psr}
\label{fig:no_psr}
\end{figure}

\paragraph{Simplification}
When increasing mesh resolution, we would ideally control the size and stability of the parameter space, as well as the efficiency of optimization, by only adding vertices to mesh regions that require finer detail. Thus, we follow subdivision with mesh simplification.
If, after subdivision, the number of faces exceeds a threshold $\mathcal{D}$, we decimate the mesh using quadric error simplification~\cite{QuadricErrorSimplification} until it has $\mathcal{D}$ faces remaining.
$\mathcal{D}$ is initially set to twice the number of faces, and increases by half the number of initial faces after every simplification step.
While Xu et al.~\cite{xu2019deep} seek to avoid vertices distributed non-uniformly across the surface of the mesh, we find that allowing this behavior facilitates estimation of more complex geometries.

\paragraph{Remeshing}
Due to Monte Carlo rendering noise and the limited number of target images, gradient descent on vertex positions can result in artifacts from which it cannot recover (e.g., excess triangles in the interior of the mesh).
To rectify this problem, we re-generate the mesh by point-sampling its surface via raycasting from the input viewpoints, followed by a Poisson surface reconstruction~\cite{PoissonSurfaceReconstruction} on the resulting point cloud (Figure~\ref{fig:psr}, bottom).
This strategy also allows our input mesh to be less sensitive to the target object's genus, i.e., the remeshing step can open holes in the optimized mesh to match the target (Figure~\ref{fig:psr}, top).

\begin{figure}[t!]
  \centering
  \setlength{\tabcolsep}{1pt}
    \begin{tabular}{rccc}
        & Ours & Target & Diff
        \\
        \raisebox{3em}{\rotatebox{90}{}} &
        \includegraphics[width=.32\linewidth]{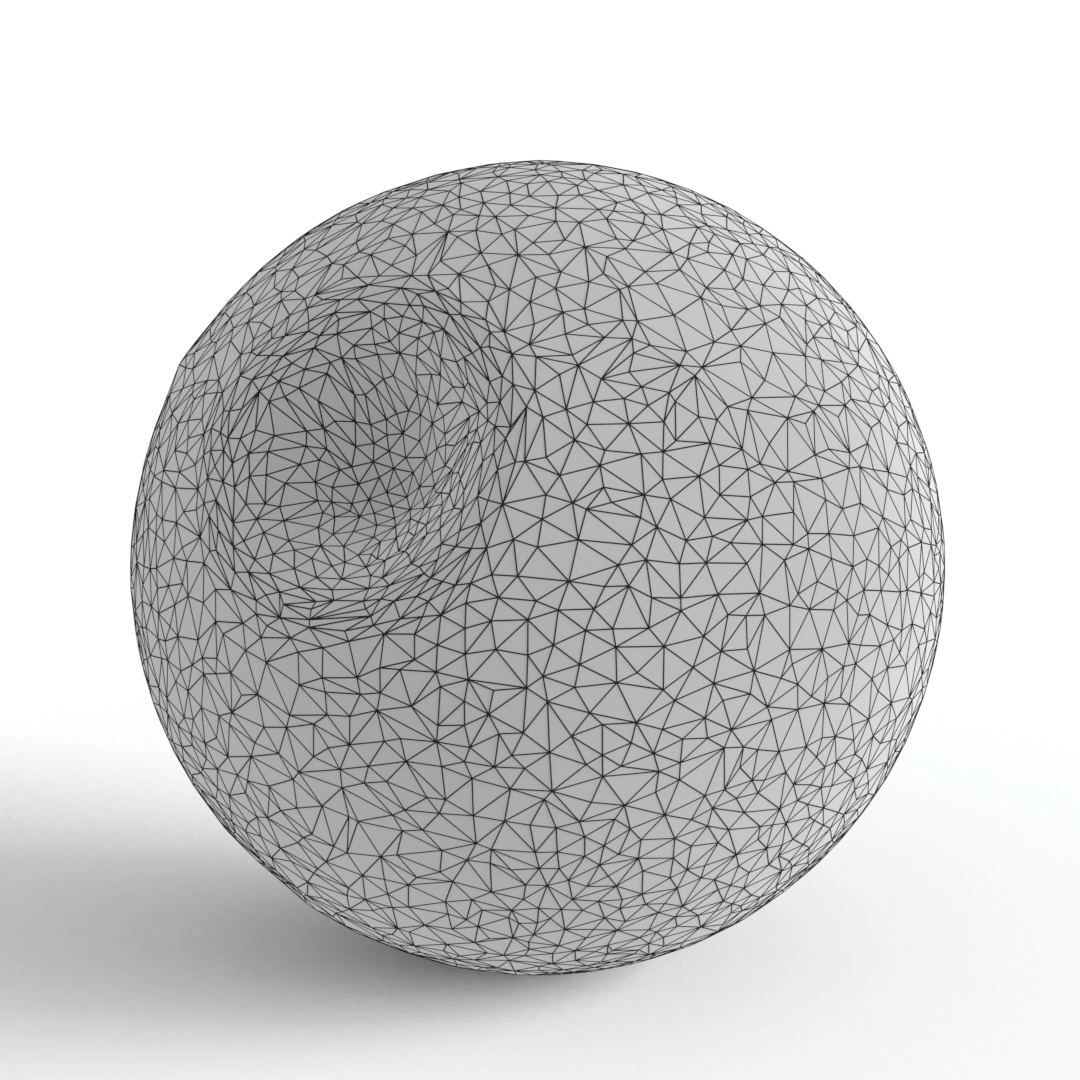} &
        \includegraphics[width=.32\linewidth]{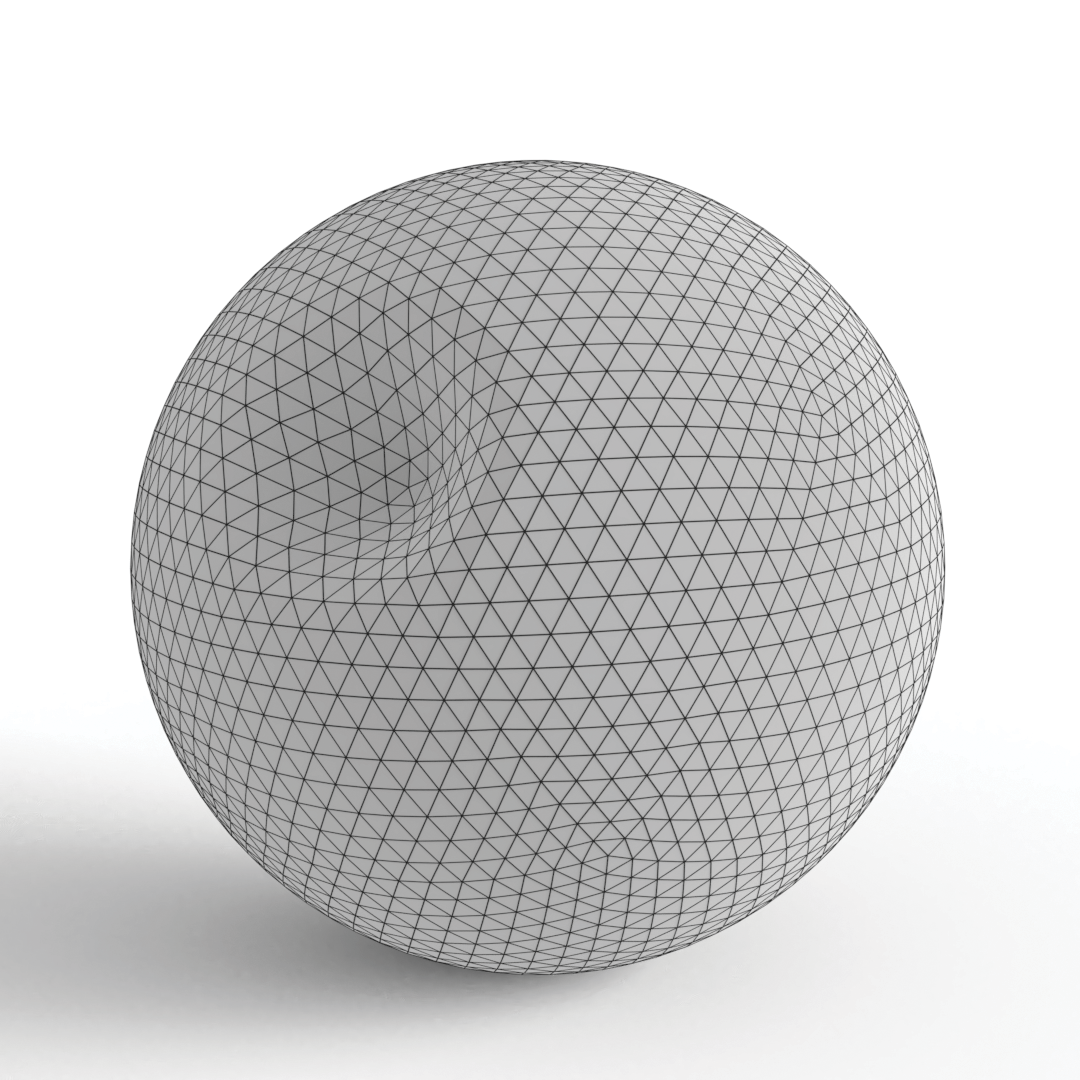} &
        \includegraphics[width=.32\linewidth]{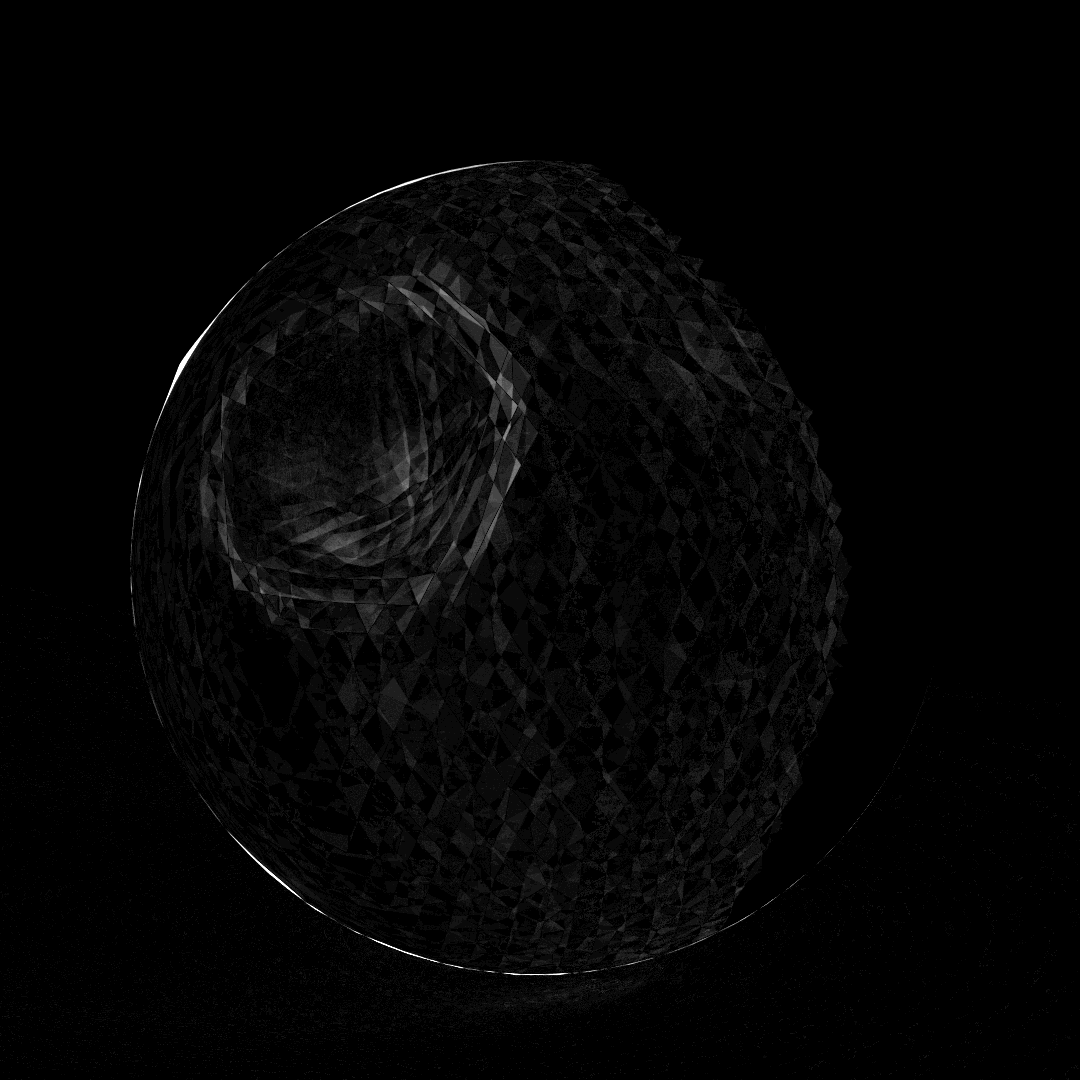}
        \\
        \raisebox{3em}{\rotatebox{90}{}} &
        \includegraphics[width=.32\linewidth]{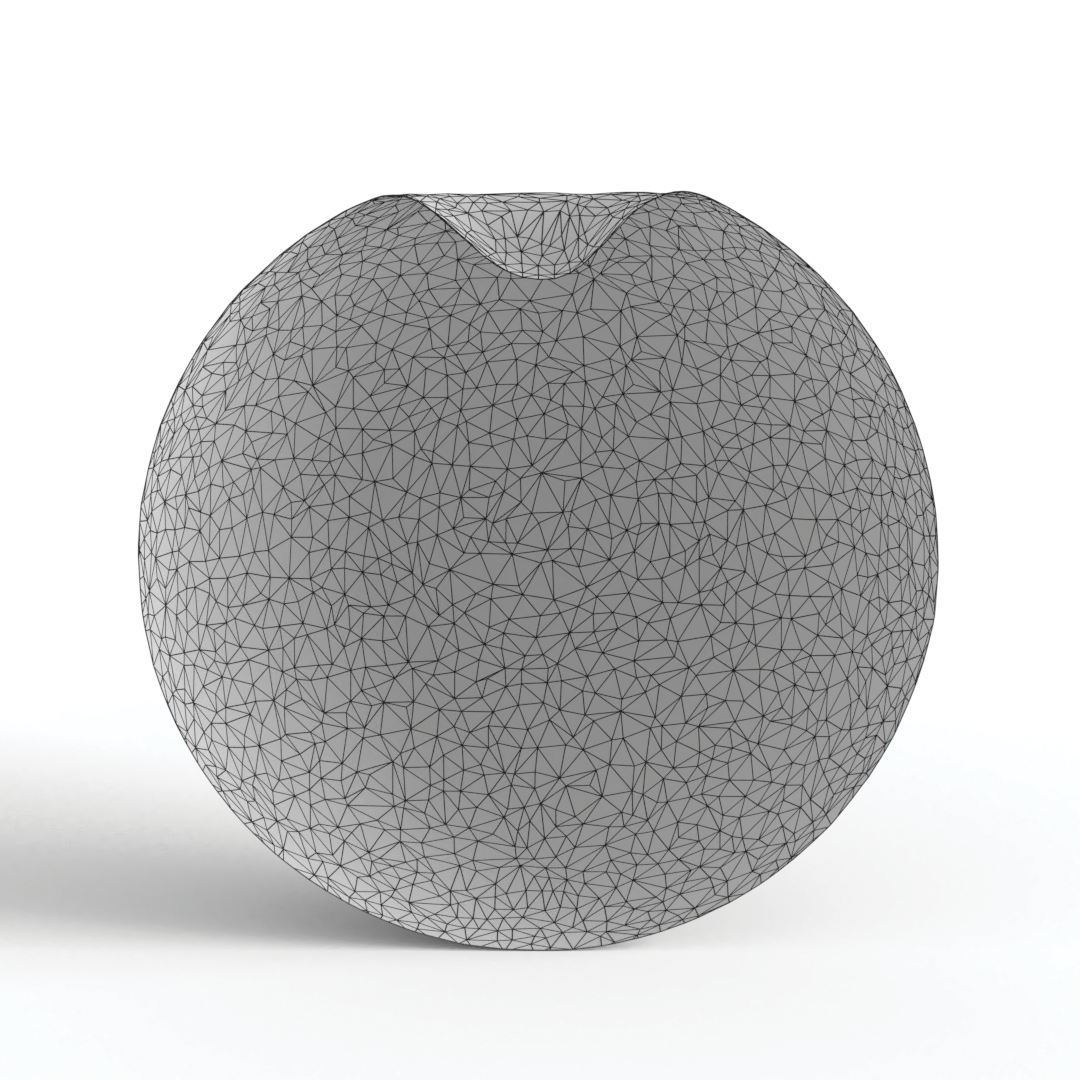} &
        \includegraphics[width=.32\linewidth]{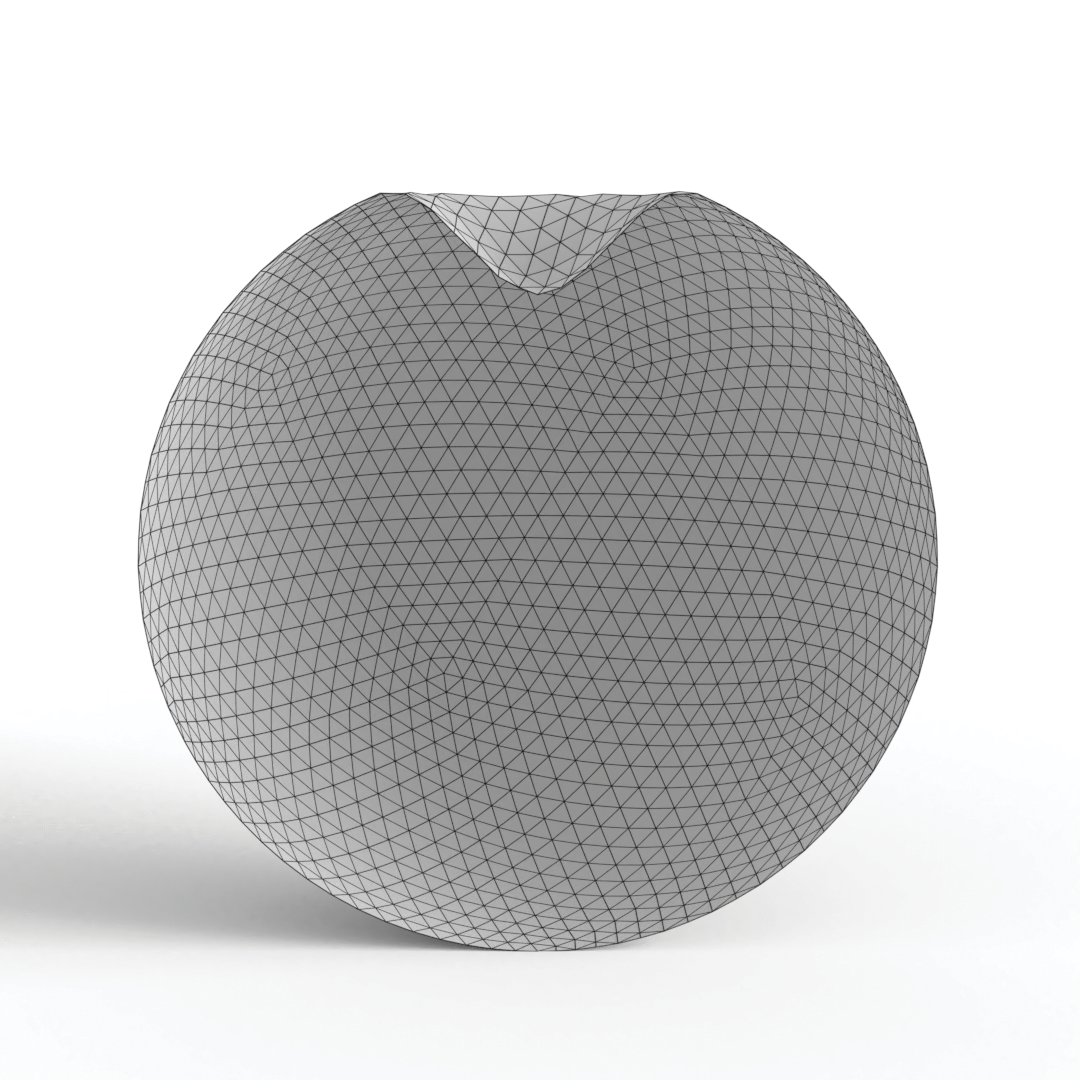} &
        \includegraphics[width=.32\linewidth]{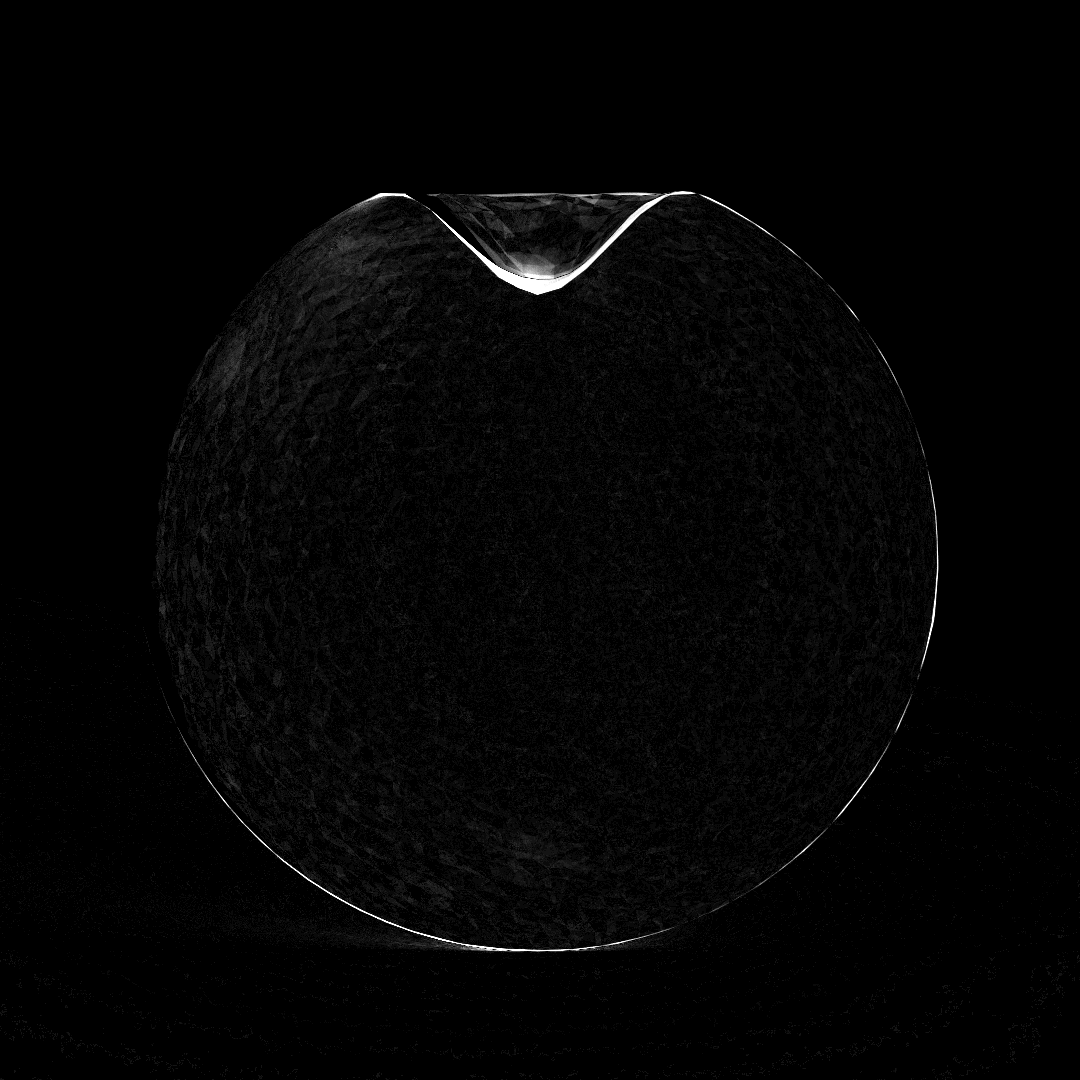}
    \end{tabular}
\caption{
Reconstruction of a sphere-with-divot, starting from a sphere, using 5 cameras each with 4 light angles. With no silhouette cues, from only shading information, our optimization pushes down the divot to just short of the required depth, as demonstrated by the difference image (right, intensity magnified 10x). 
}
\label{fig:deathstar}
\end{figure}

\paragraph{Shape From Shading}
\label{sec:shadingvssilhouette}
Since we use multiple views, it is possible that the quality of geometry reconstruction is mainly due to the large percentage of the target object which is seen in silhouette.
We conduct an experiment to reconstruct geometry in the absence of silhouette edges and from shading only.
The target object is a sphere with a divot that does not affect the shape's silhouette. 
Figure~\ref{fig:deathstar} shows reconstruction results for this case.
Starting with an initial sphere mesh, our optimizer depresses the surface down to just short of the target depth, demonstrating that concave surface detail can be accurately captured. This behavior is critical to reconstruct surfaces whose concavities cannot be fully captured by silhouette-only initialization.

\subsection{Material Optimization}
\label{sec:matopt}
\revision{In alternation with geometry optimization, we also optimize for the material of the object.
The remainder of this section motivates and details each component of our material optimization stage.}

\paragraph{Representation}
\revision{We represent material as a Torrance-Sparrow BRDF~\cite{torrancesparrow}, a physically-based microfacet model commonly used in material acquisition research~\cite{guarnera2016brdf}.}
It is parameterized by a diffuse albedo and a specular roughness, traditionally represented by UV-mapped textures.
However, UV maps are ill-suited for our setting as our mesh is constantly changing, and we would need to simultaneously optimize the UV surface parameterization with the contents of the texture image. 
Instead, we use mesh colors \cite{Yuksel:2010:MC:1731047.1731053}: an adaptive-resolution extension of vertex colors. This is suitable for our optimization as it (a) provides an automatic coarse-to-fine level of detail which is tied to the underlying triangle mesh geometry, (b) does not require a surface parameterization, and (c) is seamless by construction.

\revision{ We modify the path tracing framework of Li et al.~\cite{redner} to support mesh colors.
A mesh colors texture is stored in a 1-D array $\mathbf{a}$, the size of which is determined by the number of triangles in the mesh and an integer resolution level $r$.
Given the barycentric coordinates $(\alpha, \beta)$ of a point in triangle number $t$, the texel for that point is
\begin{align*}
    \mathbf{m}(r, t, \alpha, \beta) &= \mathbf{a}[k]\\
    k &= \frac{t \cdot (r + 1) \cdot (r + 2) + i \cdot (2r - i + 3)}{2} + j\\
    (i, j) &= \lfloor{r \cdot (\alpha, \beta)}\rfloor
\end{align*}
This storage scheme duplicates edge and vertex detail for parallelism at the slight cost of additional memory. We use finite differences to compute derivatives.}

\paragraph{Initialization}
We optimize texture maps in a coarse-to-fine manner.
For the first five optimization cycles, we optimize for a single spatially invariant diffuse color and specular color per material. 
Using a low-degree-of-freedom material representation while the geometry is initially refined prevents the material from `baking in' appearance effects from small-scale geometry. 
To distinguish between materials, we use per-material image-space masks to segment distinct materials.
To correct for any inconsistencies in clustering from view to view, we determine which vertices belong to each material cluster by counting which material they are assigned to most often in image-space.

Every time the mesh changes topology (i.e., after a remeshing step), we re-optimize from a neutral gray color to avoid bias towards previous errors in geometry or texture. The texture loss is:
\begin{equation*}
    \mathcal{L}(\mathbf{x}, \mathbf{m}) = \frac{1}{n}\sum_{i=1}^{n}\sum_{j=1}^{s}(\mathcal{M}_{i,j} \cdot \mathcal{T}_i - \mathcal{M}_{i,j} \cdot R(\mathbf{x}, \mathbf{m}, c_i) )^2
\end{equation*}
where colors or per-triangle texels $\mathbf{m}$ are optimizable, and $\mathcal{M}_{i,j}$ is the $j$th material mask for the $i$th target image.

\paragraph{Spatially-varying Diffuse}

After the first five optimization cycles, we expand the parameter space to allow a spatially-varying diffuse material, initialized from the last spatially invariant diffuse color, while keeping the single specular value. \revision{We estimate areas where we expect specular highlights to occur by rendering the geometry as a perfect mirror and thresholding bright areas. Then we mask out these bright regions from renders. This removes gradients where specular highlights occur.} As highlights vary from target frame to frame and have high radiance, they otherwise tend to `bake' into spatially-varying diffuse texture.

\paragraph{Spatially-varying Specular}
For the last optimization cycles, we optimize for a spatially-varying specular material.
As low-variance (and often constant) specular maps are a common artistic choice, we add a variance penalty on the specular mesh colors $\mathbf{m}_s$:
    $\mathcal{L}(\mathbf{x}, \mathbf{m}) = \frac{1}{n}\sum_{i=1}^{n}(\mathcal{M}_i \cdot \mathcal{T}_i - \mathcal{M}_i \cdot R(\mathbf{x}, \mathbf{m}, c_i) )^2 + \lambda \text{Var}(\mathbf{m}_s)$.

\begin{figure}[t!]
  \centering
  \setlength{\tabcolsep}{1pt}
    \begin{tabular}{rccc}
        & Target & Local + Shadows & Global
        \\
        \raisebox{2em}{\rotatebox{90}{Render}} &
        \includegraphics[width=.30\linewidth]{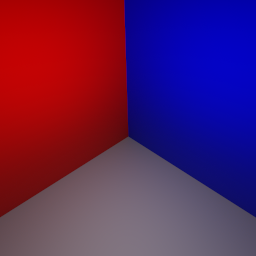} &
        \includegraphics[width=.30\linewidth]{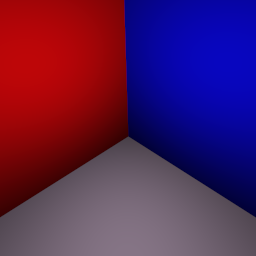} &
        \includegraphics[width=.30\linewidth]{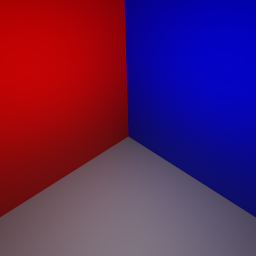}
        \\
        \raisebox{2em}{\rotatebox{90}{Albedo}} &
        \includegraphics[width=.30\linewidth]{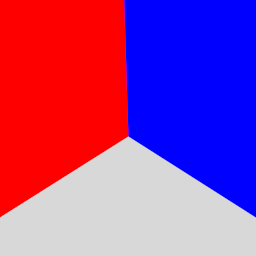} &
        \includegraphics[width=.30\linewidth]{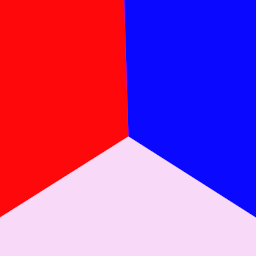} &
        \includegraphics[width=.30\linewidth]{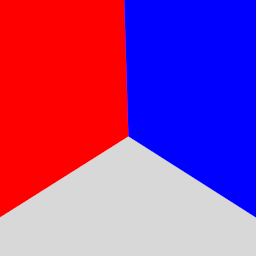}
    \end{tabular}
\caption{
\revision{Material optimization using local vs.~global illumination. Top: each strategy's output---all quite similar. Bottom: unlit albedo of each strategy. The local case \emph{(Bottom Middle)} `bakes' purple diffuse interreflection into the floor's albedo, while a differentiable path tracer disambiguates this effect \emph{(Bottom Right)}.}
}
\label{fig:color_bleeding}
\end{figure}

\paragraph{Effects of Global Illumination}
\label{sec:matgi}
Previous differentiable renderering-based reconstruction methods do not use global illumination. We justify the use of global illumination by investigating the effect of diffuse interreflection on material estimation.
We set up a virtual scene consisting of three intersecting planes, each with a constant diffuse albedo: red, blue, and white.
The geometry is known, and we optimize a SVBRDF material for this geometry given a rendered target.
We compare results to a version in which the renderer uses only one-bounce illumination. Figure~\ref{fig:color_bleeding} shows the results.
Optimization with global illumination correctly reconstructs the ground-truth albedos, while the one-bounce version explains the purplish floor (caused by color bleeding) by baking this color into the ground plane's albedo.

\subsection{Optimization Details}


Our optimization procedure alternates between solving for geometry and material, switching once the loss has converged.
Optimization proceeds until the loss stops improving between successive cycles.

\section{Results}
\label{sec:results}

In this section, we explore reconstructing complex object geometry and material in simulation and from real-world input.
All results were produced on desktops with an AMD Ryzen 2700X and an NVIDIA GTX 1080Ti.

\subsection{Reconstructing Simulated Objects}

\begin{figure}[t!]
  \centering
  \setlength{\tabcolsep}{1pt}
    \begin{tabular}{rcccc}
        & \textbf{Target} & \textbf{Init} & \textbf{Reconstruction} & 
        \\
        \raisebox{1em}{\rotatebox{90}{Render}} &
        \includegraphics[width=.23\linewidth]{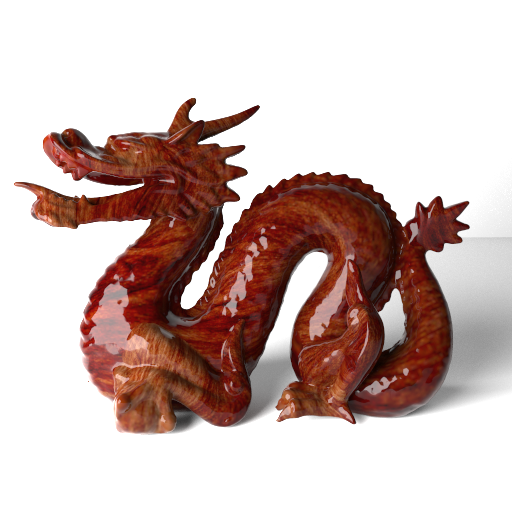} &
        \includegraphics[width=.23\linewidth]{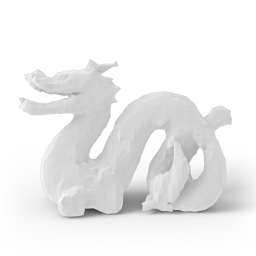} &
        \includegraphics[width=.23\linewidth]{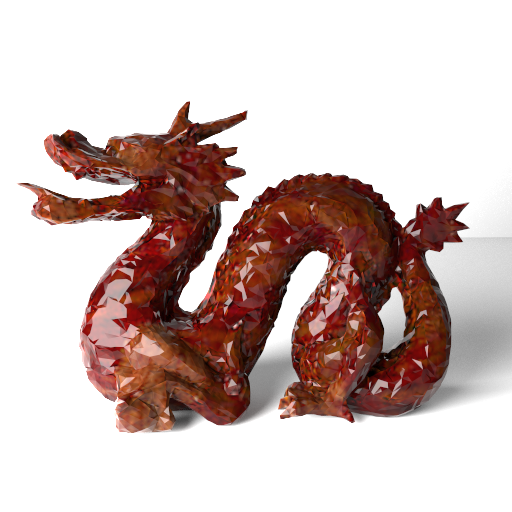}
        \includegraphics[width=.23\linewidth]{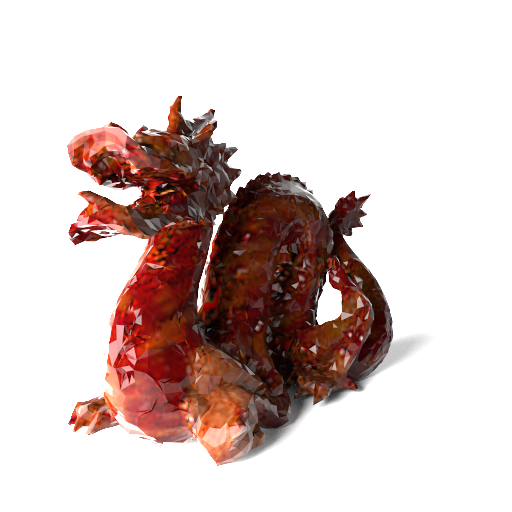}
        \\
        \raisebox{2em}{\rotatebox{90}{Geo}} &
        \includegraphics[width=.23\linewidth]{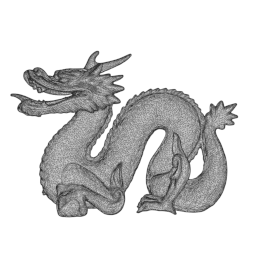} &
        \includegraphics[width=.23\linewidth]{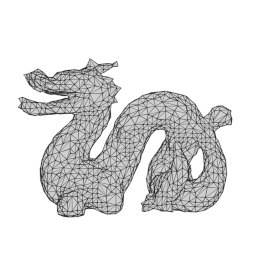} &
        \includegraphics[width=.23\linewidth]{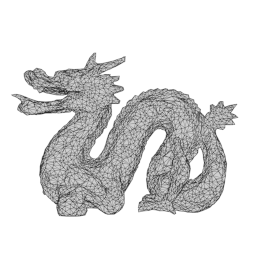}
        \includegraphics[width=.23\linewidth]{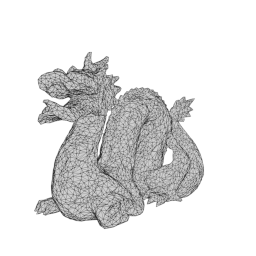}
        \\
        \raisebox{1em}{\rotatebox{90}{Albedo}} &
        \includegraphics[width=.23\linewidth]{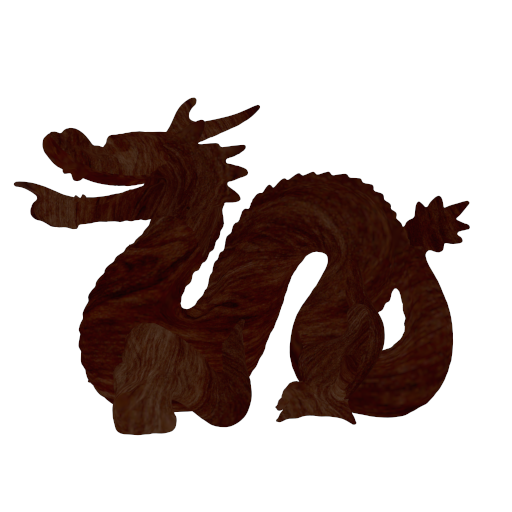} &
        \includegraphics[width=.23\linewidth]{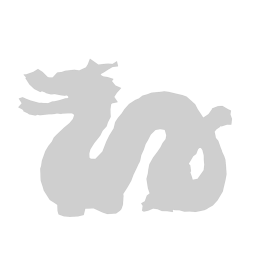} &
        \includegraphics[width=.23\linewidth]{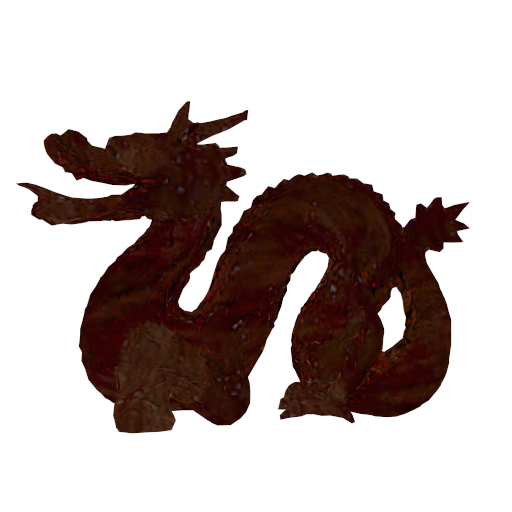}
        \includegraphics[width=.23\linewidth]{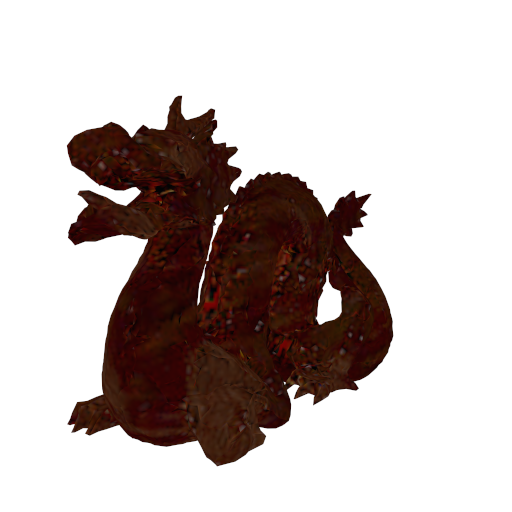}
        \\
        \raisebox{1em}{\rotatebox{90}{Specular}} &
        \includegraphics[width=.23\linewidth]{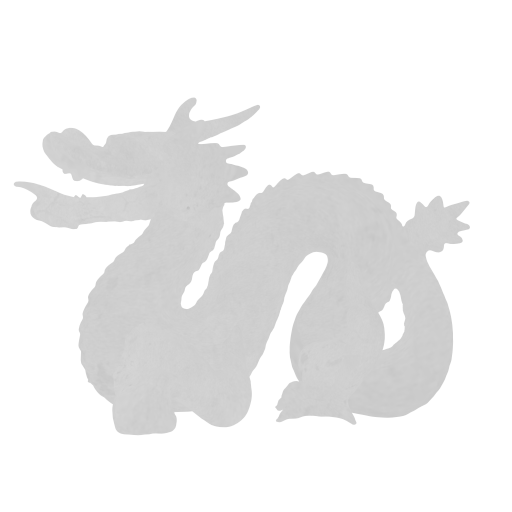} &
        \includegraphics[width=.23\linewidth]{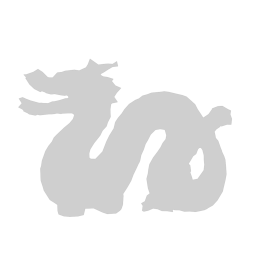} &
        \includegraphics[width=.23\linewidth]{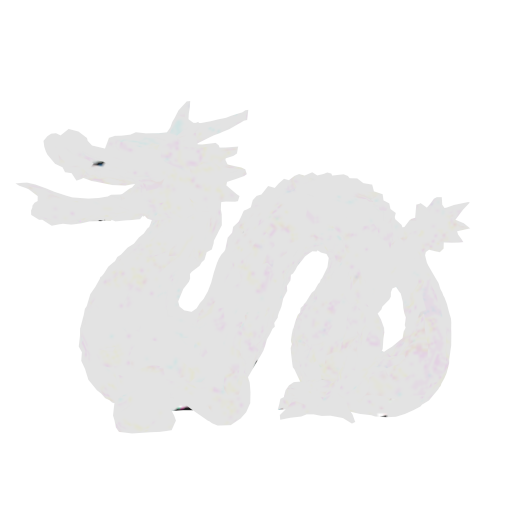}
        \includegraphics[width=.23\linewidth]{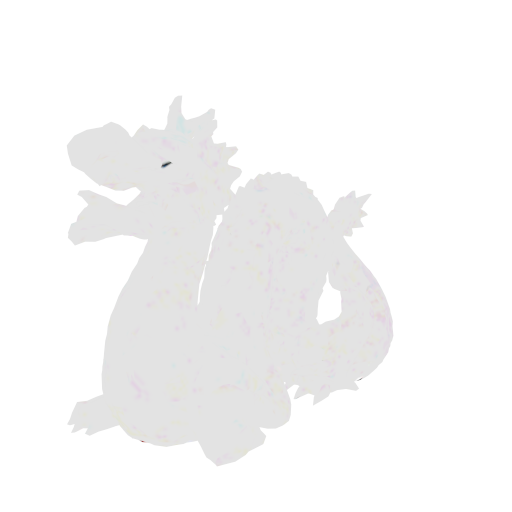}
    \end{tabular}
    \caption{
    Reconstruction of the Stanford dragon inside a Cornell Box, with a glossy wood material.
     The image resolution was $128 \times 128$. \revision{Col. 2 shows initializations; Col. 4 shows a novel view. Although the geometry reconstruction exhibits dense triangle clusters, it is intersection-free: clusters are caused by our adaptive remeshing, which concentrates vertices in high detail areas. Re-renderings were produced using the Blender Cycles renderer.}. 
    }
\label{fig:dragon_results}
\end{figure}

\begin{table}[t!]
    \centering
    \revision{
    \caption{Quantitative results for Figure~\ref{fig:dragon_results}. We report Chamfer distance and F1 scores between initialization/target (condition 1) and reconstruction/target geometries (condition 2).  We report PSNR \emph{(Left)} and SSIM \emph{(Right)} between initialization/target and reconstruction/target renders.}
    \setlength{\tabcolsep}{2pt}
    \small
    \resizebox{\linewidth}{!}
    {
    \begin{tabular}{@{}llccccc@{}}
        \toprule
        \textbf{Object} & \textbf{Cond.} & \textbf{Chamfer} & \textbf{F1} & \textbf{Full} & \textbf{Diffuse} & \textbf{Specular}\\
        \midrule
        \multirow{ 2}{*}{\emph{Dragon}}
        & Init. & 0.191 & 44.61 & 9.82, 0.69 & 10.44, 0.68 & 27.37, 0.70\\
        & Recon. & \textbf{0.149} & \textbf{87.51} & \textbf{32.19}, \textbf{0.91} & \textbf{33.89}, \textbf{0.87} & \textbf{43.78}, \textbf{0.98} \\
        \midrule
        \multirow{ 2}{*}{\emph{Armadillo}}
        & Init. & 0.158 & 47.97 & 12.16, 0.73 & 12.34, 0.72 & 28.94,0.69 \\
        & Recon. & \textbf{0.147} & \textbf{98.72} &  \textbf{29.25}, \textbf{0.89} & \textbf{26.58}, \textbf{0.84} & \textbf{36.83},\textbf{0.94} \\
        \midrule
        \multirow{ 2}{*}{\emph{Buddha}}
        & Init. & 0.172 & 47.97 & 15.20, 0.85 & 15.98, 0.84 & 22.47, 0.72 \\
        & Recon. & \textbf{0.155} & \textbf{87.68} & \textbf{31.96}, \textbf{0.96} & \textbf{33.80}, \textbf{0.94} & \textbf{32.24}, \textbf{0.88} \\
        \bottomrule
    \end{tabular}
    }
    }
    \label{tab:simulation_results}
    \vspace{-0.25cm}
\end{table}

\revision{We test our findings on the task of reconstructing 3D objects with known geometry and material in simulation, using meshes varying in genus and texture patterns.

We use 32 cameras distributed on a Fibonacci sphere surrounding the object within a Cornell box, 
with two light position variations per view;} one light is aligned with the camera and another is placed at a constant offset of 1 unit in the camera's tangent direction. Using multiple light positions is critical to distinguish between geometric surface details, diffuse albedo, and specular highlights.
\revision{Figure~\ref{fig:dragon_results} shows qualitative results (with additional results in supplemental material).
Table~\ref{tab:simulation_results} shows numerical results.}


\begin{figure*}[t!]
    \centering
    \setlength{\tabcolsep}{1pt}
    \begin{tabular}{cc c cc c cc c cc c cc}
      \multicolumn{2}{c}{Two Input Views} &
      \hspace{0.4em} &
      \multicolumn{2}{c}{COLMAP Initialization} &
      \hspace{0.4em} &
      \multicolumn{2}{c}{Space Carving} &
      \hspace{0.4em} &
      \multicolumn{2}{c}{Our Mesh + SVBRDF} &
      \hspace{0.4em} &
      \multicolumn{2}{c}{Ours (re-lit)}
      \\
      \includegraphics[width=0.09\linewidth]{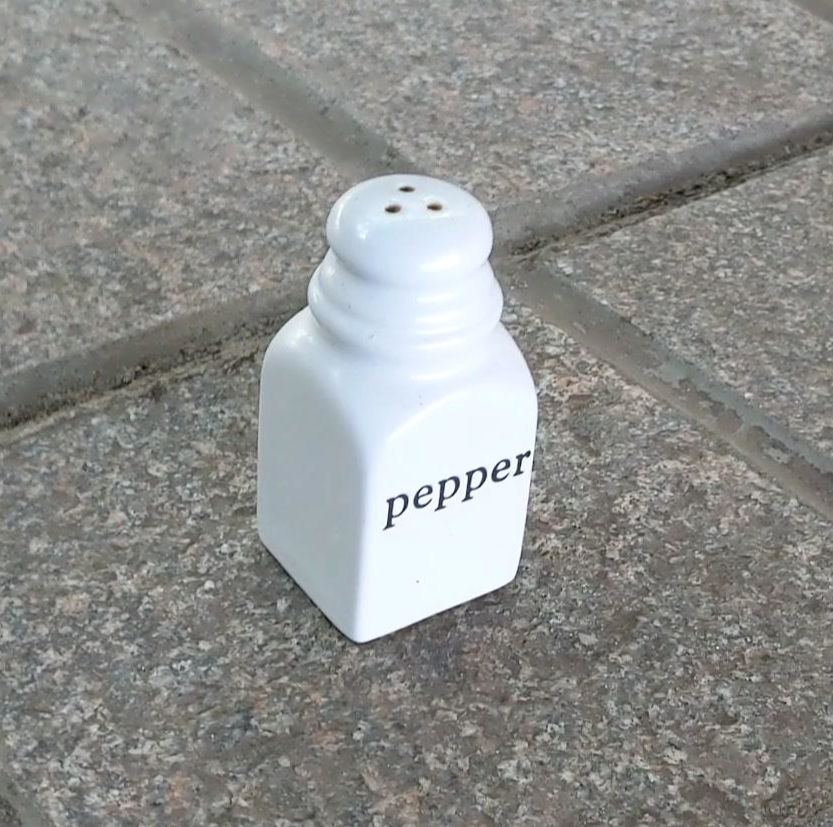} &
      \includegraphics[width=0.09\linewidth]{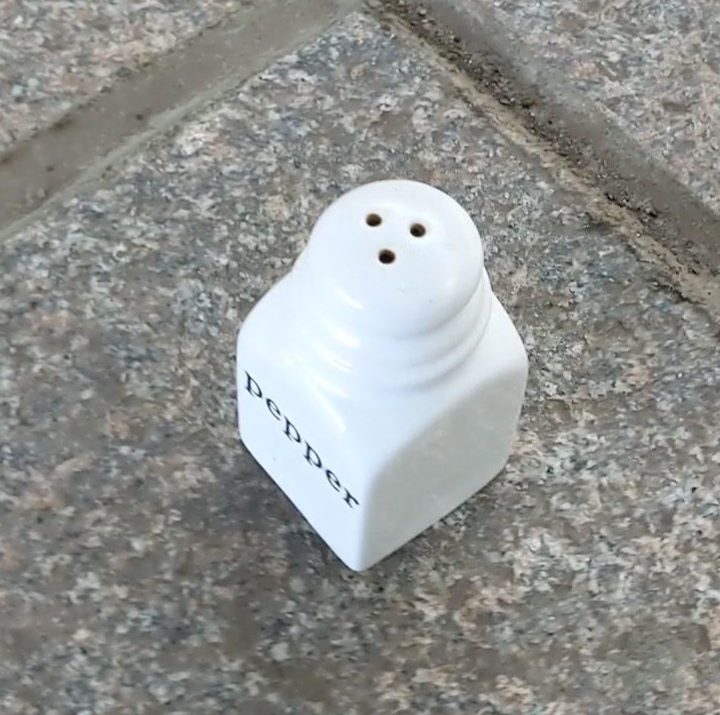} &
      &
      \includegraphics[width=0.09\linewidth]{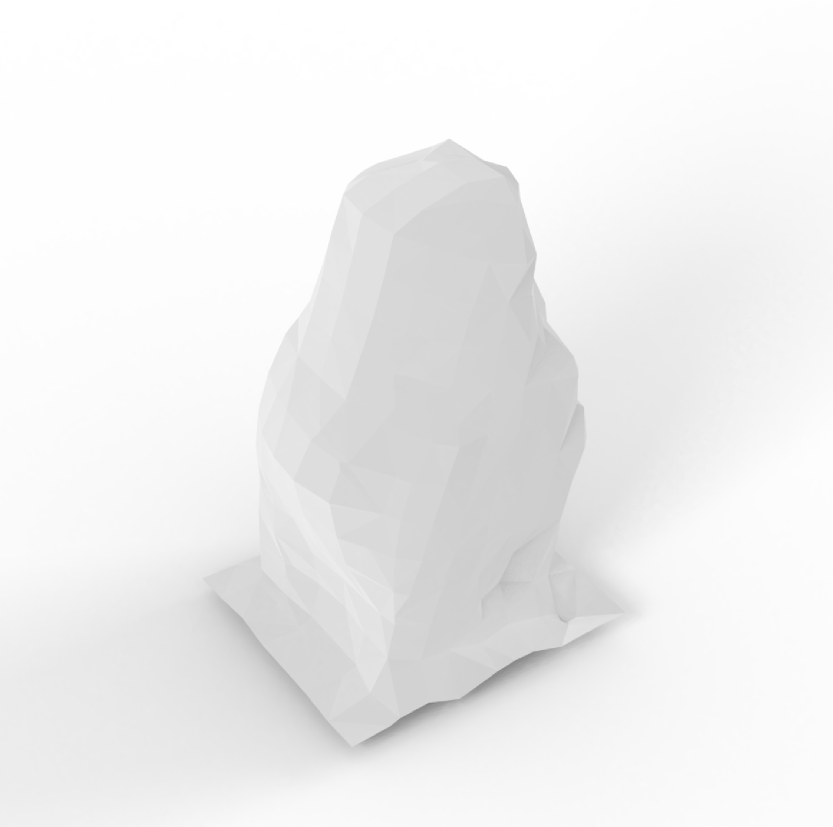} &
      \includegraphics[width=0.09\linewidth]{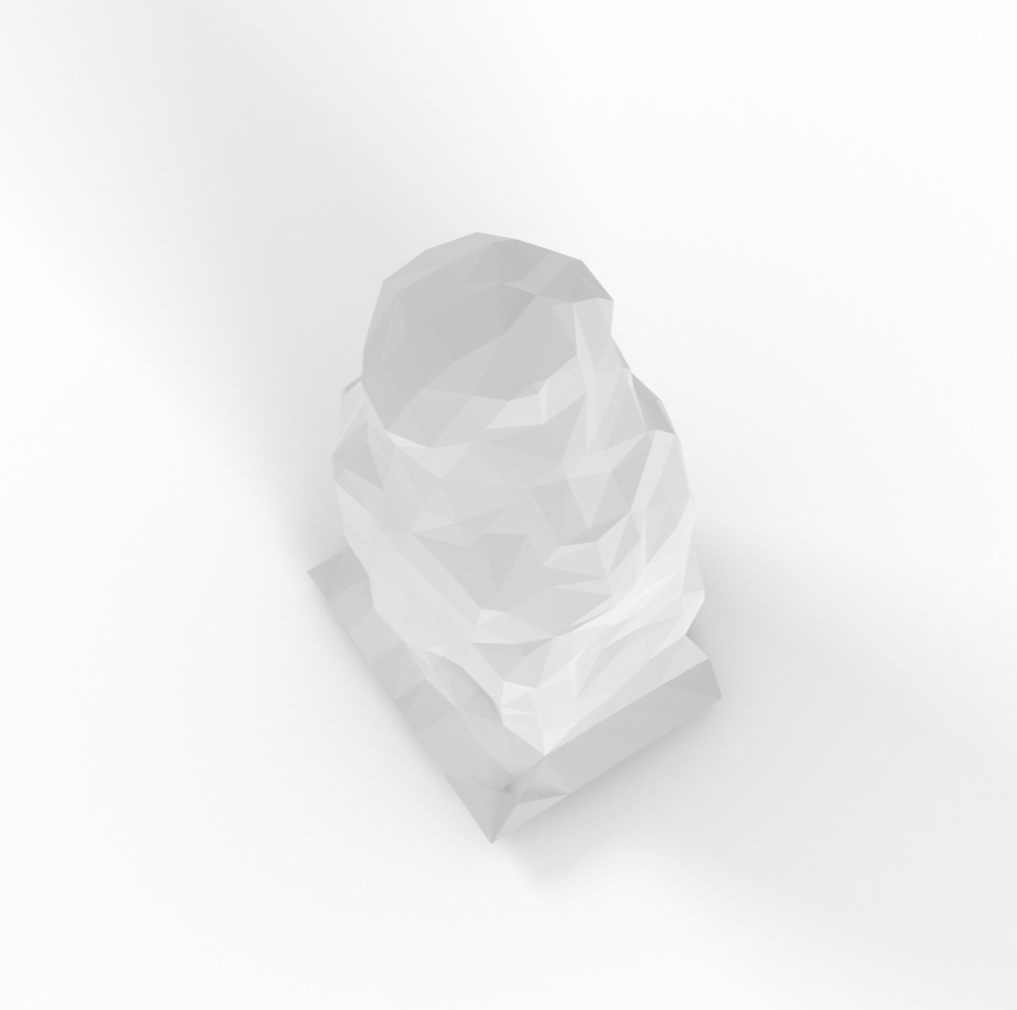} &
      &
      \includegraphics[width=0.09\linewidth]{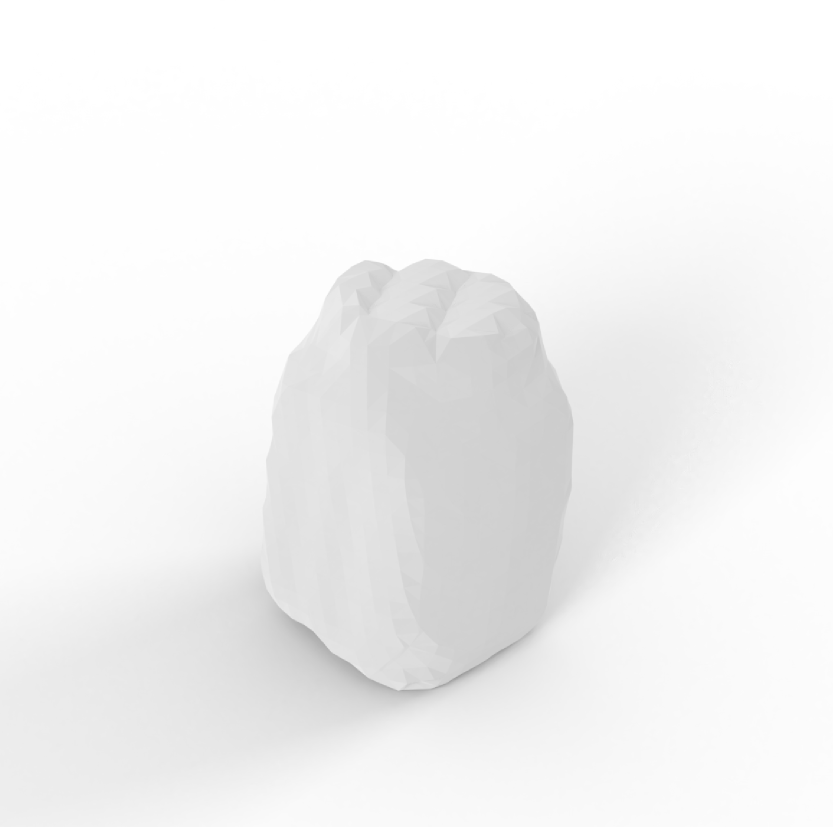} &
      \includegraphics[width=0.09\linewidth]{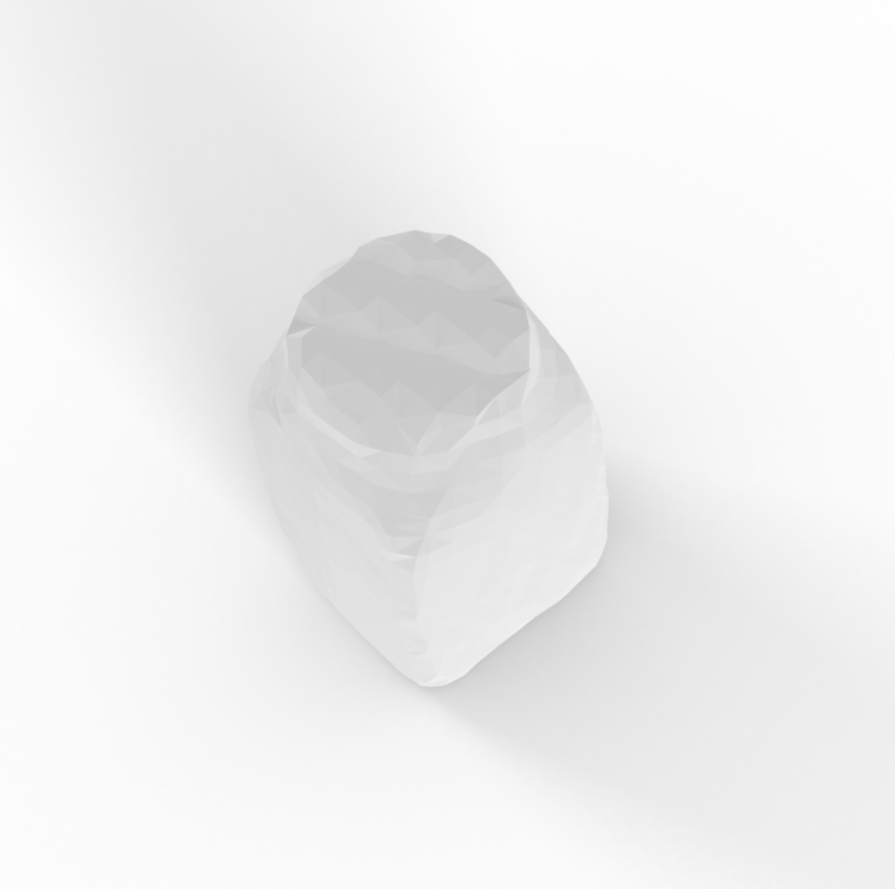} &
      &
      \includegraphics[width=0.09\linewidth]{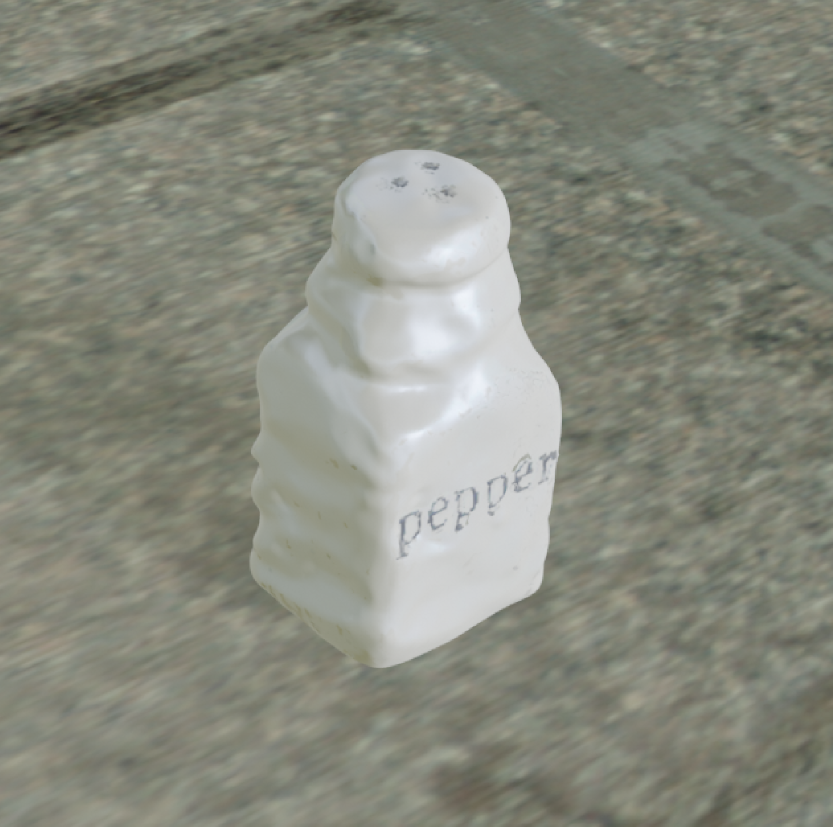} &
      \includegraphics[width=0.09\linewidth]{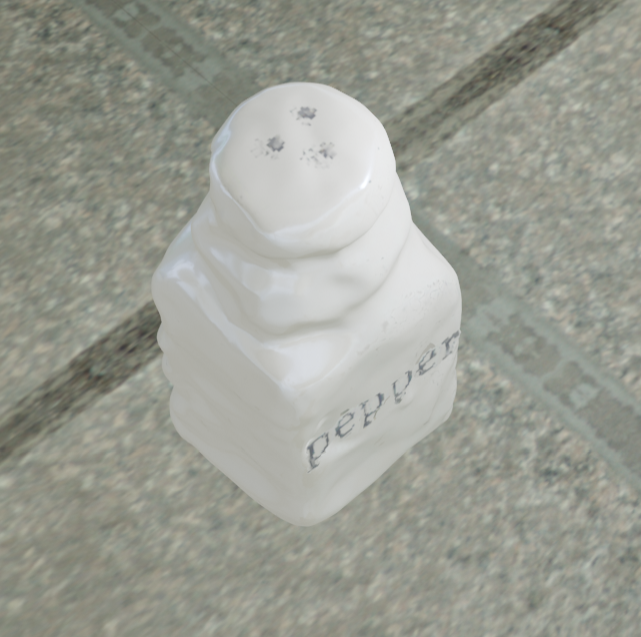} &
      &
      \includegraphics[width=0.09\linewidth]{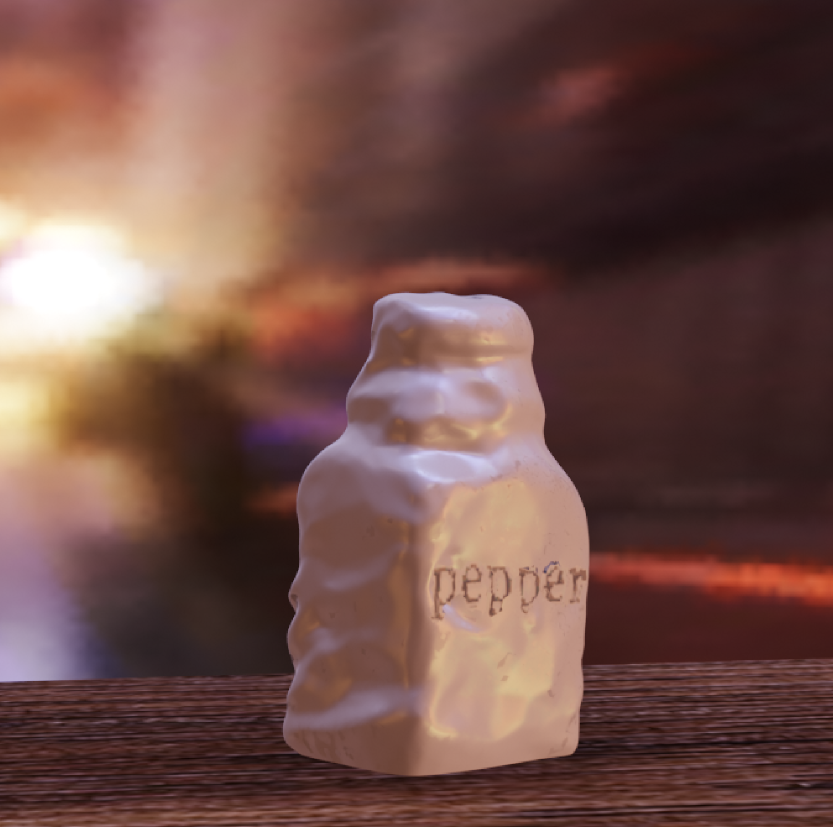} &
      \includegraphics[width=0.09\linewidth]{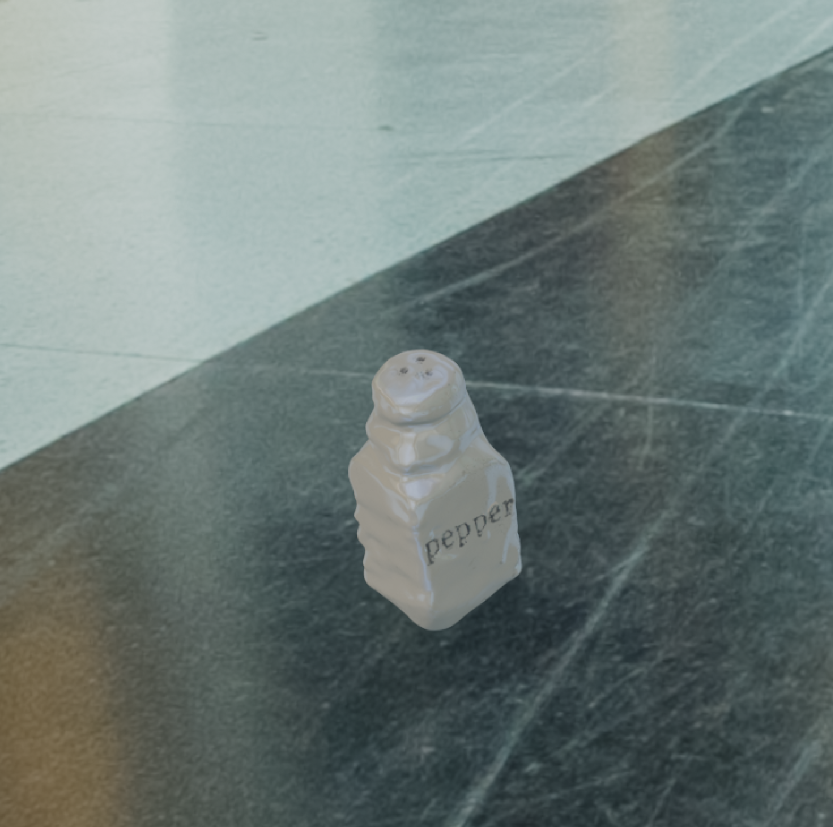}
      \\
      \includegraphics[width=0.09\linewidth]{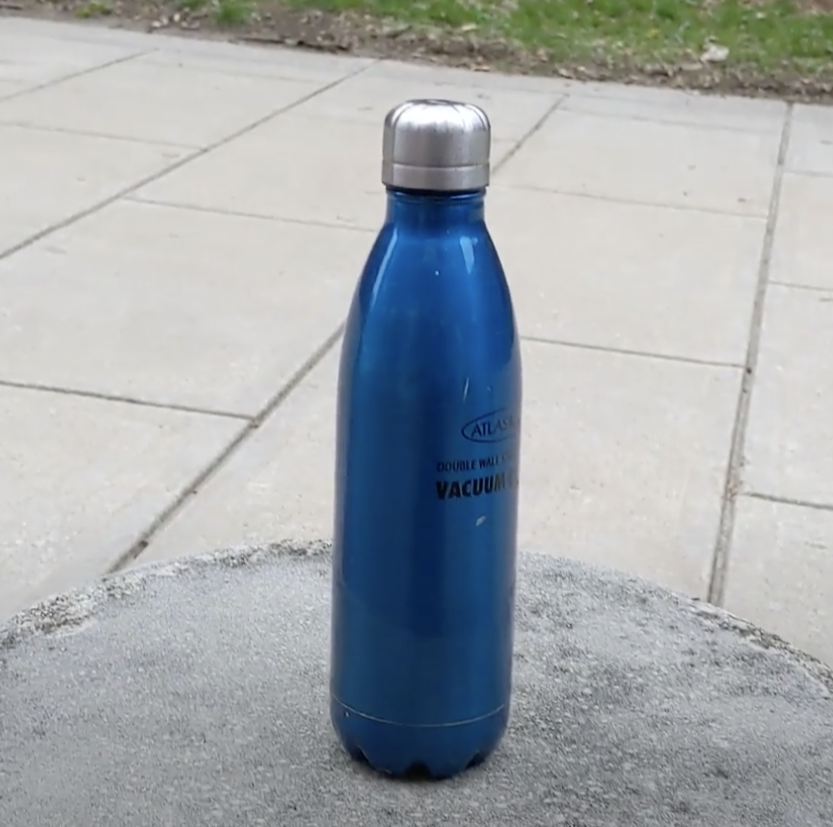} &
      \includegraphics[width=0.09\linewidth]{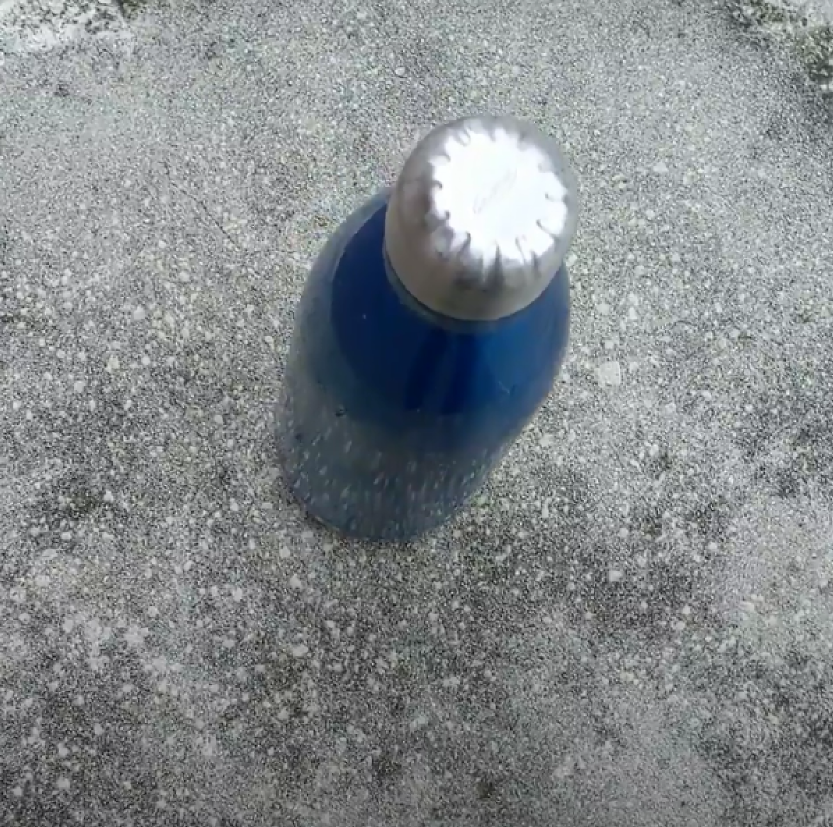} &
      &
      \includegraphics[width=0.09\linewidth]{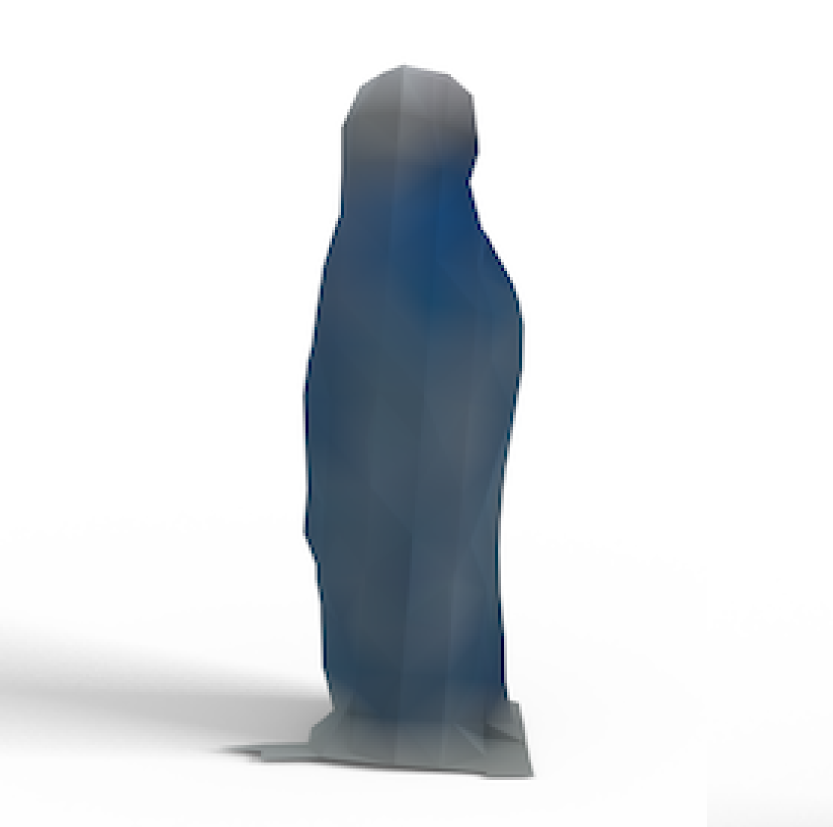} &
      \includegraphics[width=0.09\linewidth]{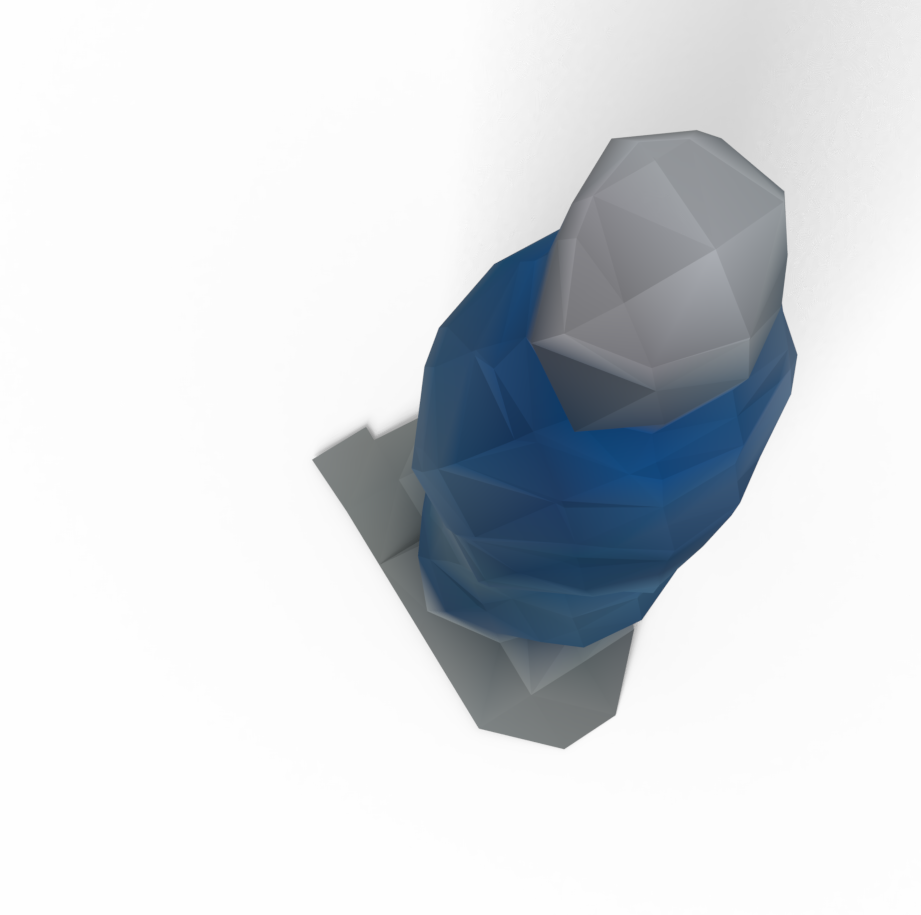} &
      &
      \includegraphics[width=0.09\linewidth]{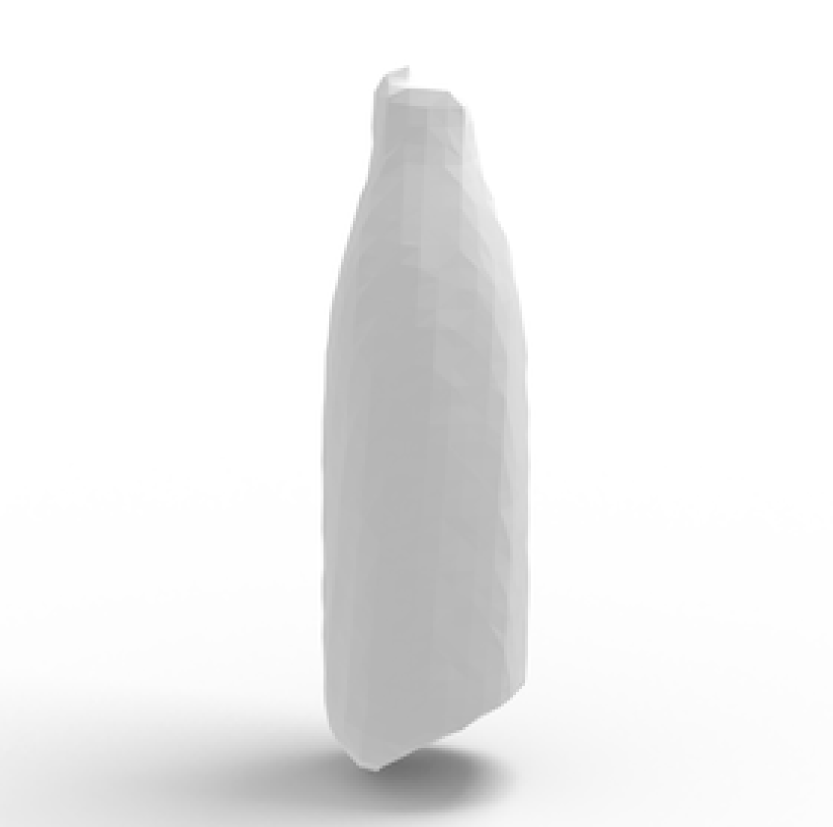} &
      \includegraphics[width=0.09\linewidth]{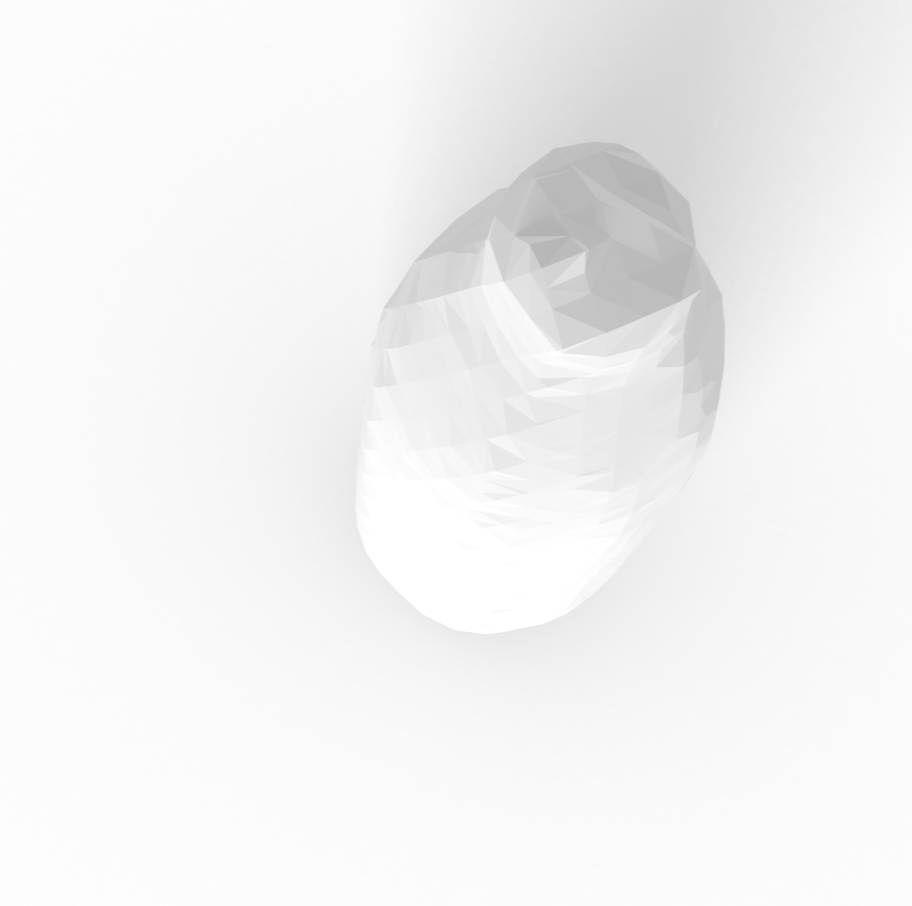} &
      &
      \includegraphics[width=0.09\linewidth]{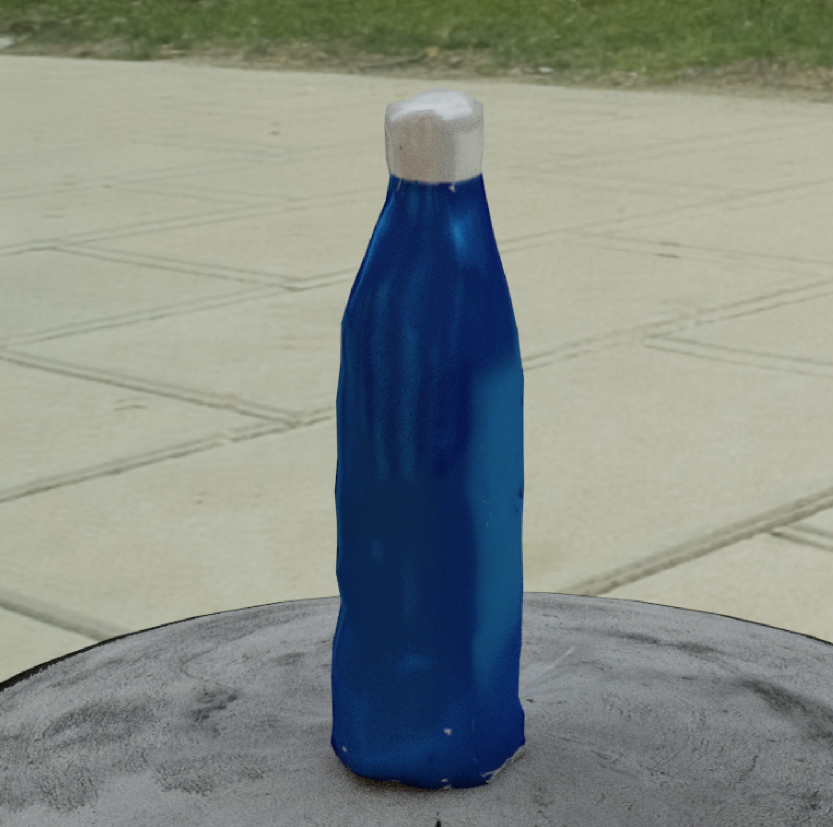} &
      \includegraphics[width=0.09\linewidth]{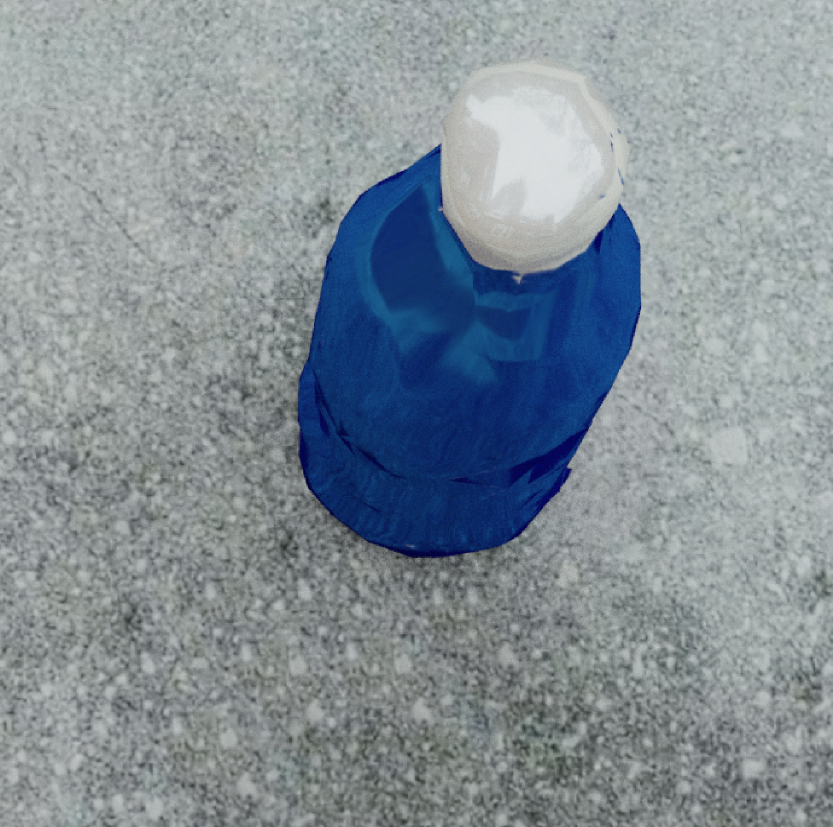} &
      &
      \includegraphics[width=0.09\linewidth]{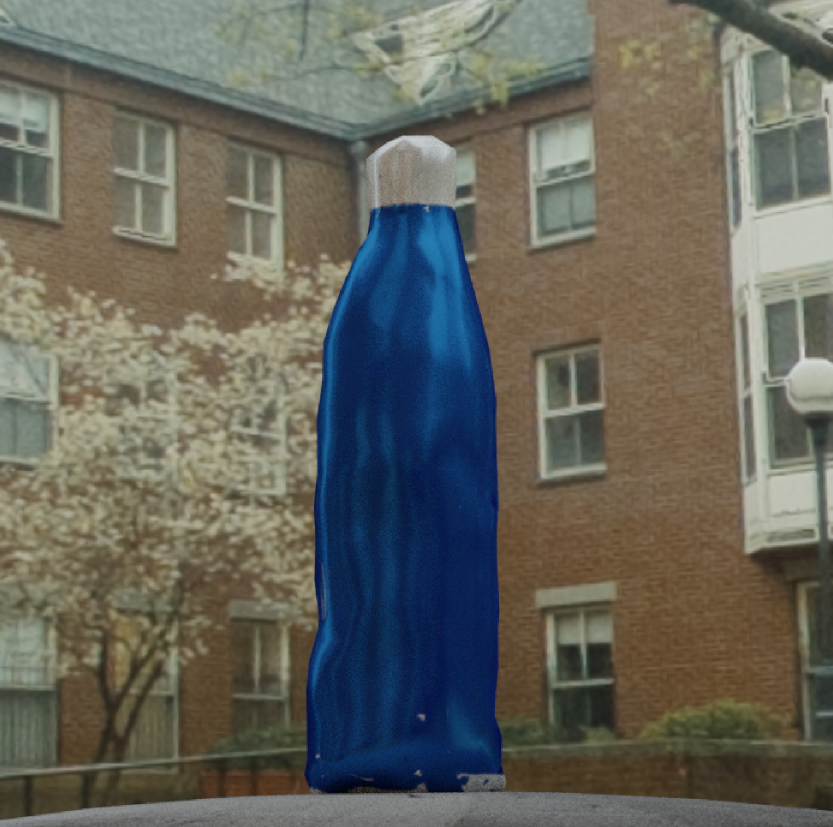} &
      \includegraphics[width=0.09\linewidth]{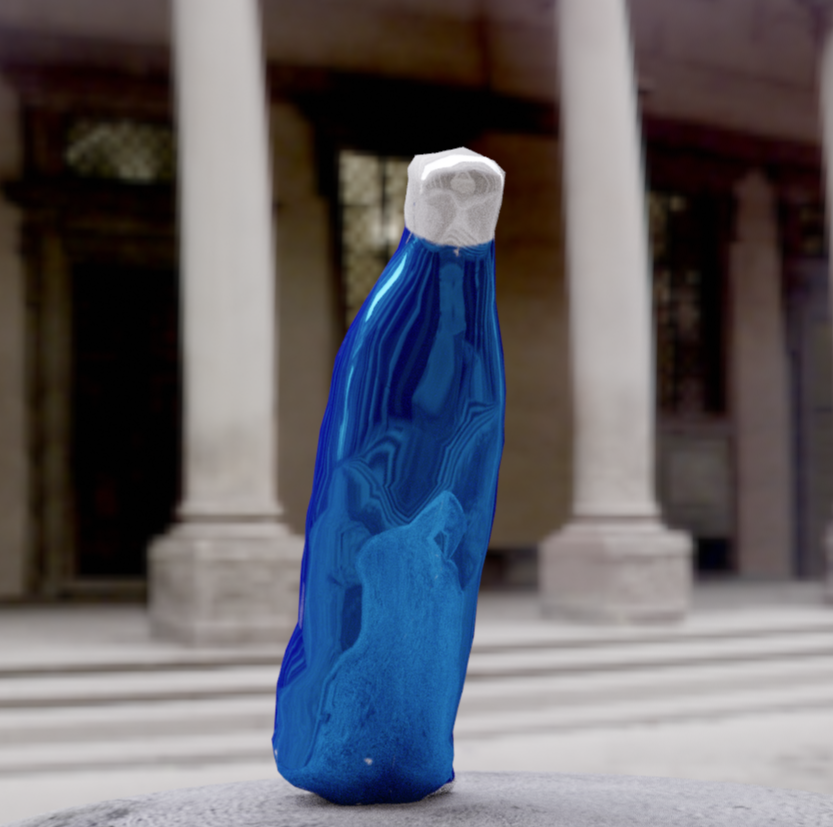}
      \\
      \includegraphics[width=0.09\linewidth]{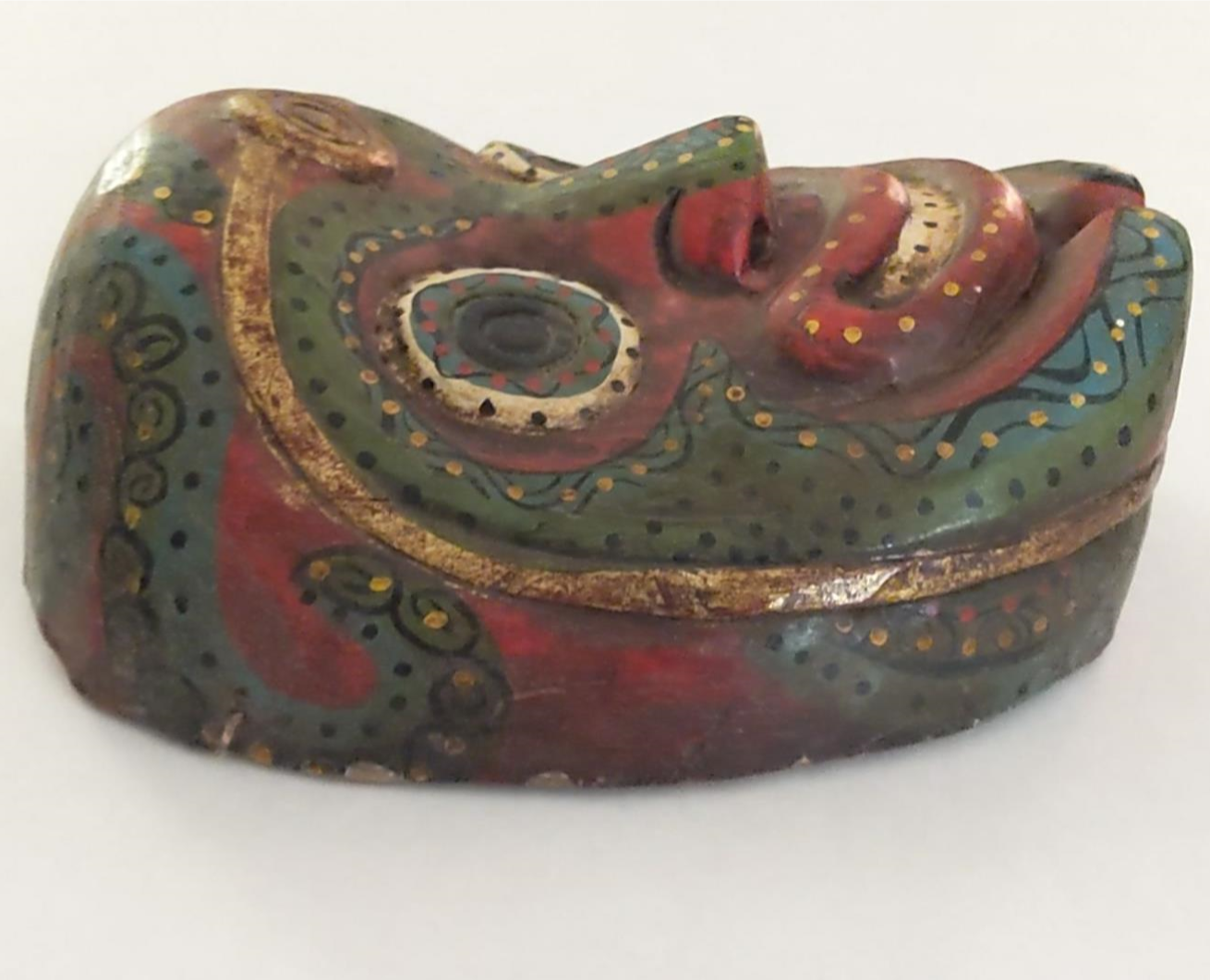} &
      \includegraphics[width=0.09\linewidth]{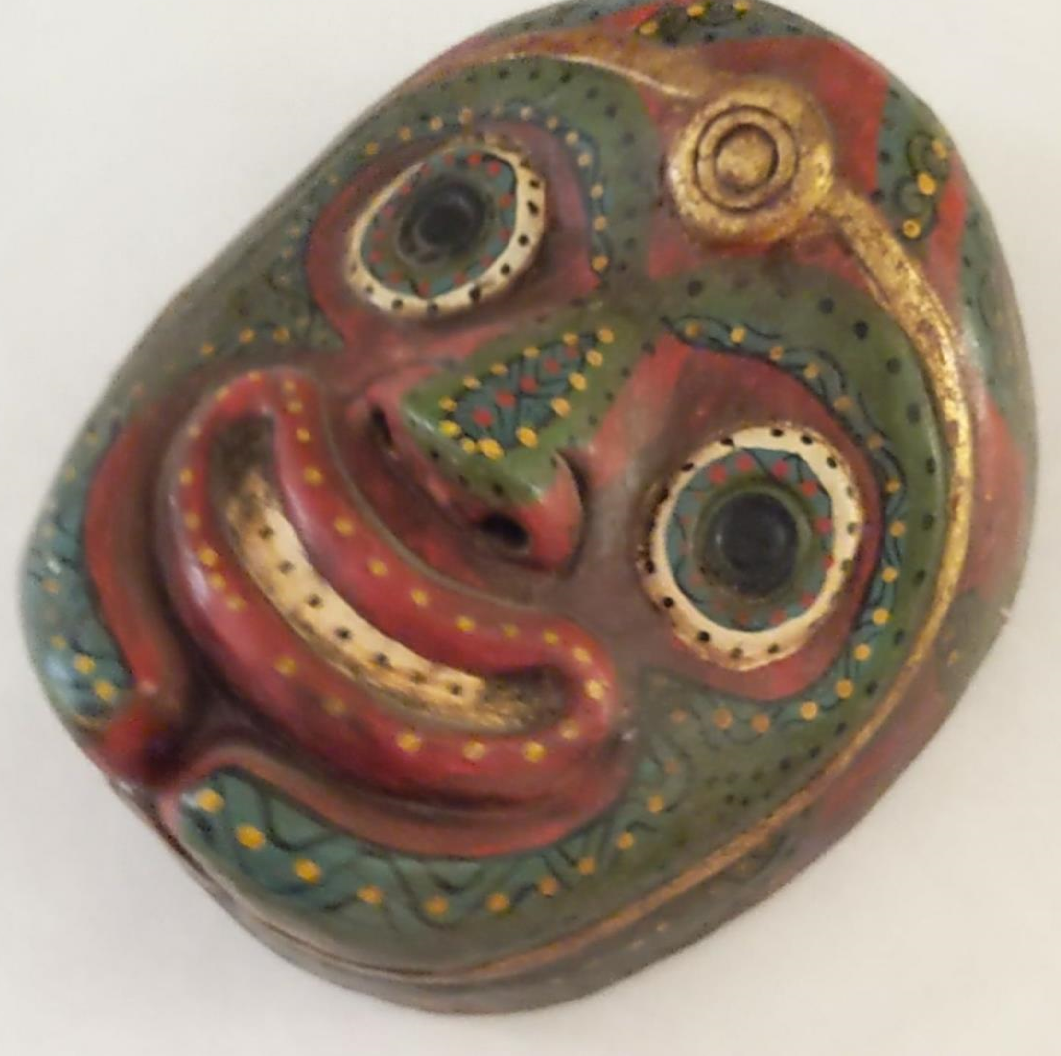} &
      &
      \includegraphics[width=0.09\linewidth]{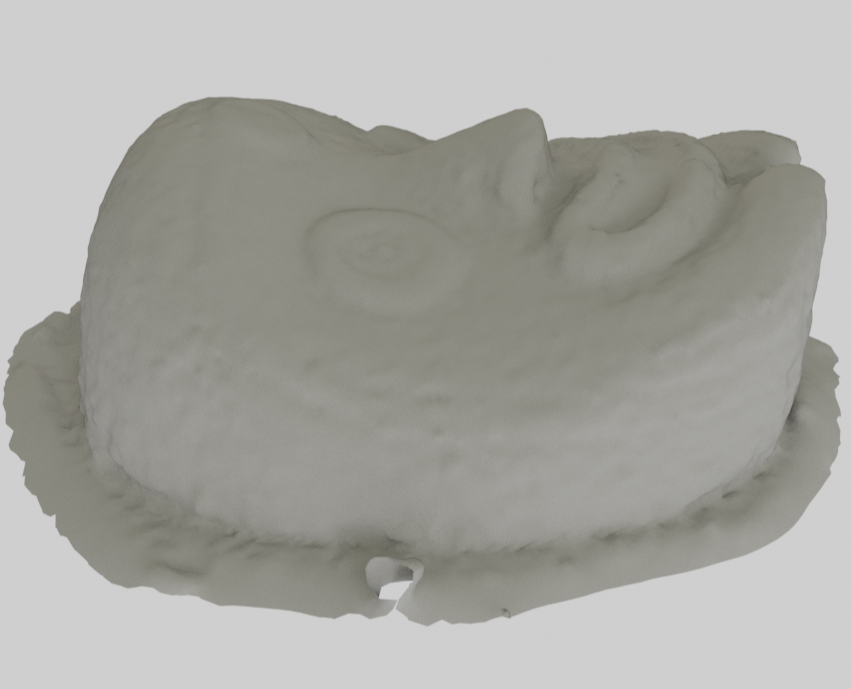} &
      \includegraphics[width=0.09\linewidth]{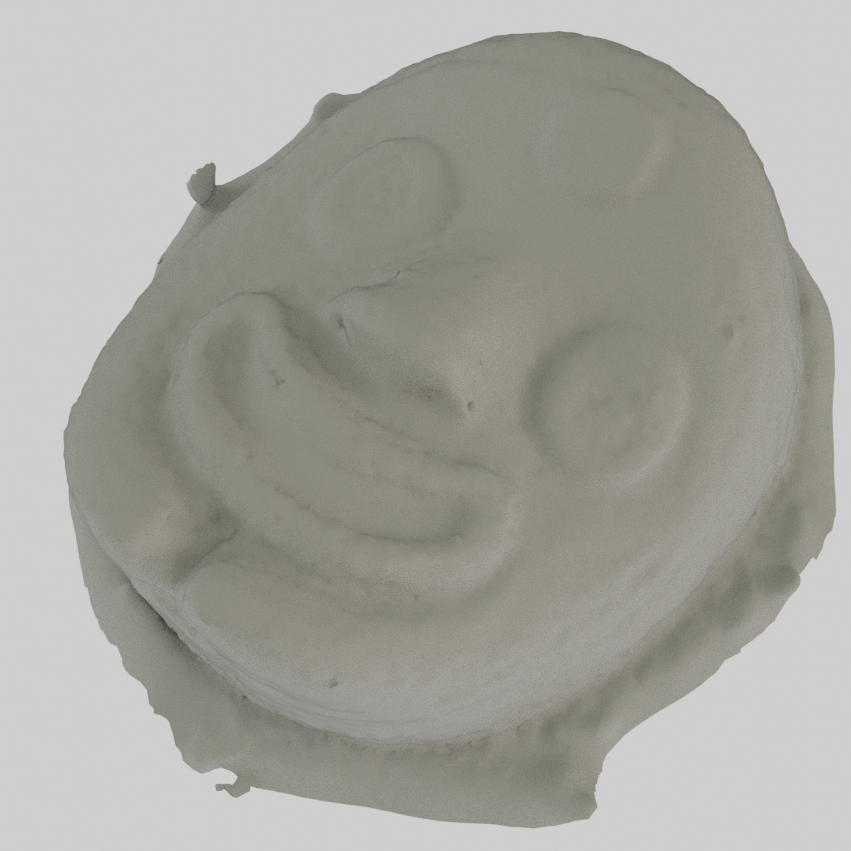} &
      &
      \includegraphics[width=0.09\linewidth]{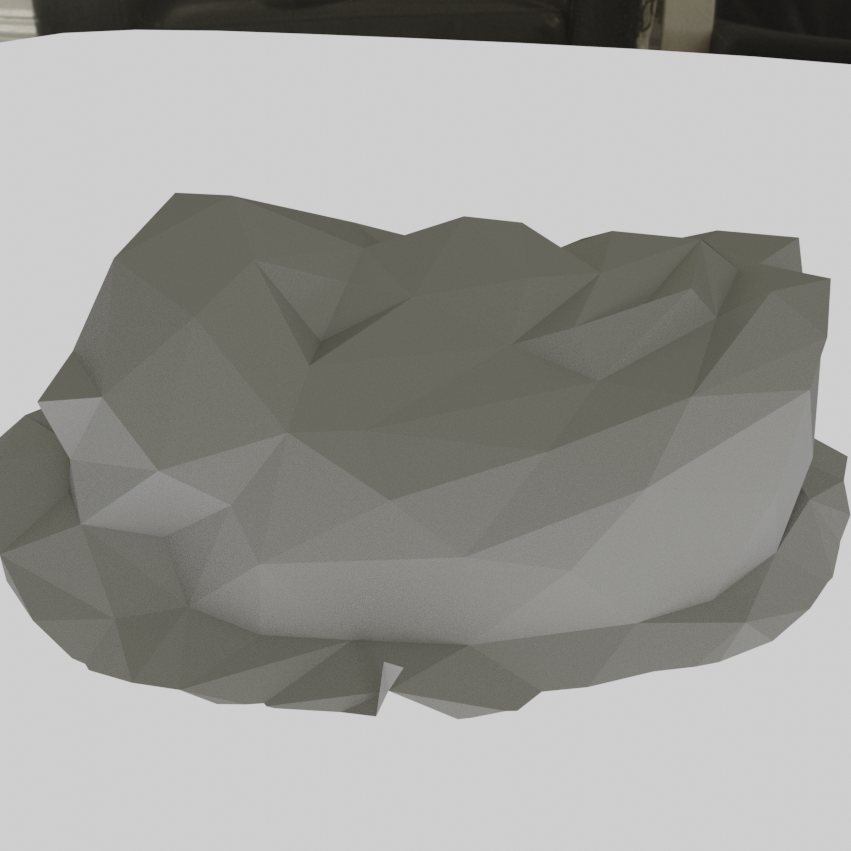} &
      \includegraphics[width=0.09\linewidth]{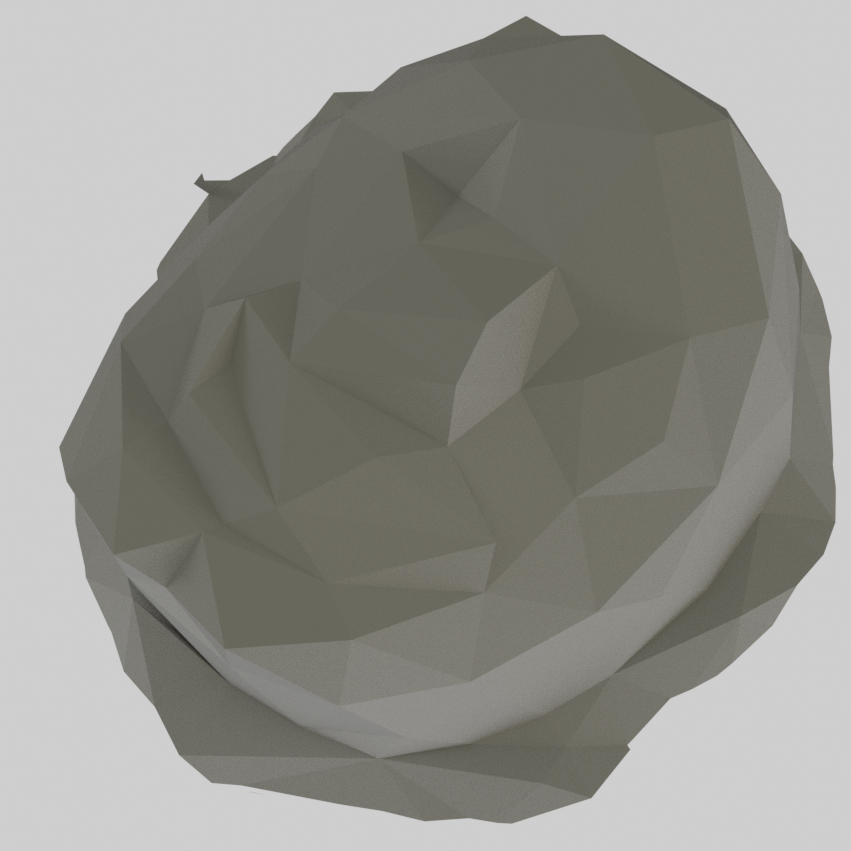} &
      &
      \includegraphics[width=0.09\linewidth]{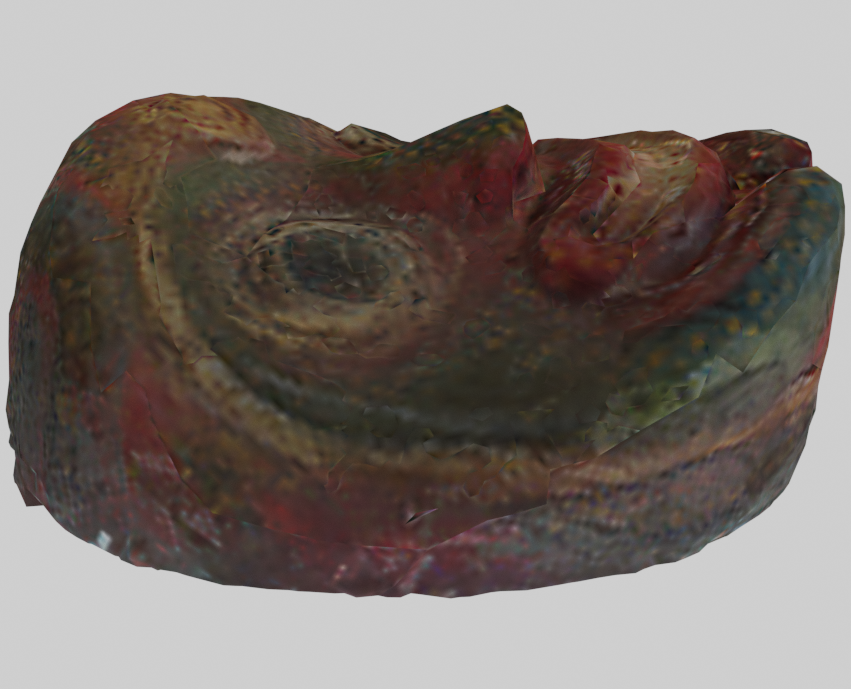} &
      \includegraphics[width=0.09\linewidth]{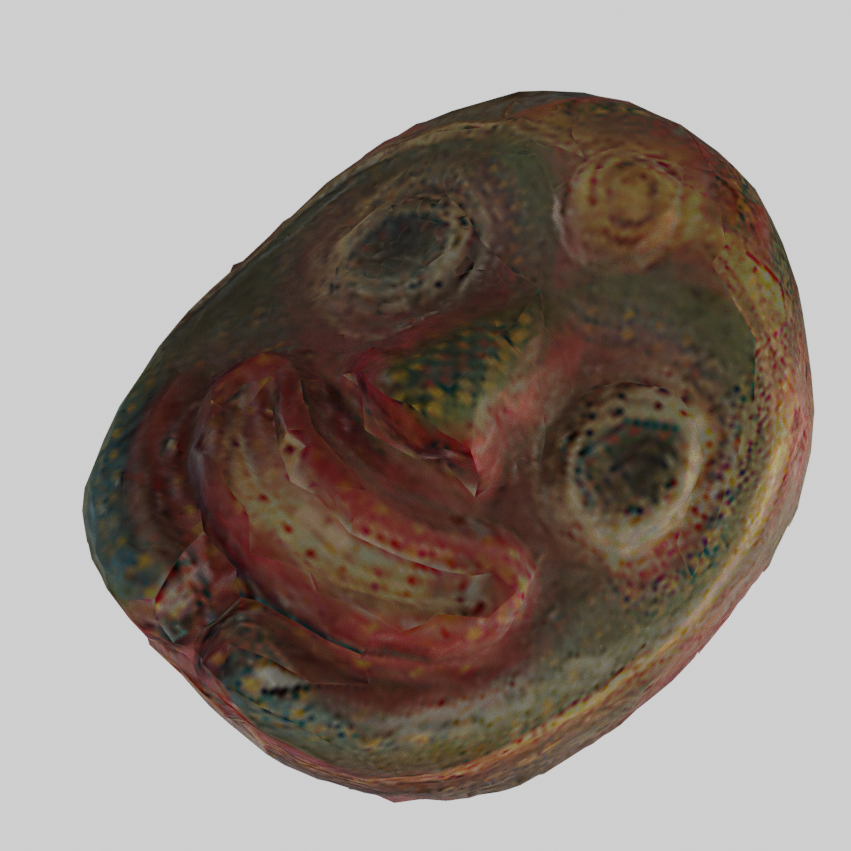} &
      &
      \includegraphics[width=0.09\linewidth]{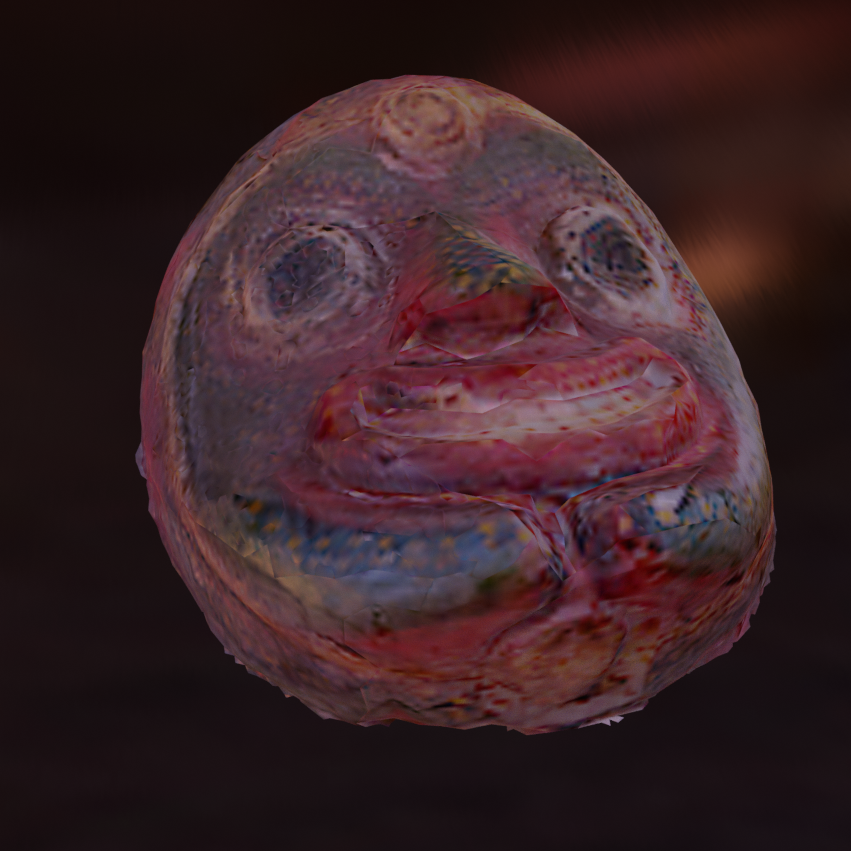} &
      \includegraphics[width=0.09\linewidth]{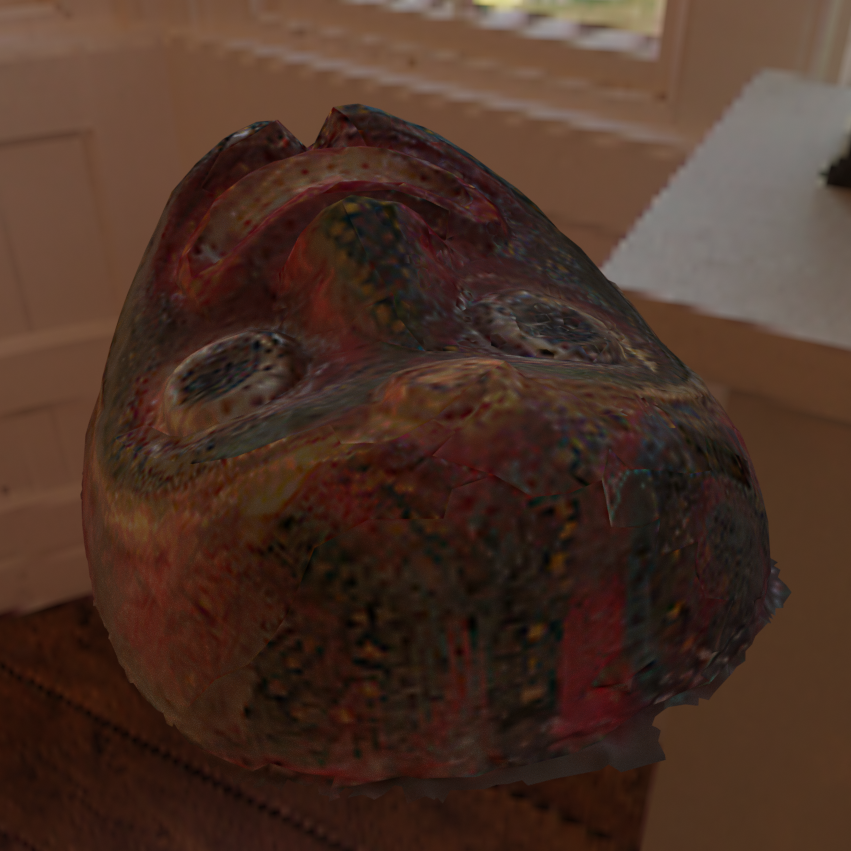}
  \end{tabular}
    \caption{\revision{Results from real world capture. We show two video frames from the smartphone camera \emph{(Col. 1)}, and the geometry initialization \emph{(Col. 2)}. Our geometry estimations compare favorably against their initializations and space carving reconstruction \emph{(Col. 3)}: note the entire upper half of the pepper shaker.  
    We show our final reconstruction rendered under captured \emph{(Col. 4)} and novel illumination \emph{(Col. 5)}. }
    }
    \label{fig:real_world_results}
\end{figure*}
\subsection{Reconstructing Real-World Objects}

Most existing real-world methods requires objects be photographed in HDR in a controlled environment, e.g., in a dark room with a few fixed lights. 
Here, we use differentiable path tracing to address a more challenging ill-posed scenario: an object casually-captured outdoors with a smartphone video and a 360 camera HDR. This is a difficult setting for reconstruction as we assume no constraints on illumination and only require a coarse estimate of scene lighting and a sparse number of views with estimated poses.

\vspace{-0.1cm}
\paragraph{Data capture}
We capture input views and lighting using consumer hardware. For input views, we use a Pixel 3A phone to record video of an object while walking around it, using the built-in camera app and the H.264/AVC video format.
\revision{For lighting capture, we place a Insta360 ONE camera in the same position as the object, take an exposure bracket of the environment, and fuse these into a HDR environment map. The supplemental material shows our setup.}

\vspace{-0.1cm}
\paragraph{Reconstruction methodology}
We sample every 60th frame of the video, use COLMAP to create the geometry initialization and estimate each frame's associated camera pose, and align the environment map manually.
To account for different camera responses between real and simulated cameras, we optimize for an environment map brightening factor during the first round of texture estimation.

\vspace{-0.1cm}
\paragraph{Results}
Fig.~\ref{fig:real_world_results} shows results from our real-world reconstruction experiment.
Despite the relatively unconstrained capture set up, our prototype recovers spatially varying texture detail, and a more accurate geometry prediction as a refinement to the COLMAP reconstructions. These reconstructions do not recover the same level of detail that they do in simulated environments; we expect some inaccuracy in both geometry and SVBRDF predictions due to compounding error in COLMAP's camera calibration and our environment map alignment. Improving these parts of scene set-up is required to recover finer-scale detail.

\section{Discussion}
\label{sec:discussion}

Our investigations assume known camera positions and some estimate of lighting conditions. This constraint might be relaxed in the future with simultaneous optimization of camera position, texture, geometry, and lighting. 
\revision{In addition, our optimizations require a reasonable initialization so that differentiable path tracing can produce meaningful gradients.} As such, we see it as a refinement technique.
\revision{Our method is comparable in runtime to a dense COLMAP initialization, taking 3--5 hours depending on the resolution and number of views.} This is expected due to the natural expense of path tracing and the added expense of backpropagating gradients.
Recent ray tracing hardware may offer improvements to differentiable path tracing speed.
Finally, our current mesh colors representation is not mipmapped or anisotropically filtered, and these additions would improve texture estimation across pixels of varied depth.

One avenue of future work is more complex materials, especially those exhibiting ideal reflection/refraction, subsurface scattering, or volumetric effects. 
Another possible area of investigation is to raise the ceiling on object reconstruction size, broadening the scope of possible targets from single small objects to larger scenes like rooms and city landscapes. The sparsity of the mesh representation we use lends itself to such reconstruction tasks. This would require a reevaluation of lighting environment and viewing angle assumptions.
Another possible direction is how best to combine learned neural network priors with our methodology to leverage the advantages of both. For instance, learned priors can encode artistic intent and class-specific patterns, while our approach recovers fine-scale geometry details. A hybrid approach could lead to flexible physically-accurate reconstruction systems.

\section{Conclusion}

\revision{We have investigated how to use differentiable path tracing to jointly estimate the shape and material of a 3D object under known lighting conditions from a series of target images. Starting from a coarse geometry initialization, we alternate between texture and geometry steps and gradually increase the parameter space for optimization. We motivate pipeline stages with several experiments, and show that optimizing over global illumination effects can help handle interreflections and self shadows in reconstruction. We find that optimization via a differentiable path tracer is a promising avenue of research for shape and material reconstruction in unconstrained settings}.
Finally, we show that our method can refine results on real-world data from largely unconstrained capture setups using smartphone videos.
\section*{Acknowledgments}

\revision{We would like to thank the anonymous reviewers for their thoughtful and helpful suggestions. We also thank Prof.~Min H.~Kim for discussions about object capture and reconstruction, and Prof.~Tzu-Mao~Li for discussions about differentiable path tracing. 
This research was supported by GPU donations from NVIDIA and computer donations from Valve Corporation.}




{\small
\bibliographystyle{ieee}
\bibliography{main}

\begin{thebibliography}{10}\itemsep=-1pt

\bibitem{InversePathTracing}
D.~Azinovi\'c, T.-M. Li, A.~Kaplanyan, and M.~Nie{\ss}ner.
\newblock Inverse path tracing for joint material and lighting estimation.
\newblock In {\em CVPR}, 2019.

\bibitem{9105209}
C.~{Che}, F.~{Luan}, S.~{Zhao}, K.~{Bala}, and I.~{Gkioulekas}.
\newblock Towards learning-based inverse subsurface scattering.
\newblock In {\em 2020 IEEE International Conference on Computational
  Photography (ICCP)}, pages 1--12, 2020.

\bibitem{DIBR}
W.~Chen, J.~Gao, H.~Ling, E.~Smith, J.~Lehtinen, A.~Jacobson, and S.~Fidler.
\newblock Learning to predict 3d objects with an interpolation-based
  differentiable renderer.
\newblock In {\em Advances In Neural Information Processing Systems}, 2019.

\bibitem{Gaurav}
A.~Dib, G.~Bharaj, J.~Ahn, C.~Thebault, P.-H. Gosselin, and L.~Chevallier.
\newblock Face reflectance and geometry modeling via differentiable ray
  tracing, 2019.

\bibitem{gao2019deep}
D.~Gao, X.~Li, Y.~Dong, P.~Peers, K.~Xu, and X.~Tong.
\newblock Deep inverse rendering for high-resolution {SVBRDF} estimation from
  an arbitrary number of images.
\newblock {\em ACM Transactions on Graphics (TOG)}, 38(4):134, 2019.

\bibitem{QuadricErrorSimplification}
M.~Garland and P.~S. Heckbert.
\newblock Surface simplification using quadric error metrics.
\newblock In {\em Proceedings of the 24th Annual Conference on Computer
  Graphics and Interactive Techniques}, SIGGRAPH '97, pages 209--216, 1997.

\bibitem{FitMorphableModel}
K.~Genova, F.~Cole, A.~Maschinot, A.~Sarna, D.~Vlasic, and W.~T. Freeman.
\newblock Unsupervised training for 3d morphable model regression.
\newblock In {\em The IEEE Conference on Computer Vision and Pattern
  Recognition (CVPR)}, June 2018.

\bibitem{Goldman2010}
D.~B. {Goldman}, B.~{Curless}, A.~{Hertzmann}, and S.~M. {Seitz}.
\newblock Shape and spatially-varying {BRDFs} from photometric stereo.
\newblock {\em IEEE Transactions on Pattern Analysis and Machine Intelligence},
  32(6):1060--1071, June 2010.

\bibitem{guarnera2016brdf}
D.~Guarnera, G.~C. Guarnera, A.~Ghosh, C.~Denk, and M.~Glencross.
\newblock {BRDF} representation and acquisition.
\newblock In {\em Computer Graphics Forum}, volume~35, pages 625--650. Wiley
  Online Library, 2016.

\bibitem{DIRT}
P.~Henderson and V.~Ferrari.
\newblock Learning single-image {3D} reconstruction by generative modelling of
  shape, pose and shading.
\newblock {\em International Journal of Computer Vision}, 2019.

\bibitem{DeepMVS}
P.~Huang, K.~Matzen, J.~Kopf, N.~Ahuja, and J.~Huang.
\newblock Deepmvs: Learning multi-view stereopsis.
\newblock {\em CoRR}, abs/1804.00650, 2018.

\bibitem{SDFDiff}
Y.~Jiang, D.~Ji, Z.~Han, and M.~Zwicker.
\newblock Sdfdiff: Differentiable rendering of signed distance fields for 3d
  shape optimization, 2019.

\bibitem{Kang2019IllumMultiplexing}
K.~Kang, C.~Xie, C.~He, M.~Yi, M.~Gu, Z.~Chen, K.~Zhou, and H.~Wu.
\newblock Learning efficient illumination multiplexing for joint capture of
  reflectance and shape.
\newblock {\em ACM Trans. Graph.}, 38(6), Nov. 2019.

\bibitem{kato2018renderer}
H.~Kato, Y.~Ushiku, and T.~Harada.
\newblock Neural 3d mesh renderer.
\newblock In {\em The IEEE Conference on Computer Vision and Pattern
  Recognition (CVPR)}, 2018.

\bibitem{PoissonSurfaceReconstruction}
M.~Kazhdan, M.~Bolitho, and H.~Hoppe.
\newblock Poisson surface reconstruction.
\newblock In {\em Proceedings of the fourth Eurographics symposium on Geometry
  processing}, pages 61--70. Eurographics Association, 2006.

\bibitem{SpaceCarving}
K.~N. Kutulakos and S.~M. Seitz.
\newblock A theory of shape by space carving.
\newblock {\em Int. J. Comput. Vision}, 38(3):199--218, July 2000.

\bibitem{langguth2016shading}
F.~Langguth, K.~Sunkavalli, S.~Hadap, and M.~Goesele.
\newblock Shading-aware multi-view stereo.
\newblock In {\em European Conference on Computer Vision}, pages 469--485.
  Springer, 2016.

\bibitem{redner}
T.-M. Li, M.~Aittala, F.~Durand, and J.~Lehtinen.
\newblock Differentiable monte carlo ray tracing through edge sampling.
\newblock {\em ACM Trans. Graph. (Proc. SIGGRAPH Asia)}, 37(6):222:1--222:11,
  2018.

\bibitem{MaterialsForMasses}
Z.~Li, K.~Sunkavalli, and M.~Chandraker.
\newblock Materials for masses: {SVBRDF} acquisition with a single mobile phone
  image.
\newblock {\em CoRR}, abs/1804.05790, 2018.

\bibitem{li2018learning}
Z.~Li, Z.~Xu, R.~Ramamoorthi, K.~Sunkavalli, and M.~Chandraker.
\newblock Learning to reconstruct shape and spatially-varying reflectance from
  a single image.
\newblock In {\em SIGGRAPH Asia 2018 Technical Papers}, page 269. ACM, 2018.

\bibitem{Lin2019MaterialAcquisition}
Y.~Lin, P.~Peers, and A.~Ghosh.
\newblock On-site example-based material appearance acquisition.
\newblock {\em Computer Graphics Forum}, 38(4):15--25, 2019.

\bibitem{SoftRas}
S.~Liu, T.~Li, W.~Chen, and H.~Li.
\newblock Soft rasterizer: A differentiable renderer for image-based 3d
  reasoning.
\newblock {\em The IEEE International Conference on Computer Vision (ICCV)},
  Oct 2019.

\bibitem{FieldProbing}
S.~Liu, S.~Saito, W.~Chen, and H.~Li.
\newblock Learning to infer implicit surfaces without 3d supervision.
\newblock In H.~Wallach, H.~Larochelle, A.~Beygelzimer, F.~d\textquotesingle
  Alch\'{e}-Buc, E.~Fox, and R.~Garnett, editors, {\em Advances in Neural
  Information Processing Systems 32}, pages 8293--8304. Curran Associates,
  Inc., 2019.

\bibitem{Lombardi_2016}
S.~Lombardi and K.~Nishino.
\newblock Radiometric scene decomposition: Scene reflectance, illumination, and
  geometry from rgb-d images.
\newblock {\em 2016 Fourth International Conference on 3D Vision (3DV)}, Oct
  2016.

\bibitem{NeuralVolumes}
S.~Lombardi, T.~Simon, J.~Saragih, G.~Schwartz, A.~Lehrmann, and Y.~Sheikh.
\newblock Neural volumes: Learning dynamic renderable volumes from images.
\newblock {\em ACM Trans. Graph.}, 38(4):65:1--65:14, July 2019.

\bibitem{OpenDR}
M.~M. Loper and M.~J. Black.
\newblock Opendr: An approximate differentiable renderer.
\newblock In D.~Fleet, T.~Pajdla, B.~Schiele, and T.~Tuytelaars, editors, {\em
  ECCV}, 2014.

\bibitem{Mallett2019}
I.~Mallett, L.~Seiler, and C.~Yuksel.
\newblock Patch textures: Hardware implementation of mesh colors.
\newblock In {\em High-Performance Graphics (HPG 2019)}. The Eurographics
  Association, 2019.

\bibitem{mildenhall2020nerf}
B.~Mildenhall, P.~P. Srinivasan, M.~Tancik, J.~T. Barron, R.~Ramamoorthi, and
  R.~Ng.
\newblock Nerf: Representing scenes as neural radiance fields for view
  synthesis, 2020.

\bibitem{Nam2018PracticalAcquisition}
G.~Nam, J.~H. Lee, D.~Gutierrez, and M.~H. Kim.
\newblock Practical {SVBRDF} acquisition of 3d objects with unstructured flash
  photography.
\newblock {\em ACM Trans. Graph.}, 37(6), Dec. 2018.

\bibitem{Mitsuba2}
M.~Nimier-David, D.~Vicini, T.~Zeltner, and W.~Jakob.
\newblock Mitsuba 2: A retargetable forward and inverse renderer.
\newblock {\em Transactions on Graphics (Proceedings of SIGGRAPH Asia)}, 38(6),
  Nov. 2019.

\bibitem{park2020seeing}
J.~J. Park, A.~Holynski, and S.~Seitz.
\newblock Seeing the world in a bag of chips, 2020.

\bibitem{Pix2Vex}
F.~Petersen, A.~H. Bermano, O.~Deussen, and D.~Cohen{-}Or.
\newblock Pix2vex: Image-to-geometry reconstruction using a smooth
  differentiable renderer.
\newblock {\em CoRR}, abs/1903.11149, 2019.

\bibitem{Pytorch3D}
N.~Ravi, J.~Reizenstein, D.~Novotny, T.~Gordon, W.-Y. Lo, J.~Johnson, and
  G.~Gkioxari.
\newblock Pytorch3d.
\newblock \url{https://github.com/facebookresearch/pytorch3d}, 2020.

\bibitem{Rhodin_2015_ICCV}
H.~Rhodin, N.~Robertini, C.~Richardt, H.-P. Seidel, and C.~Theobalt.
\newblock A versatile scene model with differentiable visibility applied to
  generative pose estimation.
\newblock In {\em The IEEE International Conference on Computer Vision (ICCV)},
  December 2015.

\bibitem{schoenberger2016sfm}
J.~L. Sch\"{o}nberger and J.-M. Frahm.
\newblock Structure-from-motion revisited.
\newblock In {\em Conference on Computer Vision and Pattern Recognition
  (CVPR)}, 2016.

\bibitem{schoenberger2016mvs}
J.~L. Sch\"{o}nberger, E.~Zheng, M.~Pollefeys, and J.-M. Frahm.
\newblock Pixelwise view selection for unstructured multi-view stereo.
\newblock In {\em European Conference on Computer Vision (ECCV)}, 2016.

\bibitem{MultiViewStereoComparison}
S.~M. Seitz, B.~Curless, J.~Diebel, D.~Scharstein, and R.~Szeliski.
\newblock A comparison and evaluation of multi-view stereo reconstruction
  algorithms.
\newblock In {\em Proceedings of the 2006 IEEE Computer Society Conference on
  Computer Vision and Pattern Recognition - Volume 1}, CVPR '06, pages
  519--528, Washington, DC, USA, 2006. IEEE Computer Society.

\bibitem{sitzmann2019scene}
V.~Sitzmann, M.~Zollhöfer, and G.~Wetzstein.
\newblock Scene representation networks: Continuous 3d-structure-aware neural
  scene representations, 2019.

\bibitem{torrancesparrow}
K.~Torrance and E.~Sparrow.
\newblock Theory for off-specular reflection from roughened surfaces.
\newblock {\em Journal of The Optical Society of America}, 57, 09 1967.

\bibitem{Tunwattanapong2013}
B.~Tunwattanapong, G.~Fyffe, P.~Graham, J.~Busch, X.~Yu, A.~Ghosh, and
  P.~Debevec.
\newblock Acquiring reflectance and shape from continuous spherical harmonic
  illumination.
\newblock {\em ACM Trans. Graph.}, 32(4), July 2013.

\bibitem{TensorflowGraphics}
J.~Valentin, C.~Keskin, P.~Pidlypenskyi, A.~Makadia, A.~Sud, and S.~Bouaziz.
\newblock Tensorflow graphics: Computer graphics meets deep learning.
\newblock 2019.

\bibitem{egt.20161032}
M.~Weinmann, F.~Langguth, M.~Goesele, and R.~Klein.
\newblock {Advances in Geometry and Reflectance Acquisition}.
\newblock In A.~Sousa and K.~Bouatouch, editors, {\em EG 2016 - Tutorials}. The
  Eurographics Association, 2016.

\bibitem{wu2011high}
C.~Wu, B.~Wilburn, Y.~Matsushita, and C.~Theobalt.
\newblock High-quality shape from multi-view stereo and shading under general
  illumination.
\newblock In {\em CVPR 2011}, pages 969--976. IEEE, 2011.

\bibitem{wu2019detectron2}
Y.~Wu, A.~Kirillov, F.~Massa, W.-Y. Lo, and R.~Girshick.
\newblock Detectron2.
\newblock \url{https://github.com/facebookresearch/detectron2}, 2019.

\bibitem{RSR-UI}
R.~Xia, Y.~Dong, P.~Peers, and X.~Tong.
\newblock Recovering shape and spatially-varying surface reflectance under
  unknown illumination.
\newblock {\em ACM Transactions on Graphics}, 35(6), December 2016.

\bibitem{xu2019deep}
Z.~Xu, S.~Bi, K.~Sunkavalli, S.~Hadap, H.~Su, and R.~Ramamoorthi.
\newblock Deep view synthesis from sparse photometric images.
\newblock {\em ACM Transactions on Graphics (TOG)}, 38(4):1--13, 2019.

\bibitem{PerspectiveTransformerNets}
X.~Yan, J.~Yang, E.~Yumer, Y.~Guo, and H.~Lee.
\newblock Perspective transformer nets: Learning single-view {3D} object
  reconstruction without {3D} supervision.
\newblock In {\em Advances in Neural Information Processing Systems (NeurIPS)},
  2016.

\bibitem{Yuksel:2010:MC:1731047.1731053}
C.~Yuksel, J.~Keyser, and D.~H. House.
\newblock Mesh colors.
\newblock {\em ACM Trans. Graph.}, 29(2):15:1--15:11, Apr. 2010.

\bibitem{DiffRadiativeTransfer}
C.~Zhang, L.~Wu, C.~Zheng, I.~Gkioulekas, R.~Ramamoorthi, and S.~Zhao.
\newblock A differential theory of radiative transfer.
\newblock {\em ACM Trans. Graph.}, 38(6), 2019.

\end{thebibliography}
}

\end{document}